%% file: nertstyle.tex
\pdfoutput=1 % ensure pdflatex (for arXiv)

\documentclass[11pt,a4paper]{article}
\usepackage[hyperref]{acl2020}
\usepackage{nert}
\usepackage{times}
\usepackage{microtype}

\usepackage{standalone}% Need standalone package
\usepackage{pdflscape}
\usepackage{afterpage}
\usepackage{tabularx}
\usepackage{xcolor}
\usepackage{txfonts}
\usepackage[T1]{fontenc}

\aclfinalcopy % Uncomment this line for the final submission
\newcommand{\modelname}[1]{\texttt{#1}\xspace}
\newcommand{\testname}[1]{\texttt{#1}\xspace}

\NewDocumentCommand\emojisnake{}{
    \includegraphics[scale=0.4]{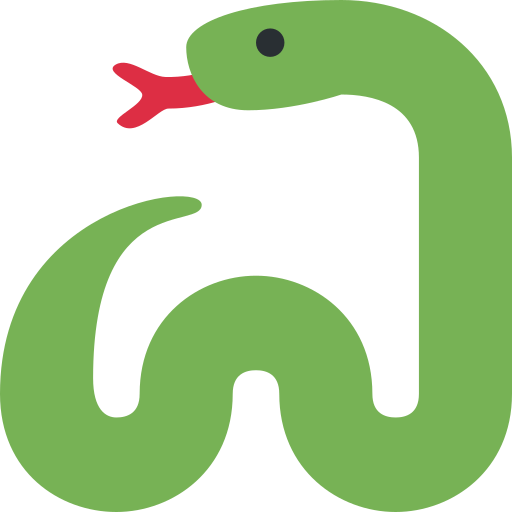}}
\NewDocumentCommand\emojilizard{}{
    \includegraphics[scale=0.4]{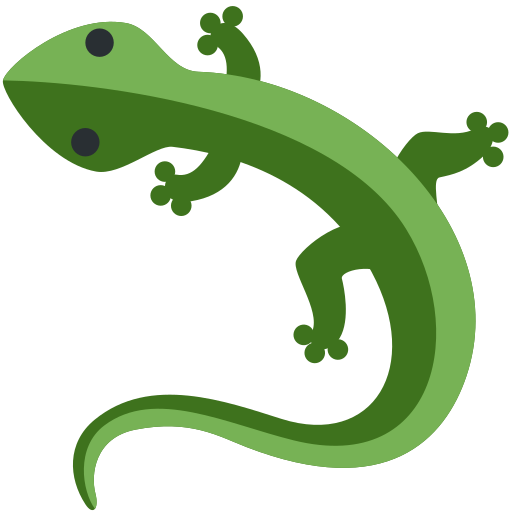}}
\NewDocumentCommand\emojiarrow{}{
    \includegraphics[scale=0.012]{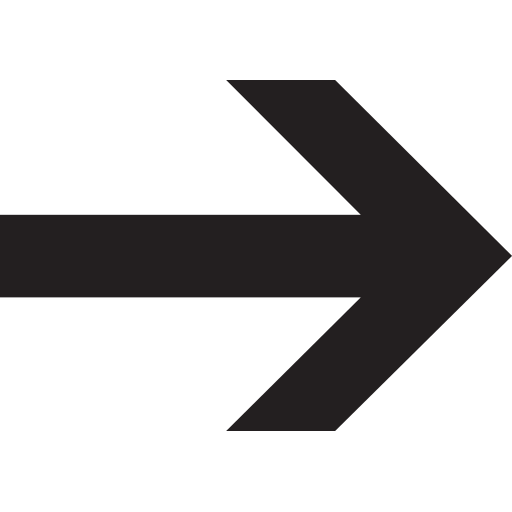}}
\newcommand{\contributor}[2]{\href{mailto:#1}{\textcolor{black}{#2}}}

\definecolor{KDpurple}{rgb}{0.6,0.18,0.64}

\newcommand{\transformation}[2]{\subsection{\href{https://github.com/GEM-benchmark/NL-Augmenter/tree/main/transformations/#1}{#2}}}
\newcommand{\filter}[2]{\subsection{\href{https://github.com/GEM-benchmark/NL-Augmenter/tree/main/filters/#1}{#2}}}

\definecolor{SGcolor}{rgb}{0.3,0.9,0.3}

\newcommand{\veryshortarrow}[1][3pt]{\mathrel{%
   \hbox{\rule[\dimexpr\fontdimen22\textfont2-.2pt\relax]{#1}{.4pt}}%
   \mkern-4mu\hbox{\usefont{U}{lasy}{m}{n}\symbol{41}}}}

\newcommand{\colbox}[2]{\colorbox{#1}{#2}}
\definecolor{caddback}{rgb}{0.90, 0.98, 0.96}
\definecolor{cadd}{rgb}{0, 0.47, 0.34}
\definecolor{cdelback}{rgb}{1, 0.94, 0.92}
\definecolor{cdel}{rgb}{0.83, 0.32, 0.16}
\def \arrow{$\veryshortarrow$}
\newcommand{\add}[1]{\colbox{caddback}{\color{cadd}#1\xspace}} %

\newcommand{\remove}[1]{\colbox{cdelback}{{\color{cdel}#1\xspace}}}%

\title{NL-Augmenter \emojilizard \emojiarrow \emojisnake \\ A Framework for Task-Sensitive Natural Language Augmentation \\[15pt]
\smaller\normalfont\em December 5, 2021
}

\author{
\begin{minipage}[t]{\textwidth}
\centering
\normalsize

\contributor{kdhole@emory.edu}{Kaustubh~D.~Dhole}$^{3,18\dagger}$,
\contributor{vgangal@cs.cmu.edu}{Varun~Gangal}$^{7\dagger}$,
\contributor{gehrmann@google.com}{Sebastian~Gehrmann}$^{23\dagger}$,
\contributor{aadesh.gupta@ipsoft.com}{Aadesh~Gupta}$^{3\dagger}$,
\contributor{zhenhao.li18@imperial.ac.uk}{Zhenhao~Li}$^{32\dagger}$,
\contributor{saad.mahamood@trivago.com}{Saad~Mahamood}$^{90\dagger}$,
\contributor{abinaya.m02@mphasis.com}{Abinaya~Mahendiran}$^{45\dagger}$,
\contributor{simon.mille@upf.edu}{Simon~Mille}$^{53\dagger}$,
\contributor{ashish3586@gmail.com}{Ashish~Shrivastava}$^{2\dagger}$,
\contributor{samson.tan@salesforce.com}{Samson~Tan}$^{48,91\dagger}$
% ...

{\footnotesize \normalfont 

% ...
$^{2}$Agara,
$^{3}$Amelia R\&D, New York,
% ...
$^{7}$Carnegie Mellon University,
% ...
$^{18}$Emory University,
% ...
$^{23}$Google Research,
% ...
$^{32}$Imperial College, London,
% ...
$^{45}$Mphasis NEXT Labs,
% ...
$^{48}$National University of Singapore,
% ...
$^{53}$Pompeu Fabra University,
% ...
$^{90}$trivago N.V.,
$^{91}$Salesforce Research Asia,
% ...
}
\end{minipage}
}

\date{}

\begin{document}
\maketitle
\blfootnote{\textbf{\dagger} Organizers \& Steering Committee}
\blfootnote{\textbf{*} Please send requests to the correspondence email: \UrlFont{nlaugmenter@googlegroups.com}.}

\begin{abstract}
Data augmentation is an important component in the robustness evaluation of models in natural language processing (NLP) and in enhancing the diversity of the data they are trained on. In this paper, we present NL-Augmenter, a new participatory Python-based natural language augmentation framework which supports the creation of both transformations (modifications to the data) and filters (data splits according to specific features). We describe the framework and an initial set of $117$ transformations and $23$ filters for a variety of natural language tasks. We demonstrate the efficacy of NL-Augmenter by 
using several of its tranformations to analyze the robustness of popular natural language models. 
The infrastructure, datacards and robutstness analysis results are available publicly on the NL-Augmenter repository (\url{https://github.com/GEM-benchmark/NL-Augmenter}).
\end{abstract}

\begin{figure*}[!thb]
    \centering
    \includegraphics[width=\textwidth]{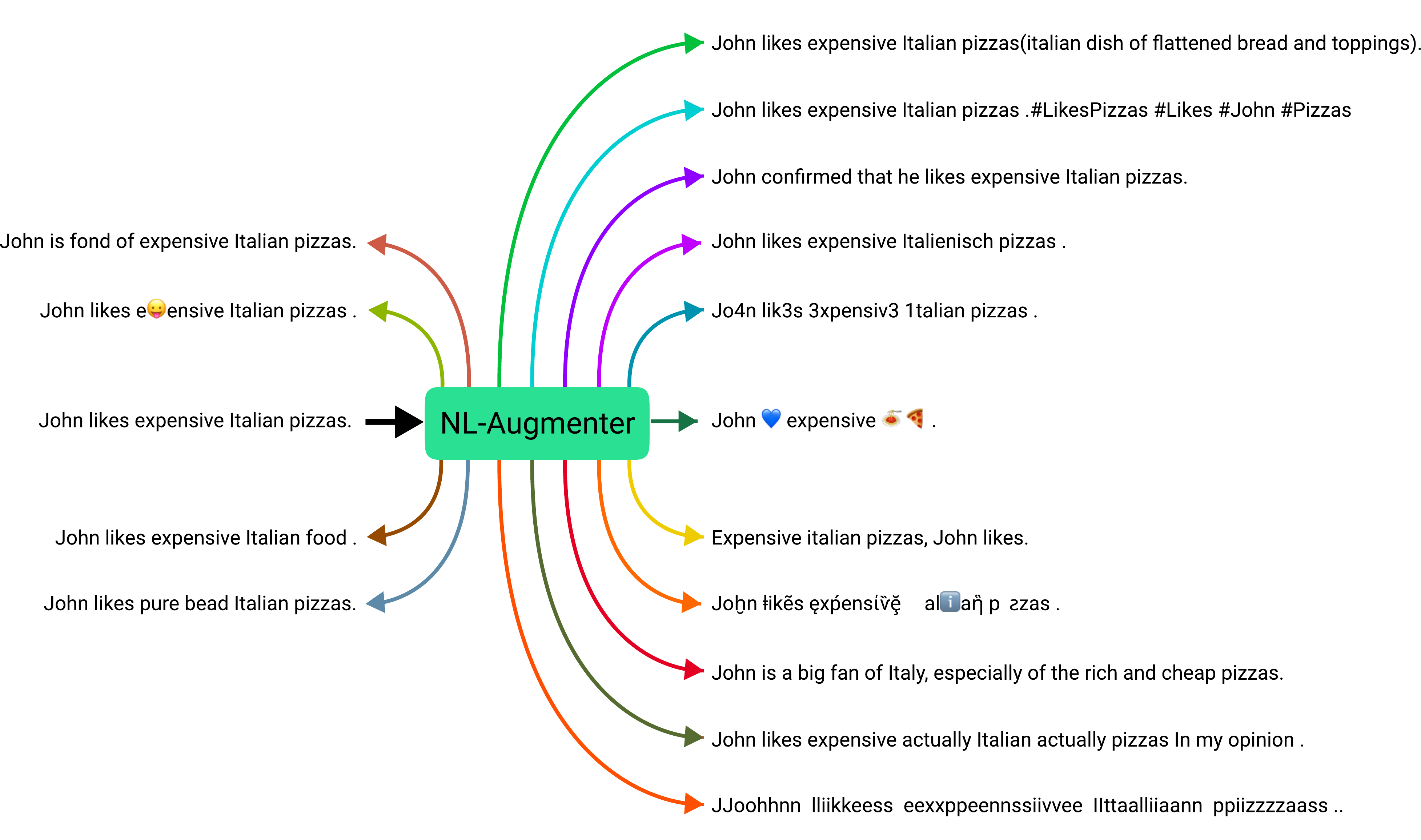}
    \caption{A few randomly chosen transformations of NL-Augmenter for the original sentence \textit{John likes expensive pizzas.} While the meaning (almost) always remains the same and identifiable by humans, models can have a much harder time representing the transformed sentences.}
    \label{fig:randomtr}
\end{figure*}

\section{Introduction}
Data augmentation, the act of creating new datapoints by slightly modifying copies or creating synthetic data based on existing data, is an important component in the robustness evaluation of models in natural language processing (NLP) and in enhancing the diversity of their training data.
Most data augmentation techniques create examples through transformations of existing examples which are based on prior task-specific knowledge~\cite{feng2021survey, chen2021empirical}.  Such transformations seek to disrupt model predictions or can be used as training candidates for improving regularization and denoising models, for example, through consistency training~\cite{xie2020unsupervised}; Figure~\ref{fig:randomtr} shows a few possible transformations for a sample sentence.

However, a vast majority of transformations do not alter the structure of examples in drastic and meaningful ways, rendering them qualitatively less effective as potential training or test examples.  
Moreover, different NLP tasks may benefit from transforming different linguistic properties. for example, changing the word “happy” to “very happy” in an input is more relevant for sentiment analysis than for summarization~\cite{mille2021automatic}. 
Despite this, many transformations may be universially useful, for example changing places to ones from different geographic regions, or changing names to those from different cultures. 
As such, having a single place to collect both task-specific and task-independent augmentations will ease the barrier to creating appropriate suites of augmentations that should be applied to different tasks.

Natural language and its long-tailed nature~\citep{bamman2017natural} lead to a very high diversity of possible surface forms. If only a handful number of ways to paraphrase a text are available, it can be hard to generalize across radically different surface forms. Besides, data drawn i.i.d. from such long-tailed distribution represents itself exactly in proportion to its occurence in the dataset. Evaluating NLP systems on such data means that the head of the distribution is emphasized even in the test dataset and that rare phenomena are implicitly ignored. 
However, informed transformations or the identification of such tail examples may require a wide range of domain knowledge or specific cultural backgrounds. 
We thus argue that a collection of transformations that should be applied to NLP datasets should be done by capitalizing on the ``wisdom-of-researchers''.

% To that end, 
To enable more diverse and better characterized data during testing and training, 
we create a Python-based natural language augmentation framework, NL-Augmenter\footnote{\url{https://github.com/GEM-benchmark/NL-Augmenter}}.
% , and w
With the help of researchers across subfields in computational linguistics and NLP, we collect many creative ways to augment data for natural language tasks. To encourage task-specific implementations, we tie each transformation to a widely-used data format (e.g. text pair, a question-answer pair, etc.) along with various task types (e.g. entailment, tagging, etc.) that they intend to benefit. We demonstrate the efficacy of NL-Augmenter by
using some of its transformations to analyze the robustness of popular natural language models.
% measuring the capability of some of the transformations for analyzing the robustness of multiple popular natural language models.

A majority of the augmentations that the framework supports are transformations of single sentences that aim to paraphrase these sentences in various ways. NL-Augmenter loosens the definition of ``transformations'' from the logic-centric view of strict equivalence to the more descriptive view of linguistics, closely resembling~\citet{bhagat2013paraphrase}'s ``quasi-paraphrases''. We extend this to accommodate noise, intentional and accidental human mistakes, sociolinguistic variation, semantically-valid style, syntax changes, as well as artificial constructs that are unambiguous to humans~\citep{tan-etal-2021-reliability}. Some transformations vary the socio-linguistic perspective permitting a crucial source of variation wherein language goals span beyond conveying ideas and content. 

%\sg{Maybe add something about filters here?} 
In addition to transformations, NL-Augmenter also provides a variety of filters, which can be used to filter data and create subpopulations of given inputs, according to features such as input complexity, input size, etc. Unlike a transformation, the output of a filter is a boolean value, indicating whether the input meets the filter criterion, e.g. whether the input text is toxic. The filters allow splitting existing datasets and hence evaluating models on subsets with specific linguistic properties.

In this paper, we apply the collected transformations and filters to several datasets and show to what exent the different types of perturbations do affect some models.
 
The paper is organized as follows. We first discuss the rise of participatory benchmarks in Section~\ref{sec:related_work}. In Section~\ref{sec:nl_augmenter}, we introduce the participatory workshop and the repository of NL-Augmenter. In Section~\ref{sec:robustness_analysis}, we present the robustness analysis performed on the participants' submissions, and in Section~\ref{sec:robustness_analysis} we provide a broader impact discussion. All filters and transformations are listed and described in details in the Appendix.

\section{Related Work}
NL-Augmenter enables both data augmentation and robustness testing by constructing the library in a \emph{participatory} fashion. We provide an overview of the related work in these lines of research.
\label{sec:related_work}
\input{sections/related_work.tex}

\begin{figure*}[!thb]
    \centering
    \includegraphics[width=0.5\textwidth]{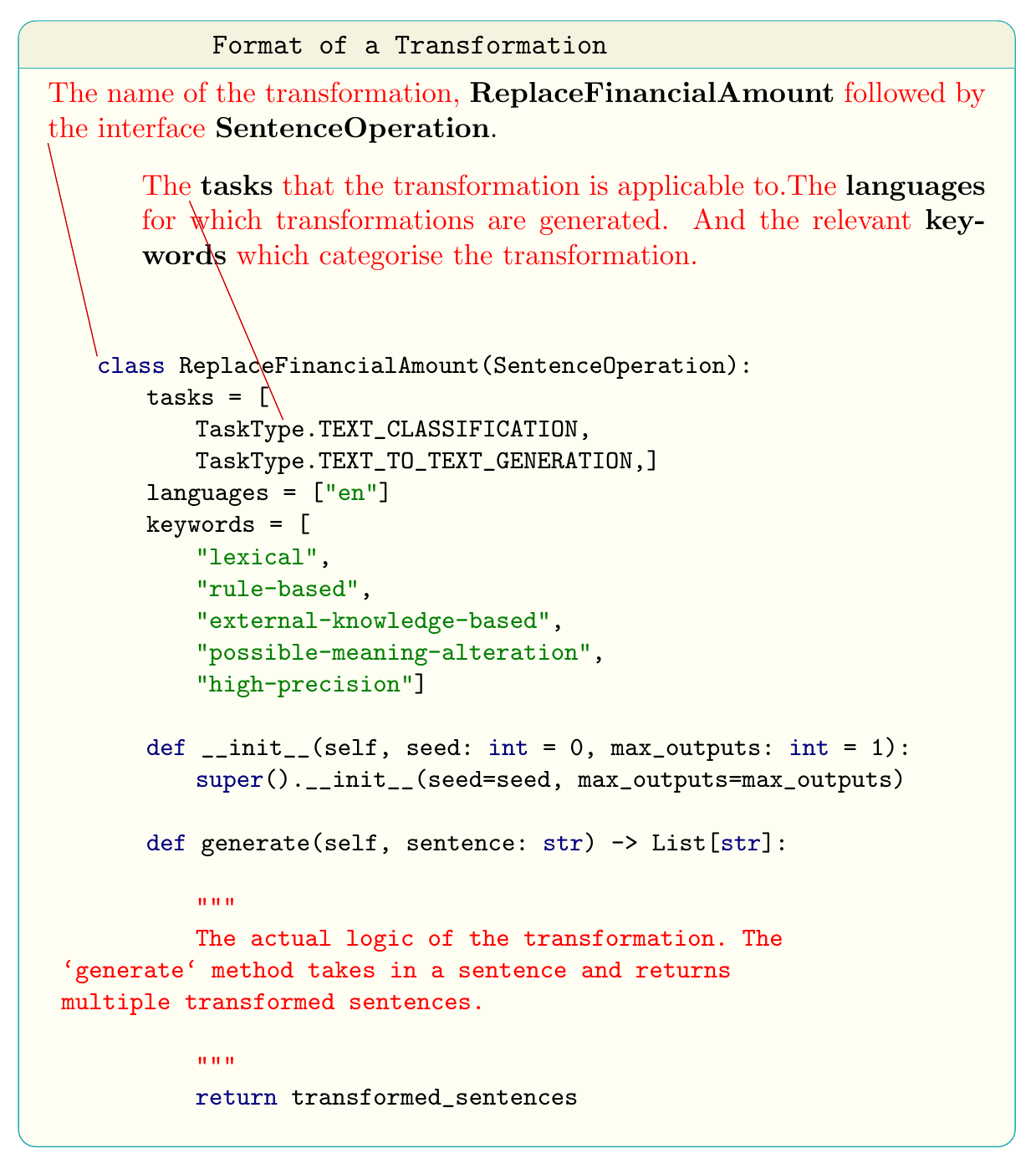}
    \caption{Participants were expected to write their python class adhering to the above format.}
    \label{fig:tr}
\end{figure*}

\section{NL-Augmenter \emojilizard \emojiarrow \emojisnake}
\label{sec:nl_augmenter}
NL-Augmenter is a crowd-sourced suite to facilitate rapid augmentation of data for NLP tasks to assist in training and evaluating models. NL-augmenter was introduced in~\citet{mille2021automatic} in the context of the creation of evaluation suites for the GEM benchmark~\citep{gehrmann2021gem, gem2-0}; three types of evaluation sets were proposed: (i) transformations, i.e. original test sets are perturbed in different ways (e.g. backtranslation, introduction of typographical errors, etc.), (ii) subpopulations, i.e. test subsets filtered according to features such as input complexity, input size, etc.; and (iii) data shifts, i.e. new test sets that do not contain any of the original test set material.

In this paper, we present a participant-driven repository for creating and testing \textbf{transformations} and \textbf{filters}, and for applying them to all dataset splits (training, development, evaluation) and to all NLP tasks (NLG, labeling, question answering, etc.). As shown by~\citet{mille2021automatic}, applying filters and tranformations to development/evaluation data splits allows for testing the robustness of models and for identifying possible biases; on the other hand, applying transformations and filters to training data (data augmentation) allows for possibly mitigating the detected robustness and bias issues~\cite{wang-etal-2021-closer, pruksachatkun-etal-2021-robustness,si-etal-2021-better}.

In this section, we provide organizational details, list the transformations and filters that the repository currently contains, and we present the list of tags we associated to tranformations and filters and how we introduced them.

\subsection{Participatory Workshop on GitHub}

A workshop was organized towards constructing this full-fledged participant-driven repository. Unlike a traditional workshop wherein people submit papers, participants were asked to submit python implementations of transformations to the GitHub repository. Organizers of this workshop created a base repository extending~\citet{mille2021automatic}'s NLG evaluation suite and incorporated a set of~\textit{interfaces}, each of which catered to popular NL example formats. This formed the backbone of the repository. A sample set of transformations and filters alongwith evaluation scripts were provided as starter code. Figure~\ref{fig:tr} show an annotated code snippet of a submission.
Following the format of BIG-Bench's review process, multiple review criteria were designed for accepting contributions. The review criteria (see Appendix~\ref{sec:append-review}) guided participants to follow a style guide, incorporate test cases in JSON format, and encouraged novelty and specificity. Apart from the general software development advantages of test cases, they made reviewing simpler by providing an overview of the transformation's capability and scope of generations.

\input{tf-table-full-list}

\subsection{Transformations and filters}
\label{subsec:tranfo-and-filters}

\input{filters-table-full-list}

Tables~\ref{tab:transformations-list} and \ref{tab:filters-list} list respectively the 117 transformations and 23 filters that are currently found in the NL-Augmenter repository (alphabetically ordered according to the submission name in the repository). For each transformation/filter, a link to the corresponding Appendix subsection is provided, where a detailed description, illustrations and an exernal link to implementation in the NL-Augmenter repository can be found. 

\input{sections/tags-perturbations}

\input{sections/robustness-tables-pre}

\section{Robustness Analysis}
\label{sec:robustness_analysis}
All authors of the accepted perturbations were asked to provide the task performance scores for each of their respective transformations or filters. In Section~\ref{subsec:robustness-analysis-experiment} we provide details on how the scores were obtained, and in Section~\ref{subsec:results} we provide a first analysis of these scores.
\subsection{Experiment}
\label{subsec:robustness-analysis-experiment}
The perturbations are currently split into three groups, according to the task(s) they will be evaluated on: text classification tasks, tagging tasks, and question-answering tasks. For experiments in this paper, we focus on text classification and on the relevant perturbations. We compare the models' performance on the original data and on the perturbed data. The percentage of sentences being changed by a transformation (\textit{transformation rate}) and the percentage of performance drop on the perturbed data compared to the performance on the original data (\textit{score variation}) are reported.
%\zl{The definition of score variation to be confirmed.}\st{this is usually called robust accuracy} 

\textbf{Tasks.}
We choose four evaluation datasets among three English NLP tasks: 
(1) sentiment analysis on both short sentences (SST-2~\cite{socher2013recursive}) and full paragraphs (IMDB Movie Review~\cite{maas2011learning}), (2) Duplicate question detection (QQP)~\cite{wang2018glue}, and (3) Natural Language Inference (MNLI)~\cite{williams2017broad}. 
These tasks cover both classifications on single sentences, as well as pairwise comparisons, and have been widely used in various counterfactual analysis and augmentation experiments~\cite{wu2021polyjuice, kaushik2019learning, gardner2020evaluating, ribeiro2020beyond}. 

\textbf{Evaluation models}. We represent each dataset/task with its corresponding most downloaded large model hosted on Huggingface~\cite{wolf-etal-2020-transformers}, resulting in four models for evaluation:
\modelname{roberta-base-SST-2}, \modelname{roberta-base-imdb}, \modelname{roberta-large-mnli}, and \modelname{bert-base-uncased-QQP}.\footnote{\UrlFont{huggingface.co/\{\\
textattack/roberta-base-SST-2,
textattack/roberta-base-imdb,
textattack/bert-base-uncased-QQP,
roberta-large-mnli
\}}}

\textbf{Perturbation strategy.}
For each task, we perturb a random sample of 20\% of the validation set.
Since all the transformations are on single text snippets, for datasets with sentence pairs, i.e., QQP and MNLI, we perturb the first question and the premise sentence, respectively.

\input{sections/robustness.tex}

\section{Discussion and Broader Impact}

\paragraph{Limitations} In Section~\ref{subsec:results}, we analyze the results of applying some of the transformations on existing datasets and running models on the perturbed data. Even though it was not possible to test all of the currently existing perturbations (mostly due to time constraints), the overall results show that the tested perturbations do pose a challenge to different models on different tasks, with quasi-systematic score drops. However, with so many transformations applied to four different datasets, the presented robustness analysis can only be shallow, and a separate analysis of each transformation would be needed in order to get more informative insights. Second, our superficial analysis above relies on tags which were in many cases annotated by hand, and some of the suprising results (e.g. meaning-preserving transformations are more challenging than non-meaning-preserving ones) may reflect a lack of consistency in the annotations. We believe that assessing the quality of the tag assignment so as to ensure a high inter-annotator agreement will be needed for reliable analyses in the future. Finally, the current robustness analysis only shows that the perturbations are effective for detecting a possible weakness in a model; further experiments are needed to demonstrate that the perturbations can also help mitigating the weaknesses they bring to light.

\paragraph{Dilution of Contributions} While this is not our intent, there is a risk in large scale collections of work like this that individual contributions are being less appreciated than releasing them as a standalone project. This risk is a tradeoff with the advantage that it becomes much easier to switch between different transformations, which can lead to a better adoption of introduced methods. To proactively give appropriate credit, each transformation has a data card mentioning the contributors and all participants are listed as co-authors of this paper. We further encourage all users of our repository to cite the work that a specific implementation builds on, if appropriate. The relevant citations are listed on the respective data cards and in the description in the appendix. In the same vein, there is a risk of NL-Augmenter as a whole to monopolize the augmentation space due to its large scope, leading to less usage of related work which may cover additional transformations or filters. While this is not our intention and we actively worked with contributors to related repositories to integrate their work, we encourage researchers to try other solutions as well. 

\paragraph{Participatory Setup} Conducting research in environments with a shared mission, a low barrier of entry, and directly involving affected communities was popularizied by \citet{nekoto-etal-2020-participatory}. This kind of participatory work has many advantages, most notably that it changes the typically prescriptive research workflow toward a more inclusive one. Another advantage is that through open science, anyone can help shape the overall mission and improve the end result. Following the related BIG-bench \cite{srivastava2022beyond} project, we aimed to design NL-Augmenter in a similar spirit -- by providing the infrastructure, the participation barrier is reduced to filling a templated interface and providing test example. By making the interface as flexible as possible, the contributions range from filters for subpopulations with specific protected attributes to transformations via neural style transfer. Through this wide range, we hope that researchers can apply a wider range of augmentation and evaluations strategies to their data and models. 

\section{Conclusion} In this paper, we introduced NL-Augmenter, a framework for text transformations and filters with the goal to assisst in robustness testing and other data augmentation tasks. We demonstrated that through an open participation strategy, NL-Augmenter can cover a much wider set of languages, tasks, transformations, and filters than related work without a loss of focus. 
In total, our repository provides $>$ 117 transformations and $>$ 23 filters which are all documented and tested and will contribute toward more robust NLP models and an evaluation thereof. 
As we point out in our analysis, there is much room to improve NL-Augmenter. We welcome future contributions to improve its coverage of the potential augmentation space and to address its current shortcomings. Future work may further include data augmentation experiments at a larger scale to investigate the effect on model robustness.

\section{Organization}
NL-Augmenter is a large effort organized by researchers and developers ranging across different niches of NLP. To acknowledge everyone’s contributions, we list the contribution statements below for all.

\paragraph{Steering Committee:} Kaustubh Dhole, Varun Gangal, Sebastian Gehrmann, Aadesh Gupta, Zhenhao Li, Saad Mahmood, Simon Mille, Jascha Sohl-Dickstein, Ashish Shrivastava, Samson Tan, Tongshuang Wu and Abinaya Mahendiran make up the steering committee. Jinho D. Choi, Eduard Hovy \& Sebastian Ruder provided guidance and feedback. Kaustubh Dhole coordinates and leads the NL-Augmenter effort. All others provide feedback and discuss larger decisions regarding the direction of NL-Augmenter and act as organizers and reviewers.

\paragraph{Repository:} Kaustubh, Aadesh, Zhenhao, Tongshuang, Ashish, Saad, Varun \& Abinaya created the interfaces and the base repository NL-Augmenter for participants to contribute. This was also a continuation of the repository developed for creating challenge sets~\cite{mille2021automatic} for GEM~\cite{gehrmann2021gem}. All the other authors expanded this repository with their implementations.

%\paragraph{Reliability Testing:} Samson Tan

%\paragraph{AugmentedParaREL:} Varun Gangal

\paragraph{Reviewers:} Kaustubh, Simon, Zhenhao, Sebastian, Varun, Samson, Abinaya, Saad, Tongshuang, Aadesh, Ondrej were involved in reviewing the submissions of participants of the first phase. In the 2nd phase, all other authors performed a cross-review, in which participants were paired with 3 other partcipants. This was followed by a meta review by the organizers.

\paragraph{Robustness Evaluation:} Ashish, Tongshuang, Kaustubh \& Zhenhao created the evaluation engine. Simon, Kaustubh, Saad, Abinaya \& Tongshuang performed the robustness analysis.

\paragraph{Website:} Aadesh and Sebastian created the webpages for the project.

\bibliography{nertstyle}
\bibliographystyle{acl_natbib}

\appendix
\input{sections/appendix-review-criteria}

\input{tf_list}

\input{filter_list}

\end{document}

%% file: sections/related_work.tex
\subsection{Evolving Participatory Benchmarks}
\label{subsec:benchmarks}
%Two primary inspirations CheckList and BigBench 
%\st{Not sure what the talking points are for these two}

To address the problem of under-resourced African languages in machine translation, Masakhane adopted a bottom-up, participatory approach to construct machine translation benchmarks for over thirty languages~\citep{nekoto-etal-2020-participatory}. This collaborative approach is increasingly adopted in the NLP communtiy to create evolving benchmarks in response to the rapid pace of NLP progress. The Generation Evaluation and Metrics benchmark~\citep{gehrmann2021gem}, which started the development of NL-Augmenter, is a participatory project to document and improve tasks and their evaluation in natural language generation. BIG-Bench\footnote{\url{https://github.com/google/BIG-bench}} proposes a collaborative framework to collect a large suite of few-shot tasks to gauge the abilities of large, pretrained language models. DynaBench~\citep{kiela-etal-2021-dynabench} proposes to iteratively evaluate models in a human-in-the-loop fashion by constructing more challenging examples after each round of model evaluation. SyntaxGym  \cite{gauthier2020syntaxgym} provides a standardized platform for researchers to contribute and use evaluation sets and focuses on targeted syntactic evaluation of Language Models (LMs), particularly psycholinguistically motivated ones. In contrast, our transformations operate on a larger variety of tasks and model types --- they are not required to be syntactically or even linguistically motivated.

 \subsection{Wisdom-of-Researchers}
There are many ways to introduce variation in a sentence without altering its meaning; the lived experiences of a diverse group of individuals could help with identifying and codifying the myriad dimensions of variation as executable transformations~\citep{tan-etal-2021-reliability}. Leveraging the wisdom-of-the-crowd \cite{galton1907vox,yi2010wisdom} is common in our field of natural language processing, with the use of Amazon Mechanical Turk to generate and annotate data in exchange for monetary returns. The aforementioned BIG-Bench project, hosted on GitHub, offers co-authorship in exchange for task contribution. We can think of this as a sort of \emph{wisdom-of-researchers}. Similarly, we crowd-source transformations, in the form of Python code snippets, in return for co-authorship.

\subsection{Robustness Evaluation Tools}
There are many projects with similar goals that inspired NL-Augmenter. For example~\citet{gardner2020evaluating} proposed creating ``contrast'' sets of perturbed test examples. In their approach, each example is manually perturbed, which may lead to higher-quality results but is costly to replicate for each new task due to scale and annotator cost. 
TextAttack~\citep{morris2020textattack} is a library enabling the adversarial evaluation of English NLP models. Partially overcoming this limitation, TextFlint~\citep{wang-etal-2021-textflint}, supports robustness evaluation in English and Chinese. It covers linguistic and task-specficic transformations, adversarial attacks, and subpopulation analyses. In contrast, while the majority are focused on English, NL-Augmenter comprises transformations and filters that work for many different languages and each contribution can specify a set of supported languages.

Robustness Gym~\citep{goel2021robustness} unifies four different types of robustness tests --- subpopulations, transformations, adversarial attacks, and evaluation sets --- under a single interface. However, it depends on existing libraries for its transformations. In contrast, NL-Augmenter focuses on compiling a set of transformations in an open source and collaborative fashion, which is reflected in its size and diversity. Checklist~\citep{ribeiro2020beyond} argues for the need to go beyond simple accuracy and evaluate the model on basic linguistic capabilities, for example their response to negations. Polyjuice~\citep{wu2021polyjuice} perturbs examples using GPT-2 --- though this is automatic and scalable, it offers limited control over type of challenging examples generated, making fine-grained analysis beyond global challenge-set level difficult. In contrast, our method offers a richer taxonomy with $117$ (and growing) transformations for extensive analysis and comparison.

\citet{tan-etal-2021-reliability} propose decomposing each real world environment into a set of dimensions before using randomly sampled and adversarially optimized transformations to measure the model's average- and worst-case performance along each dimension. NL-Augmenter can be used, out-of-the-box, to measure average-case performance and we plan to extend it to support worst-case evaluation.

%% file: tf-table-full-list.tex
\begin{table*}[!thb]
	\centering\small
	%So the table takes a full page
	\begin{tabular}{llll}
    & & & \\
    & & & \\
    & & & \\
    & & & \\
    \end{tabular}
	\noindent\begin{tabular}{ll|ll}
	\textbf{Transformation} & \textbf{App.} & \textbf{Transformation} & \textbf{App.}\\
	\toprule
Abbreviation Transformation & \ref{transfo:abbreviation-transformation} & Mix transliteration & \ref{transfo:mix-transliteration}\\
Add Hash-Tags & \ref{transfo:add-hashtags} & MR Value Replacement & \ref{transfo:mr-value-replacement}\\
Adjectives Antonyms Switch & \ref{transfo:adjectives-antonyms-switch} & Multilingual Back Translation & \ref{transfo:multilingual-back-translation}\\
AmericanizeBritishizeEnglish & \ref{transfo:AmericanizeBritishizeEnglish} & Multilingual Dictionary Based Code Switch & \ref{transfo:multilingual-dictionary-based-code-switch}\\
AntonymsSubstitute & \ref{transfo:AntonymsSubstitute} & Multilingual Lexicon Perturbation & \ref{transfo:multilingual-lexicon-perturbation}\\
Auxiliary Negation Removal & \ref{transfo:auxiliary-negation-removal} & Causal Negation and Strengthening & \ref{transfo:negate-strengthen}\\
AzertyQwertyCharsSwap & \ref{transfo:AzertyQwertyCharsSwap} & Question Rephrasing transformation & \ref{transfo:neural-question-paraphraser}\\
BackTranslation & \ref{transfo:back-translation} & English Noun Compound Paraphraser [N+N] & \ref{transfo:noun-compound-paraphraser}\\
BackTranslation for Named Entity Recognition & \ref{transfo:back-translation-ner} & Number to Word & \ref{transfo:number-to-word}\\
Butter Fingers Perturbation & \ref{transfo:butter-fingers-perturbation} & Numeric to Word & \ref{transfo:numeric-to-word}\\
Butter Fingers Perturbation For Indian Languages & \ref{transfo:butter-fingers-perturbation-for-IL} & OCR Perturbation & \ref{transfo:ocr-perturbation}\\
Change Character Case & \ref{transfo:change-char-case} & Add Noun Definition & \ref{transfo:p1-noun-transformation}\\
Change Date Format & \ref{transfo:change-date-format} & Pig Latin Cipher & \ref{transfo:pig-latin}\\
Change Person Named Entities & \ref{transfo:change-person-named-entities} & Pinyin Chinese Character Transcription & \ref{transfo:pinyin}\\
Change Two Way Named Entities & \ref{transfo:change-two-way-ne} & SRL Argument Exchange & \ref{transfo:propbank-srl-roles}\\
Chinese Antonym and Synonym Substitution & \ref{transfo:chinese-antonym-synonym-substitution} & ProtAugment Diverse Paraphrasing & \ref{transfo:protaugment-diverse-paraphrase}\\
Chinese Pinyin Butter Fingers Perturbation & \ref{transfo:chinese-butter-fingers-perturbation} & Punctuation & \ref{transfo:punctuation}\\
Chinese Person NE and Gender Perturbation & \ref{transfo:chinese-person-named-entities-gender} & Question-Question Paraphraser for QA & \ref{transfo:qq-paraphraser}\\
Chinese (Simplified and Traditional) Perturbation & \ref{transfo:chinese-simplified-traditional-perturbation} & Question in CAPS & \ref{transfo:question-in-caps}\\
City Names Transformation & \ref{transfo:city-names-transformation} & Random Word Deletion & \ref{transfo:random-deletion}\\
Close Homophones Swap & \ref{transfo:close-homophones-swap} & Random Upper-Case Transformation & \ref{transfo:random-upper-transformation}\\
Color Transformation & \ref{transfo:color-transformation} & Double Context QA & \ref{transfo:redundant-context-for-qa}\\
Concatenate Two Random Sentences (Bilingual) & \ref{transfo:concat} & Replace Abbreviations and Acronyms & \ref{transfo:replace-abbreviation-and-acronyms}\\
Concatenate Two Random Sentences (Monolingual) & \ref{transfo:concat-monolingual} & Replace Financial Amounts & \ref{transfo:replace-financial-amounts}\\
Concept2Sentence & \ref{transfo:concept2sentence} & Replace Numerical Values & \ref{transfo:replace-numerical-values}\\
Contextual Meaning Perturbation & \ref{transfo:contextual-meaning-perturbation} & Replace Spelling & \ref{transfo:replace-spelling}\\
Contractions and Expansions Perturbation & \ref{transfo:contraction-expansions} & Replace nouns with hyponyms or hypernyms & \ref{transfo:replace-with-hyponyms-hypernyms}\\
Correct Common Misspellings & \ref{transfo:correct-common-misspellings} & Sampled Sentence Additions & \ref{transfo:sentence-additions}\\
Country/State Abbreviation & \ref{transfo:country-state-abbreviation-transformation} & Sentence Reordering & \ref{transfo:sentence-reordering}\\
Decontextualisation of the main Event & \ref{transfo:decontextualize-sentence} & Emoji Addition for Sentiment Data & \ref{transfo:sentiment-emoji-augmenter}\\
Diacritic Removal & \ref{transfo:diacritic-removal} & Shuffle Within Segments & \ref{transfo:shuffle-within-segments}\\
Disability/Differently Abled Transformation & \ref{transfo:disability-transformation} & Simple Ciphers & \ref{transfo:simple-ciphers}\\
Discourse Marker Substitution & \ref{transfo:discourse-marker-substitution} & Slangificator & \ref{transfo:slangificator}\\
Diverse Paraphrase Generation & \ref{transfo:diverse-paraphrase} & Spanish Gender Swap & \ref{transfo:spanish-gender-swap}\\
Dislexia Words Swap & \ref{transfo:dyslexia-words-swap} & Speech Disfluency Perturbation & \ref{transfo:speech-disfluency-perturbation}\\
Emoji Icon Transformation & \ref{transfo:emoji-icon-transformation} & Paraphrasing through Style Transfer & \ref{transfo:style-paraphraser}\\
Emojify & \ref{transfo:emojify} & Subject Object Switch & \ref{transfo:subject-object-switch}\\
English Inflectional Variation & \ref{transfo:english-inflectional-variation} & Sentence Summarizaiton & \ref{transfo:summarization-transformation}\\
English Mention Replacement for NER & \ref{transfo:entity-mention-replacement-ner} & Suspecting Paraphraser for QA & \ref{transfo:suspecting-paraphraser}\\
Filler Word Augmentation & \ref{transfo:filler-word-augmentation} & Swap Characters Perturbation & \ref{transfo:swap-characters}\\
Style Transfer from Informal to Formal & \ref{transfo:formality-change} & Synonym Insertion & \ref{transfo:synonym-insertion}\\
French Conjugation Substitution & \ref{transfo:french-conjugation-transform} & Synonym Substitution & \ref{transfo:synonym-substitution}\\
Gender And Culture Diversity Name Changer & \ref{transfo:gender-culture-diverse-name} & Syntactically Diverse Paraphrasing & \ref{transfo:syntactically-diverse-paraphrase}\\
Neopronoun Substitution & \ref{transfo:gender-neopronouns} & Subsequence Substitution for Seq. Tagging & \ref{transfo:tag-subsequence-substitution}\\
Gender Neutral Rewrite & \ref{transfo:gender-neutral-rewrite} & Tense & \ref{transfo:tense}\\
GenderSwapper & \ref{transfo:gender-swap} & Token Replacement Based on Lookup Tables & \ref{transfo:token-replacement}\\
GeoNames Transformation & \ref{transfo:GeoNames-Transformation} & Transformer Fill & \ref{transfo:transformer-fill}\\
German Gender Swap & \ref{transfo:GermanGenderSwap} & Added Underscore Trick & \ref{transfo:underscore-trick}\\
Grapheme to Phoneme Substitution & \ref{transfo:grapheme-to-phoneme-transformation} & Unit converter & \ref{transfo:unit-converter}\\
Greetings and Farewells & \ref{transfo:Greetings-and-Farewells} & Urban Thesaurus Swap & \ref{transfo:urban-dict-swap}\\
Hashtagify & \ref{transfo:hashtagify} & Use Acronyms & \ref{transfo:use-acronyms}\\
Insert English and French Abbreviations & \ref{transfo:insert-abbreviation} & Visual Attack Letter & \ref{transfo:visual-attack-letters}\\
Leet Transformation & \ref{transfo:leet-letters} & Weekday Month Abbreviation & \ref{transfo:weekday-month-abbreviation}\\
Lexical Counterfactual Generator & \ref{transfo:lexical-counterfactual-generator} & Whitespace Perturbation & \ref{transfo:whitespace-perturbation}\\
Longer Location for NER & \ref{transfo:longer-location-ner} & Context Noise for QA & \ref{transfo:word-noise}\\
Longer Location Names for testing NER & \ref{transfo:longer-location-ner2} & Writing System Replacement & \ref{transfo:writing-system-replacement}\\
Longer Names for NER & \ref{transfo:longer-names-ner} & Yes-No Question Perturbation & \ref{transfo:yes-no-question}\\
Lost in Translation & \ref{transfo:lost-in-translation} & Yoda Transformation & \ref{transfo:yoda-transform}\\
Mixed Language Perturbation & \ref{transfo:mixed-language-perturbation} &  & \\
	\bottomrule
	\end{tabular}
	\vspace*{0.5cm}
	\caption{List of transformations and link to their detailed descriptions in Appendix}
	\label{tab:transformations-list}
\end{table*}

%% file: filters-table-full-list.tex
\begin{table*}[!thb]
	\centering\footnotesize
	\noindent\begin{tabular}{ll|ll}
	\textbf{Filter} & \textbf{App.} & \textbf{Filter} & \textbf{App.}\\
	\toprule
	Code-Mixing Filter & \ref{filter:code-mixing} & Polarity Filter & \ref{filter:polarity}\\
	Diacritics Filter & \ref{filter:diacritic} & Quantitative Question Filter & \ref{filter:quantitative-ques}\\
Encoding Filter & \ref{filter:encoding} & Question type filter & \ref{filter:question-filter}\\
Englishness Filter & \ref{filter:englishness} & Repetitions Filter & \ref{filter:repetitions}\\
Gender Bias Filter & \ref{filter:gender-bias} & Phonetic Match Filter & \ref{filter:soundex}\\
Group Inequity Filter & \ref{filter:group-inequity} & Special Casing Filter & \ref{filter:special-casing}\\
Keyword Filter & \ref{filter:keywords} & Speech-Tag Filter & \ref{filter:speech-tag}\\
Language Filter & \ref{filter:lang} & Token-Amount filter & \ref{filter:token-amount}\\
Length Filter & \ref{filter:length} & Toxicity Filter & \ref{filter:toxicity}\\
Named-entity-count Filter & \ref{filter:named-entity-count} & Universal Bias Filter & \ref{filter:universal-bias}\\
Numeric Filter & \ref{filter:numeric} & Yes/no question filter & \ref{filter:yesno-question} \\
Oscillatory Hallucinations Filter & \ref{filter:oscillatory-hallucination} \\
	\bottomrule
	\end{tabular}
	\vspace*{0.3cm}
	\caption{List of filters and link to their detailed descriptions in Appendix}
	\label{tab:filters-list}
\end{table*}

%% file: sections/tags-perturbations.tex
\subsection{Tags for the classification of perturbations}
\label{subsec:tags-perturbations}

 We defined a list of tags which are useful for an efficient navigation in the pool of existing perturbations and for understanding the performance characteristics of the contributed transformations and filters (see e.g. the robustness analysis presented in Section~\ref{subsec:results}). There are three main categories of tags: (i) General properties tags, (ii) Output properties tags, and (iii) Processing properties tags.
 
\begin{table*}[!thb]
	\centering\small
	\noindent\begin{tabular}{p{0.19\textwidth}p{0.35\textwidth}p{0.35\textwidth}}
	\textbf{Property} & \textbf{Definition} & \textbf{Tags}\\
	\toprule
	Augmented set type & Transformation or Filter (Subpopulation)? & Filter, Transformation, Multiple (specify), Unclear, N/A\\
	\midrule
	General purpose & What will the data be used for? Augmenting training data? Testing robustness? Finding and fixing biases? Etc. & Augmentation, Bias, Robustness, Other (specify), Multiple (specify), Unclear, N/A\\
	\midrule
	Task type & For which NLP task(s) will the perturbation be beneficial? & Quality estimation, Question answering, Question generation, RDF-to-text generation, Sentiment analysis, Table-to-text generation, Text classification, Text tagging, Text-to-text generation\\
	\midrule
	Language(s)  & To which language(s) is is the perturbation applied? & * \\
	\midrule
	Linguistic level & On which linguistic level does the perturbation operate? & Discourse, Semantic, Style, Lexical, Syntactic, Word-order, Morphological, Character, Other (specify), Multiple (specify), Unclear, N/A\\
	\bottomrule
	\end{tabular}
	\vspace*{0.2cm}
	\caption{Criteria and possible tags for \textbf{General Properties} of perturbations}
	\label{tab:criteria-tags-general-properties}
\end{table*}

 \textbf{General properties} tags are shown in Table~\ref{tab:criteria-tags-general-properties}, and cover the type of the augmentation, i.e. whether it is a transformation or a filter (\textit{Augmented set type}), its general purpose, i.e. whether it is intended for augmentation, robustness, etc. (\textit{General purpose}), for which NLP tasks the created data will be useful (\textit{Task type}), to which languages it has been applied (\textit{Language(s)}), and on which linguistic level of representation it operates, i.e. semantic, syntactic, lexical, etc. (\textit{Linguistic level}).
 
\begin{table*}[!thb]
	\centering\small
	\noindent\begin{tabular}{p{0.19\textwidth}p{0.35\textwidth}p{0.35\textwidth}}
	\textbf{Property} & \textbf{Definition} & \textbf{Tags}\\
	\toprule
	Output/input ratio & Does the transformation generate one single output for each input, or a few, or many? &  =1,  >1 (Low),  >1 (High), Multiple (specify), Unclear, N/A\\
	\midrule
	Input/output similarity & On which level are the input and output similar (if applicable)? & Aural, Meaning, Visual, Other (specify), Multiple (specify), Unclear, N/A\\
	\midrule
	Meaning preservation & If you compare the output with the input, how is the meaning affected by the transformation? & Always-preserved, Possibly-changed, Always-changed, Possibly-added, Always-added, Possibly-removed, Always-removed, Multiple (specify), Unclear, N/A\\
	\midrule
	Grammaticality preservation & If you compare the output with the input, how is the grammatical correctness affected by the transformation? & Always-preserved, Possibly-impaired, Always-impaired, Possibly-improved, Always-improved, Multiple (specify), Unclear, N/A\\
	\midrule
	Readability preservation & If you compare the output with the input, how is the easyness of read affected by the transformation? & Always-preserved, Possibly-impaired, Always-impaired, Possibly-improved, Always-improved, Multiple (specify), Unclear, N/A\\
	\midrule
	Naturalness preservation & If you compare the output with the input, how is the naturalness of the text affected by the transformation? & Always-preserved, Possibly-impaired, Always-impaired, Possibly-improved, Always-improved, Multiple (specify), Unclear, N/A\\
	\bottomrule
	\end{tabular}
	\vspace*{0.2cm}
	\caption{Criteria and possible tags for \textbf{Output Properties} of perturbations (applicable to tranformations only)}
	\vspace*{0.5cm}
	\label{tab:criteria-tags-output-properties}
\end{table*}

 \textbf{Output properties} tags, shown in Table~\ref{tab:criteria-tags-output-properties}, apply to transformations only; they provide indications about how the data was affected during the respective transformations. There are currently six properties in this category: one to capture the number of different outputs that a transformation can produce (\textit{Output/Input ratio}), one to capture in which aspect the input and the output are alike (\textit{Input/Ouptut similarity}), and four to capture intrinsic qualities of the produced text or structured data, namely how were the meaning, the grammaticality, the readability and the naturalness affected by the transformation (respectively \textit{Meaning preservation}, \textit{Grammaticality preservation}, \textit{Readability preservation} and \textit{Naturalness preservation}).  Note that apart from Output/Input ratio, the output properties tags need to be specified manually for each transformation/filter (see Section \ref{subsec:label-retrieval}), and are thus subject to the interpretation of the annotator.
 
\begin{table*}[!thb]
	\centering\small
	\noindent\begin{tabular}{p{0.19\textwidth}p{0.35\textwidth}p{0.35\textwidth}}
	\textbf{Property} & \textbf{Definition} & \textbf{Tags}\\
	\toprule
	Input data processing & What kind of NL processing is applied to the input? & Addition, Chunking, Paraphrasing, Parsing, PoS-Tagging, Removal, Segmentation, Simplification, Stemming, Substitution, Tokenisation, Translation, Other (specify), Multiple (specify), Unclear, N/A\\
	\midrule
	Implementation & Is the perturbation implemented as rule-based or model-based? & Model-based, Rule-based, Both, Unclear, N/A\\
	\midrule
	Algorithm type & What type of algorithm is used to implement the perturbation? & API-based, External-knowledge-based, LSTM-based, Transformer-Based, Other (specify), Multiple (specify), Unclear, N/A\\
	\midrule
	Precision/recall & To what extent does the perturbation generate what it intends to generate (precision)? To what extent does the perturbation return an output for any input (recall)? & High-precision-High-recall, High-precision-Low-recall, Low-precision-High-recall, Low-precision-Low-recall, Unclear, N/A\\
	\midrule
	GPU Required? & Is GPU needed to run the perturbation? & No, Yes, Unclear, N/A\\
	\midrule
	Computational complexity / Time & How would you assess the computational complexity of running the perturbation? Does it need a lot of time to run? & High, Medium, Low\\
	\bottomrule
	\end{tabular}
	\vspace*{0.2cm}
	\caption{Criteria and possible tags for \textbf{Processing Properties} of perturbations}
	\label{tab:criteria-tags-processing-properties}
\end{table*}

\textbf{Processing properties} tags, shown in Table~\ref{tab:criteria-tags-processing-properties}, capture information related to the type of processing applied on the input (\textit{Input data processing}), the type of algorithm used (\textit{Algorithm type}), how it is implemented (\textit{Implementation}), its estimated precision and recall (\textit{Precision/recall}) and computational complexity (\textit{Computational complexity / Time}), and whether an accelerator is required to apply the transformation/filter (\textit{GPU required?}).

\subsection{Tag retrieval and assignment}
\label{subsec:label-retrieval}
%Saad (retrieval from GitHub, manual completion by participants, all annotations stored in spreadsheet).

Transformation and filters are assigned tags for each of the properties listed in Tables~\ref{tab:criteria-tags-general-properties}-\ref{tab:criteria-tags-processing-properties}. There are two sources for the tags: (i) assigning them manually, and (ii) using existing metadata embedded in the respective source code implementations of each given transformation and filter. 
The in-code metadata (see e.g. the \textit{Keywords} field in Figure~\ref{fig:tr}) provides descriptions for each one identificable aspects such as the language(s) supported, the type of task that the transformation or filter is applicable for, and other characteristical keywords. The specification and type of this metadata was pre-defined as a requirement for all contributors to the NL-Augmenter project to enable identification of the type of transformation of filter being written by their respective author(s).   

This metadata was initally collected through the creation of an automated script which programatically iterated through each transformation and filter and gathered all stated metadata. The metadata was then mapped by the script into discrete property groups as defined in Tables~\ref{tab:criteria-tags-general-properties}-\ref{tab:criteria-tags-processing-properties}. All contributing authors were invited to review the initally collected metadata and, where possible, add additional data.
% The authors were additionally asked to provide the task performance scores for each of their respective transformations or filters, as detailed in Section~\ref{subsec:results}.

%% file: sections/robustness-tables-pre.tex
\label{subsec:results}

\begin{table*}[!htb]
	\centering\small
	\begin{tabular}{lc|ccc|ccc|ccc|ccc}
	& & \multicolumn{3}{c|}{SST-2 Roberta-base} & \multicolumn{3}{c|}{QQP BERT-base-unc.} & \multicolumn{3}{c|}{MNLI Roberta-large} & \multicolumn{3}{c}{IMDB Roberta-base}\\
	\textbf{Tag} & \textbf{\#$_{All}$} & \textbf{\#$_{Evl}$} & \textbf{R$_{T}$} & \textbf{Var$_{S}$} & \textbf{\#$_{Evl}$} & \textbf{R$_{T}$} & \textbf{Var$_{S}$} & \textbf{\#$_{Evl}$} & \textbf{R$_{T}$} & \textbf{Var$_{S}$} & \textbf{\#$_{Evl}$} & \textbf{R$_{T}$} & \textbf{Var$_{S}$}\\
	\toprule
	Augmentation & 34 & 20 & 0.63 & -13.25 & 20 & 0.75 & -6 & 18 & 0.74 & -8.89 & 17 & 0.73 & -4.41\\
	Bias & 3 & 1 & 0.5 & -5 & 2 & 0.52 & -11.5 & 2 & 0.53 & -16 & 1 & 0.71 & 0\\
	Robustness & 15 & 8 & 0.82 & -9.38 & 7 & 0.59 & -8.14 & 7 & 0.65 & -12.14 & 7 & 0.88 & -13.71\\
	Other* & 1 & 1 & 0.5 & -38 & 1 & 0.5 & -23 & 1 & 0.5 & -44 & 1 & 0.6 & 1\\
	Multiple* & 21 & 13 & 0.72 & -4.15 & 13 & 0.64 & -5.08 & 12 & 0.68 & -4.08 & 11 & 0.92 & -5.64\\
	\midrule
	Total & 74 & 43 & & & 43 & & & 40 & & & 37 & & \\
	\end{tabular}
	\vspace*{0.2cm}
	\caption{Results of the robustness evaluation from the perspective of the \textbf{General purpose} criterion (\#$_{All}$ = Total number of tags, \#$_{Evl}$ Total number of evaluations collected, R$_{T}$ = Transformation rate, Var$_{S}$ = Score variation)}
	\vspace*{0.35cm}
	\label{tab:Robust-General-purpose}
\end{table*}

\begin{table*}[!htb]
	\centering\small
	\begin{tabular}{lc|ccc|ccc|ccc|ccc}
	& & \multicolumn{3}{c|}{SST-2 Roberta-base} & \multicolumn{3}{c|}{QQP BERT-base-unc.} & \multicolumn{3}{c|}{MNLI Roberta-large} & \multicolumn{3}{c}{IMDB Roberta-base}\\
	\textbf{Tag} & \textbf{\#$_{All}$} & \textbf{\#$_{Evl}$} & \textbf{R$_{T}$} & \textbf{Var$_{S}$} & \textbf{\#$_{Evl}$} & \textbf{R$_{T}$} & \textbf{Var$_{S}$} & \textbf{\#$_{Evl}$} & \textbf{R$_{T}$} & \textbf{Var$_{S}$} & \textbf{\#$_{Evl}$} & \textbf{R$_{T}$} & \textbf{Var$_{S}$}\\
	\toprule
	Qual. estim. & 2 & 2 & 0.52 & -2.5 & 2 & 0.51 & -6 & 2 & 0.53 & -6.5 & 1 & 0.56 & 0\\
	Question ans. & 3 & 2 & 0.7 & -0.5 & 2 & 0.89 & -1.5 & 2 & 0.77 & -1 & 2 & 0.98 & -4\\
	Question gen. & 2 & 1 & 0.41 & 0 & 1 & 0.77 & -1 & 1 & 0.54 & -2 & 1 & 0.97 & -5\\
	RDF to text & 1 & 1 & 0.01 & 0 & 1 & 0.02 & 0 & 1 & 0.04 & 0 & 1 & 0.21 & 0\\
	Sentiment ana. & 4 & 1 & 0.99 & -12 & 1 & 0.99 & -14 & 1 & 0.93 & -18 & 1 & 1 & -15\\
	Table to text & 1 & 1 & 0.01 & 0 & 1 & 0.02 & 0 & 1 & 0.04 & 0 & 1 & 0.21 & 0\\
	Text class. & 95 & 52 & 0.71 & -9.27 & 52 & 0.68 & -6.21 & 49 & 0.69 & -8.33 & 43 & 0.83 & -5.74\\
	Text tagging & 25 & 17 & 0.79 & -10.94 & 17 & 0.64 & -6.82 & 16 & 0.66 & -9.75 & 13 & 0.84 & -9.23\\
	Text to text gen. & 92 & 49 & 0.69 & -8.86 & 49 & 0.66 & -5.86 & 46 & 0.68 & -7.57 & 40 & 0.79 & -5.62\\
	\midrule
	Total & 231 & 126 & & & 126 & & & 119 & & & 103 & & \\
	\end{tabular}
	\vspace*{0.35cm}
	\caption{Results of the robustness evaluation from the perspective of the \textbf{Task type} criterion (\#$_{All}$ = Total number of tags, \#$_{Evl}$ Total number of evaluations collected, R$_{T}$ = Transformation rate, Var$_{S}$ = Score variation)}
	\vspace*{0.35cm}
	\label{tab:Robust-Task-type}
\end{table*}

\begin{table*}[!htb]
	\centering\small
	\begin{tabular}{lc|ccc|ccc|ccc|ccc}
	& & \multicolumn{3}{c|}{SST-2 Roberta-base} & \multicolumn{3}{c|}{QQP BERT-base-unc.} & \multicolumn{3}{c|}{MNLI Roberta-large} & \multicolumn{3}{c}{IMDB Roberta-base}\\
	\textbf{Tag} & \textbf{\#$_{All}$} & \textbf{\#$_{Evl}$} & \textbf{R$_{T}$} & \textbf{Var$_{S}$} & \textbf{\#$_{Evl}$} & \textbf{R$_{T}$} & \textbf{Var$_{S}$} & \textbf{\#$_{Evl}$} & \textbf{R$_{T}$} & \textbf{Var$_{S}$} & \textbf{\#$_{Evl}$} & \textbf{R$_{T}$} & \textbf{Var$_{S}$}\\
	\toprule
	Semantic & 3 & 1 & 1 & -35 & 1 & 1 & -20 & 1 & 1.0 & -42 & 1 & 1 & -3\\
	Lexical & 44 & 30 & 0.67 & -5.83 & 30 & 0.61 & -5 & 30 & 0.64 & -4.4 & 25 & 0.73 & -2.44\\
	Syntactic & 3 & 1 & 1 & -8 & 1 & 0.74 & -7 & 1 & 0.85 & -15 & 1 & 1 & 0\\
	Word-order & 2 & 2 & 0.6 & -1.5 & 2 & 0.61 & -1 & 2 & 0.63 & -2 & 1 & 1 & 0\\
	Morphological & 3 & 2 & 0.75 & -25.5 & 2 & 0.75 & -21.5 & 2 & 0.75 & -28.5 & 2 & 0.8 & -4.5\\
	Character & 6 & 2 & 1 & -16.5 & 2 & 1.0 & -12.5 & 1 & 0.95 & -31 & 2 & 1 & -26\\
	Other* & 1 & 1 & 0 & 0 & 1 & 0.7 & -4 & 0 &  &  & 1 & 1 & -1\\
	Multiple* & 25 & 9 & 0.74 & -11.22 & 9 & 0.71 & -7 & 9 & 0.74 & -12.56 & 8 & 0.8 & -14.5\\
	Unclear & 1 & 1 & 1 & -46 & 1 & 0.79 & -2 & 0 &  &  & 0 &  & \\
	\midrule
	Total & 92 & 49 & & & 49 & & & 46 & & & 41 & & \\
	\end{tabular}
	\vspace*{0.2cm}
	\caption{Results of the robustness evaluation from the perspective of the \textbf{Linguistic level} criterion (\#$_{All}$ = Total number of tags, \#$_{Evl}$ Total number of evaluations collected, R$_{T}$ = Transformation rate, Var$_{S}$ = Score variation)}
	\vspace*{0.35cm}
	\label{tab:Robust-Linguistic-level}
\end{table*}

\begin{table*}[!htb]
	\centering\small
	\begin{tabular}{lc|ccc|ccc|ccc|ccc}
	& & \multicolumn{3}{c|}{SST-2 Roberta-base} & \multicolumn{3}{c|}{QQP BERT-base-unc.} & \multicolumn{3}{c|}{MNLI Roberta-large} & \multicolumn{3}{c}{IMDB Roberta-base}\\
	\textbf{Tag} & \textbf{\#$_{All}$} & \textbf{\#$_{Evl}$} & \textbf{R$_{T}$} & \textbf{Var$_{S}$} & \textbf{\#$_{Evl}$} & \textbf{R$_{T}$} & \textbf{Var$_{S}$} & \textbf{\#$_{Evl}$} & \textbf{R$_{T}$} & \textbf{Var$_{S}$} & \textbf{\#$_{Evl}$} & \textbf{R$_{T}$} & \textbf{Var$_{S}$}\\
	\toprule
	Aural & 5 & 3 & 1 & -4.33 & 3 & 0.7 & -6.67 & 2 & 0.7 & -6.5 & 3 & 0.85 & -3.67\\
	Meaning & 51 & 31 & 0.6 & -8.58 & 32 & 0.64 & -5.72 & 31 & 0.64 & -7.52 & 28 & 0.74 & -5.75\\
	Visual & 12 & 7 & 0.86 & -15.29 & 6 & 0.8 & -10.17 & 5 & 0.8 & -12.8 & 5 & 0.92 & -1\\
	Other* & 5 & 1 & 0.83 & 0 & 1 & 0.55 & -4 & 1 & 0.69 & -2 & 0 &  & \\
	Multiple* & 2 & 1 & 1 & -34 & 1 & 1 & -20 & 1 & 1.0 & -38 & 2 & 1 & -23\\
	N/A & 2 & 2 & 0.92 & -1 & 2 & 0.67 & -6 & 2 & 0.77 & -5 & 0 &  & \\
	\midrule
	Total & 77 & 45 & & & 45 & & & 42 & & & 38 & & \\
	\end{tabular}
	\vspace*{0.2cm}
	\caption{Results of the robustness evaluation from the perspective of the \textbf{Input/output similarity} criterion (\#$_{All}$ = Total number of tags, \#$_{Evl}$ Total number of evaluations collected, R$_{T}$ = Transformation rate, Var$_{S}$ = Score variation)}
	\vspace*{0.35cm}
	\label{tab:Robust-Input/output-similarity}
\end{table*}

%% file: sections/robustness.tex
\subsection{Results and Analysis}
\label{subsec:results}
In this section, Tables~\ref{tab:Robust-General-purpose} to~\ref{tab:Robust-Algorithm-type} show the results of the robustness analysis performed on the four datasets described in Section~\ref{subsec:robustness-analysis-experiment} and presented according to the tags introduced in Section~\ref{subsec:tags-perturbations}. 

% \begin{table*}[!htb]
% 	\centering\small
% 	\begin{tabular}{lc|ccc|ccc|ccc|ccc}
% 	& & \multicolumn{3}{c|}{SST-2 Roberta-base} & \multicolumn{3}{c|}{QQP BERT-base-unc.} & \multicolumn{3}{c|}{MNLI Roberta-large} & \multicolumn{3}{c}{IMDB Roberta-base}\\
% 	\textbf{Tag} & \textbf{\#$_{All}$} & \textbf{\#$_{Evl}$} & \textbf{R$_{T}$} & \textbf{Var$_{S}$} & \textbf{\#$_{Evl}$} & \textbf{R$_{T}$} & \textbf{Var$_{S}$} & \textbf{\#$_{Evl}$} & \textbf{R$_{T}$} & \textbf{Var$_{S}$} & \textbf{\#$_{Evl}$} & \textbf{R$_{T}$} & \textbf{Var$_{S}$}\\
% 	\toprule
% 	Augmentation & 34 & 20 & 0.63 & -13.25 & 20 & 0.75 & -6 & 18 & 0.74 & -8.89 & 17 & 0.73 & -4.41\\
% 	Bias & 3 & 1 & 0.5 & -5 & 2 & 0.52 & -11.5 & 2 & 0.53 & -16 & 1 & 0.71 & 0\\
% 	Robustness & 15 & 8 & 0.82 & -9.38 & 7 & 0.59 & -8.14 & 7 & 0.65 & -12.14 & 7 & 0.88 & -13.71\\
% 	Other* & 1 & 1 & 0.5 & -38 & 1 & 0.5 & -23 & 1 & 0.5 & -44 & 1 & 0.6 & 1\\
% 	Multiple* & 21 & 13 & 0.72 & -4.15 & 13 & 0.64 & -5.08 & 12 & 0.68 & -4.08 & 11 & 0.92 & -5.64\\
% 	\midrule
% 	Total & 74 & 43 & & & 43 & & & 40 & & & 37 & & \\
% 	\end{tabular}
% 	\vspace*{0.2cm}
% 	\caption{Results of the robustness evaluation from the perspective of the \textbf{General purpose} criterion (\#$_{All}$ = Total number of tags, \#$_{Evl}$ Total number of evaluations collected, R$_{T}$ = Transformation rate, Var$_{S}$ = Score variation)}
% 	\vspace*{0.35cm}
% 	\label{tab:Robust-General-purpose}
% \end{table*}

\textbf{General purpose (Table \ref{tab:Robust-General-purpose}):}
Transformations designed with a ``robustness testing'' objective displayed mean performance drops between 9\% and 13.7\% across models. Interestingly, 34 sentence transformations designed for ``augmentation'' tasks showed similar mean robustness drops ranging between 4\% and 13\%, emphasizing the need to draw on the paraphrasing literature to improve robustness testing.

\textbf{Task type (Table \ref{tab:Robust-Task-type}):} The results table shows that there is not necessarily a correlation between which task a transformation is marked to be relevant for and which task it actually challenges the robustness of the models on.

\textbf{Linguistic level (Table \ref{tab:Robust-Linguistic-level}):} Transformations making character level and morphological changes were able to show drastic levels of drops in performance compared to those making lexical or syntactic changes. These drops in performance were consistent across all four models. \modelname{roberta-large} finetuned on the MNLI dataset was the most brittle - character-level transformations on an average dropped performance by over 31\% and morphological changes dropped it by 28\% while those which made lexical changes displayed a mean drop of 4.4\%. The \testname{visual\_attack\_letters} (\ref{transfo:visual-attack-letters}) transformation, which replaces characters with similarly looking ones (like \textit{y} and \textit{v}), shows a large accuracy drop from 94\% to 56\% on the `roberta-base` model fine tuned on SST. `bert-base-uncased` fine-tuned on the QQP dataset drops from 92 to 69. \modelname{roberta-large-mnli} drops from 91 to 47. In the case of \testname{visual\_attack\_letters}, one can easily conceive a scenario in which a model is applied to OCR text which likely exhibit similar properties. In this case, one may expect similarly poor performance, arguably attributed to a narrow set of characters that the models have been exposed to.

% \begin{table*}[!htb]
% 	\centering\small
% 	\begin{tabular}{lc|ccc|ccc|ccc|ccc}
% 	& & \multicolumn{3}{c|}{SST-2 Roberta-base} & \multicolumn{3}{c|}{QQP BERT-base-unc.} & \multicolumn{3}{c|}{MNLI Roberta-large} & \multicolumn{3}{c}{IMDB Roberta-base}\\
% 	\textbf{Tag} & \textbf{\#$_{All}$} & \textbf{\#$_{Evl}$} & \textbf{R$_{T}$} & \textbf{Var$_{S}$} & \textbf{\#$_{Evl}$} & \textbf{R$_{T}$} & \textbf{Var$_{S}$} & \textbf{\#$_{Evl}$} & \textbf{R$_{T}$} & \textbf{Var$_{S}$} & \textbf{\#$_{Evl}$} & \textbf{R$_{T}$} & \textbf{Var$_{S}$}\\
% 	\toprule
% 	Aural & 5 & 3 & 1 & -4.33 & 3 & 0.7 & -6.67 & 2 & 0.7 & -6.5 & 3 & 0.85 & -3.67\\
% 	Meaning & 51 & 31 & 0.6 & -8.58 & 32 & 0.64 & -5.72 & 31 & 0.64 & -7.52 & 28 & 0.74 & -5.75\\
% 	Visual & 12 & 7 & 0.86 & -15.29 & 6 & 0.8 & -10.17 & 5 & 0.8 & -12.8 & 5 & 0.92 & -1\\
% 	Other* & 5 & 1 & 0.83 & 0 & 1 & 0.55 & -4 & 1 & 0.69 & -2 & 0 &  & \\
% 	Multiple* & 2 & 1 & 1 & -34 & 1 & 1 & -20 & 1 & 1.0 & -38 & 2 & 1 & -23\\
% 	N/A & 2 & 2 & 0.92 & -1 & 2 & 0.67 & -6 & 2 & 0.77 & -5 & 0 &  & \\
% 	\midrule
% 	Total & 77 & 45 & & & 45 & & & 42 & & & 38 & & \\
% 	\end{tabular}
% 	\vspace*{0.2cm}
% 	\caption{Results of the robustness evaluation from the perspective of the \textbf{Input/output similarity} criterion (\#$_{All}$ = Total number of tags, \#$_{Evl}$ Total number of evaluations collected, R$_{T}$ = Transformation rate, Var$_{S}$ = Score variation)}
% 	\vspace*{0.35cm}
% 	\label{tab:Robust-Input/output-similarity}
% \end{table*}

\begin{table*}[!htb]
	\centering\small
	\begin{tabular}{lc|ccc|ccc|ccc|ccc}
	& & \multicolumn{3}{c|}{SST-2 Roberta-base} & \multicolumn{3}{c|}{QQP BERT-base-unc.} & \multicolumn{3}{c|}{MNLI Roberta-large} & \multicolumn{3}{c}{IMDB Roberta-base}\\
	\textbf{Tag} & \textbf{\#$_{All}$} & \textbf{\#$_{Evl}$} & \textbf{R$_{T}$} & \textbf{Var$_{S}$} & \textbf{\#$_{Evl}$} & \textbf{R$_{T}$} & \textbf{Var$_{S}$} & \textbf{\#$_{Evl}$} & \textbf{R$_{T}$} & \textbf{Var$_{S}$} & \textbf{\#$_{Evl}$} & \textbf{R$_{T}$} & \textbf{Var$_{S}$}\\
	\toprule
	Alw. preserved & 40 & 22 & 0.65 & -9.77 & 22 & 0.63 & -7.36 & 22 & 0.61 & -11.23 & 19 & 0.72 & -9.89\\
	Poss. changed & 33 & 20 & 0.78 & -5.45 & 20 & 0.73 & -5.15 & 17 & 0.75 & -4.76 & 18 & 0.87 & -1.5\\
	Alw. changed & 12 & 5 & 0.7 & -4 & 5 & 0.54 & -5.4 & 5 & 0.61 & -6.8 & 3 & 0.78 & -7.33\\
	Alw. added & 2 & 1 & 0 & -94 & 1 & 0.7 & -4 & 1 & 0.78 & 0 & 1 & 0.99 & -1\\
	Poss. removed & 2 & 2 & 1 & -18 & 2 & 1 & -13 & 2 & 0.88 & -23.5 & 1 & 1 & -3\\
	\midrule
	Total & 89 & 50 & & & 50 & & & 47 & & & 42 & & \\
	\end{tabular}
 	\vspace*{0.2cm}
	\caption{Results of the robustness evaluation from the perspective of the \textbf{Meaning preservation} criterion (\#$_{All}$ = Total number of tags, \#$_{Evl}$ Total number of evaluations collected, R$_{T}$ = Transformation rate, Var$_{S}$ = Score variation)}
 	\vspace*{0.65cm}
	\label{tab:Robust-Meaning-preservation}
\end{table*}

\begin{table*}[!htb]
	\centering\small
	\begin{tabular}{lc|ccc|ccc|ccc|ccc}
	& & \multicolumn{3}{c|}{SST-2 Roberta-base} & \multicolumn{3}{c|}{QQP BERT-base-unc.} & \multicolumn{3}{c|}{MNLI Roberta-large} & \multicolumn{3}{c}{IMDB Roberta-base}\\
	\textbf{Tag} & \textbf{\#$_{All}$} & \textbf{\#$_{Evl}$} & \textbf{R$_{T}$} & \textbf{Var$_{S}$} & \textbf{\#$_{Evl}$} & \textbf{R$_{T}$} & \textbf{Var$_{S}$} & \textbf{\#$_{Evl}$} & \textbf{R$_{T}$} & \textbf{Var$_{S}$} & \textbf{\#$_{Evl}$} & \textbf{R$_{T}$} & \textbf{Var$_{S}$}\\
	\toprule
	Alw. preserved & 31 & 19 & 0.59 & -10.58 & 19 & 0.52 & -4.63 & 18 & 0.53 & -8.11 & 17 & 0.76 & -4.94\\
	Poss. impaired & 36 & 20 & 0.69 & -3.15 & 20 & 0.69 & -4.55 & 19 & 0.72 & -4.21 & 18 & 0.81 & -2.11\\
	Alw. impaired & 2 & 1 & 0.93 & -7 & 1 & 0.94 & -20 & 1 & 0.92 & -16 & 1 & 1 & -1\\
	Poss. improved & 6 & 6 & 0.83 & -16.33 & 6 & 0.8 & -8.17 & 5 & 0.79 & -14.8 & 2 & 0.52 & -1.5\\
	Unclear & 1 & 1 & 1 & -34 & 1 & 1 & -20 & 1 & 1.0 & -38 & 1 & 1 & -45\\
	N/A & 2 & 2 & 1 & -23.5 & 2 & 1 & -22 & 2 & 1 & -27 & 2 & 1 & -36.5\\
	\midrule
	Total & 79 & 49 & & & 49 & & & 46 & & & 41 & & \\
	\end{tabular}
 	\vspace*{0.2cm}
	\caption{Results of the robustness evaluation from the perspective of the \textbf{Grammaticality preservation} criterion (\#$_{All}$ = Total number of tags, \#$_{Evl}$ Total number of evaluations collected, R$_{T}$ = Transformation rate, Var$_{S}$ = Score variation)}
 	\vspace*{0.65cm}
	\label{tab:Robust-Grammaticality-preservation}
\end{table*}

\begin{table*}[!htb]
	\centering\small
	\begin{tabular}{lc|ccc|ccc|ccc|ccc}
	& & \multicolumn{3}{c|}{SST-2 Roberta-base} & \multicolumn{3}{c|}{QQP BERT-base-unc.} & \multicolumn{3}{c|}{MNLI Roberta-large} & \multicolumn{3}{c}{IMDB Roberta-base}\\
	\textbf{Tag} & \textbf{\#$_{All}$} & \textbf{\#$_{Evl}$} & \textbf{R$_{T}$} & \textbf{Var$_{S}$} & \textbf{\#$_{Evl}$} & \textbf{R$_{T}$} & \textbf{Var$_{S}$} & \textbf{\#$_{Evl}$} & \textbf{R$_{T}$} & \textbf{Var$_{S}$} & \textbf{\#$_{Evl}$} & \textbf{R$_{T}$} & \textbf{Var$_{S}$}\\
	\toprule
	Alw. preserved & 25 & 15 & 0.66 & -3 & 15 & 0.54 & -3.47 & 15 & 0.56 & -5.53 & 12 & 0.83 & -2.33\\
	Poss. impaired & 38 & 24 & 0.64 & -10.67 & 24 & 0.69 & -6.25 & 22 & 0.69 & -6.59 & 22 & 0.79 & -2.41\\
	Alw. impaired & 9 & 4 & 1 & -25.25 & 4 & 1.0 & -17.25 & 3 & 0.98 & -36.67 & 4 & 1 & -40\\
	Poss. improved & 4 & 4 & 0.75 & -11.75 & 4 & 0.75 & -8.75 & 4 & 0.75 & -16.25 & 2 & 0.52 & -1.5\\
	Alw. improved & 2 & 1 &  & -1 & 1 &  & -6 & 1 & 0.77 & -5 & 0 &  & \\
	Unclear & 1 & 1 & 1 & 0 & 1 & 0.06 & 0 & 1 & 0.15 & 0 & 1 & 0.32 & 0\\
	\midrule
	Total & 79 & 49 & & & 49 & & & 46 & & & 41 & & \\
	\end{tabular}
 	\vspace*{0.2cm}
	\caption{Results of the robustness evaluation from the perspective of the \textbf{Readability preservation} criterion (\#$_{All}$ = Total number of tags, \#$_{Evl}$ Total number of evaluations collected, R$_{T}$ = Transformation rate, Var$_{S}$ = Score variation)}
 	\vspace*{0.65cm}
	\label{tab:Robust-Readability-preservation}
\end{table*}

\begin{table*}[!htb]
	\centering\small
	\begin{tabular}{lc|ccc|ccc|ccc|ccc}
	& & \multicolumn{3}{c|}{SST-2 Roberta-base} & \multicolumn{3}{c|}{QQP BERT-base-unc.} & \multicolumn{3}{c|}{MNLI Roberta-large} & \multicolumn{3}{c}{IMDB Roberta-base}\\
	\textbf{Tag} & \textbf{\#$_{All}$} & \textbf{\#$_{Evl}$} & \textbf{R$_{T}$} & \textbf{Var$_{S}$} & \textbf{\#$_{Evl}$} & \textbf{R$_{T}$} & \textbf{Var$_{S}$} & \textbf{\#$_{Evl}$} & \textbf{R$_{T}$} & \textbf{Var$_{S}$} & \textbf{\#$_{Evl}$} & \textbf{R$_{T}$} & \textbf{Var$_{S}$}\\
	\toprule
	Alw. preserved & 18 & 9 & 0.59 & -3.33 & 10 & 0.52 & -3.5 & 9 & 0.51 & -7.44 & 9 & 0.75 & -2.56\\
	Poss. impaired & 45 & 29 & 0.66 & -8.48 & 29 & 0.64 & -5.38 & 27 & 0.67 & -5.15 & 24 & 0.79 & -1.75\\
	Alw. impaired & 8 & 4 & 1.0 & -20.5 & 4 & 1.0 & -16.25 & 4 & 0.97 & -23.25 & 4 & 1 & -32.25\\
	Poss. improved & 4 & 4 & 0.75 & -11.75 & 4 & 0.75 & -8.75 & 4 & 0.75 & -16.25 & 2 & 0.52 & -1.5\\
	Unclear & 1 & 1 & 1 & -34 & 1 & 1 & -20 & 1 & 1.0 & -38 & 1 & 1 & -45\\
	\midrule
	Total & 77 & 47 & & & 48 & & & 45 & & & 40 & & \\
	\end{tabular}
 	\vspace*{0.2cm}
	\caption{Results of the robustness evaluation from the perspective of the \textbf{Naturalness preservation} criterion (\#$_{All}$ = Total number of tags, \#$_{Evl}$ Total number of evaluations collected, R$_{T}$ = Transformation rate, Var$_{S}$ = Score variation)}
 	\vspace*{0.65cm}
	\label{tab:Robust-Naturalness-preservation}
\end{table*}

\begin{table*}[!htb]
	\centering\small
	\begin{tabular}{lc|ccc|ccc|ccc|ccc}
	& & \multicolumn{3}{c|}{SST-2 Roberta-base} & \multicolumn{3}{c|}{QQP BERT-base-unc.} & \multicolumn{3}{c|}{MNLI Roberta-large} & \multicolumn{3}{c}{IMDB Roberta-base}\\
	\textbf{Tag} & \textbf{\#$_{All}$} & \textbf{\#$_{Evl}$} & \textbf{R$_{T}$} & \textbf{Var$_{S}$} & \textbf{\#$_{Evl}$} & \textbf{R$_{T}$} & \textbf{Var$_{S}$} & \textbf{\#$_{Evl}$} & \textbf{R$_{T}$} & \textbf{Var$_{S}$} & \textbf{\#$_{Evl}$} & \textbf{R$_{T}$} & \textbf{Var$_{S}$}\\
	\toprule
	Addition & 1 & 1 & 0 & -94 & 1 & 0.7 & -4 & 1 & 0.78 & 0 & 1 & 0.99 & -1\\
	Paraphrasing & 5 & 5 & 0.79 & -1.8 & 5 & 0.74 & -5.6 & 4 & 0.77 & -6.25 & 3 & 0.77 & -0.67\\
	Parsing & 1 & 1 & 0.02 & 0 & 1 & 0.16 & -1 & 1 & 0.15 & 0 & 1 & 0.59 & 0\\
	PoS-Tagging & 5 & 3 & 0.44 & -11.67 & 3 & 0.54 & -6.67 & 3 & 0.54 & -14.33 & 2 & 0.98 & -1.5\\
	Removal & 2 & 2 & 1 & -4.5 & 2 & 0.74 & -6.5 & 2 & 0.81 & -10 & 1 & 1 & 0\\
	Segmentation & 3 & 1 & 1 & -4 & 1 & 0.93 & -6 & 1 & 0.94 & -5 & 1 & 1 & -4\\
	Substitution & 17 & 13 & 0.63 & -8.08 & 14 & 0.61 & -8 & 14 & 0.64 & -9.36 & 13 & 0.67 & -5\\
	Tokenisation & 23 & 9 & 0.67 & -4.89 & 9 & 0.5 & -4.22 & 9 & 0.54 & -4.56 & 10 & 0.76 & -3.8\\
	Translation & 3 & 2 & 0.99 & -11 & 2 & 0.99 & -13.5 & 2 & 0.97 & -18.5 & 1 & 1 & -15\\
	Other* & 3 & 2 & 1 & -17 & 2 & 1.0 & -10 & 1 & 0.95 & -38 & 2 & 1 & -23\\
	Multiple* & 13 & 6 & 0.69 & -1.33 & 5 & 0.6 & -2.2 & 5 & 0.58 & -4.8 & 3 & 0.72 & -2\\
	Unclear & 1 & 1 & 1 & -46 & 1 & 0.79 & -2 & 0 &  &  & 0 &  & \\
	N/A & 3 & 2 & 0.85 & -18.5 & 2 & 0.9 & -14 & 2 & 0.89 & -20.5 & 2 & 1 & -32\\
	\midrule
	Total & 81 & 48 & & & 48 & & & 45 & & & 40 & & \\
	\end{tabular}
	\caption{Results of the robustness evaluation from the perspective of the \textbf{Input data processing} criterion (\#$_{All}$ = Total number of tags, \#$_{Evl}$ Total number of evaluations collected, R$_{T}$ = Transformation rate, Var$_{S}$ = Score variation)}
	\label{tab:Robust-Input-data-processing}
\end{table*}

\begin{table*}[!htb]
	\centering\small
	\begin{tabular}{lc|ccc|ccc|ccc|ccc}
	& & \multicolumn{3}{c|}{SST-2 Roberta-base} & \multicolumn{3}{c|}{QQP BERT-base-unc.} & \multicolumn{3}{c|}{MNLI Roberta-large} & \multicolumn{3}{c}{IMDB Roberta-base}\\
	\textbf{Tag} & \textbf{\#$_{All}$} & \textbf{\#$_{Evl}$} & \textbf{R$_{T}$} & \textbf{Var$_{S}$} & \textbf{\#$_{Evl}$} & \textbf{R$_{T}$} & \textbf{Var$_{S}$} & \textbf{\#$_{Evl}$} & \textbf{R$_{T}$} & \textbf{Var$_{S}$} & \textbf{\#$_{Evl}$} & \textbf{R$_{T}$} & \textbf{Var$_{S}$}\\
	\toprule
	Model-based & 19 & 11 & 0.95 & -11.27 & 11 & 0.93 & -7.64 & 9 & 0.93 & -11.78 & 7 & 0.81 & -2.43\\
	Rule-based & 66 & 38 & 0.65 & -9.24 & 38 & 0.61 & -6.26 & 37 & 0.64 & -8.14 & 34 & 0.79 & -6.5\\
	Both & 6 & 2 & 0.31 & 0 & 2 & 0.5 & -0.5 & 2 & 0.42 & -1.5 & 1 & 0.97 & -5\\
	Unclear & 1 & 1 & 1 & -7 & 1 & 0.84 & -4 & 1 & 0.9 & -2 & 1 & 1 & -1\\
	\midrule
	Total & 103 & 52 & & & 52 & & & 49 & & & 43 & & \\
	\end{tabular}
	\caption{Results of the robustness evaluation from the perspective of the \textbf{Implementation} criterion (\#$_{All}$ = Total number of tags, \#$_{Evl}$ Total number of evaluations collected, R$_{T}$ = Transformation rate, Var$_{S}$ = Score variation)}
	\label{tab:Robust-Implementation}
\end{table*}

\begin{table*}[!htb]
	\centering\small
	\begin{tabular}{lc|ccc|ccc|ccc|ccc}
	& & \multicolumn{3}{c|}{SST-2 Roberta-base} & \multicolumn{3}{c|}{QQP BERT-base-unc.} & \multicolumn{3}{c|}{MNLI Roberta-large} & \multicolumn{3}{c}{IMDB Roberta-base}\\
	\textbf{Tag} & \textbf{\#$_{All}$} & \textbf{\#$_{Evl}$} & \textbf{R$_{T}$} & \textbf{Var$_{S}$} & \textbf{\#$_{Evl}$} & \textbf{R$_{T}$} & \textbf{Var$_{S}$} & \textbf{\#$_{Evl}$} & \textbf{R$_{T}$} & \textbf{Var$_{S}$} & \textbf{\#$_{Evl}$} & \textbf{R$_{T}$} & \textbf{Var$_{S}$}\\
	\toprule
	API-based & 22 & 14 & 0.78 & -7.86 & 14 & 0.67 & -7 & 13 & 0.73 & -9.23 & 11 & 0.88 & -11.45\\
	Ext. K.-based & 33 & 19 & 0.47 & -11 & 19 & 0.55 & -6.95 & 19 & 0.55 & -7.89 & 20 & 0.68 & -4.45\\
	LSTM-based & 1 & 1 & 1 & 0 & 1 & 1.0 & 0 & 0 & 0.9 &  & 1 & 1 & -1\\
	Transf.-based & 15 & 7 & 0.89 & -9.57 & 7 & 0.85 & -5.29 & 6 & 0.87 & -7.17 & 1 & 1 & -4\\
	Multiple* & 3 & 1 & 0.41 & 0 & 1 & 0.77 & -1 & 1 & 0.54 & -2 & 1 & 0.97 & -5\\
	Unclear & 1 & 0 &  &  & 0 &  &  & 0 &  &  & 1 &  & -1\\
	N/A & 24 & 4 & 1.0 & -13.25 & 4 & 0.77 & -8.5 & 4 & 0.75 & -18.75 & 3 & 0.89 & -6\\
	\midrule
	Total & 103 & 46 & & & 46 & & & 43 & & & 38 & & \\
	\end{tabular}
	\caption{Results of the robustness evaluation from the perspective of the \textbf{Algorithm type} criterion (\#$_{All}$ = Total number of tags, \#$_{Evl}$ Total number of evaluations collected, R$_{T}$ = Transformation rate, Var$_{S}$ = Score variation)}
	\label{tab:Robust-Algorithm-type}
\end{table*}

\textbf{Meaning preservation (Table \ref{tab:Robust-Meaning-preservation}):} 22 transformations which were marked as highly meaning preserving surprisingly showed a larger average performance drop as compared to 20 of those which were marked as possibly meaning changing. Not discounting the possibility of the noisiness of the transformation's logic, we believe further investigation could help understand whether models focus on the meaning of words or sentences or take shortcuts by focusing on commonly occuring surface forms associated with a particular prediction, as was already shown for some phenomena by \citet{mccoy-etal-2019-right}, among others.

\textbf{Grammaticality preservation (Table \ref{tab:Robust-Grammaticality-preservation}):} Preserving grammaticality did not correlate with high robustness. Transformations marked as grammaticality \testname{always-preserved} showed significant average drops of 10.6\%, 8.1\% and 4.6\% across \modelname{roberta-base-SST-2}, \modelname{roberta-large-mnli} and \modelname{bert-base-uncased-QQP} respectively. For example, the \testname{grapheme\_to\_phoneme} transformation showed drastic drops in performance: 13\%, 20\% and 13\% respectively. 

\textbf{Readability and Naturalness (Tables \ref{tab:Robust-Readability-preservation}-\ref{tab:Robust-Naturalness-preservation}):} In general, as expected, the transformations tagged as modifying the readability or naturalness show large drops across all tasks and models, in particular the ones tagged as ``always imparing'' the input.

Unsurprisingly, many of the injected perturbations, despite being artificial would not distract human readers from the actual meaning and intent of the text (e.g. \testname{simple\_ciphers} transformation (\ref{transfo:simple-ciphers})). Character level perturbations might not distract human readers as much as compared to word level perturbations but the above language models on the other hand behaved contrarily. Such departure from learning meaningful abstractions is further validated with the low correlation of grammaticality preservation and robustness. These results further re-question how we can expand these models from being just pure statistical learners to those which can incorporate meaning and surface-level abstraction, both across natural as well as artificial constructs. The large drops in performance of such perturbations necessitate looking at expanding training sets with even artificial data sources as well expand our definitions of text similarity from pure linguistic ones to those which abstract morphological, visual and other errors which can be unambiguous to humans.

Tables \ref{tab:Robust-Input/output-similarity}, \ref{tab:Robust-Input-data-processing}, \ref{tab:Robust-Implementation} and \ref{tab:Robust-Algorithm-type} show the robustness scores for \textbf{Input/Output similarity}, \textbf{Input processing}, \textbf{Implementation} and \textbf{Algorithm type} respectively. The score drops for these criteria may not be easily interpretable; e.g. that model-based implementations showed comparatively larger average drops as compared to rule-based implementations may not be due to the difference in implementation, but rather to which transformations were implemented that way .

%% file: sections/appendix-review-criteria.tex
\section{Review criteria for submission evaluation}
\label{sec:append-review}

\begin{figure*}[!thb]
    \centering
    \includegraphics[width=1\textwidth]{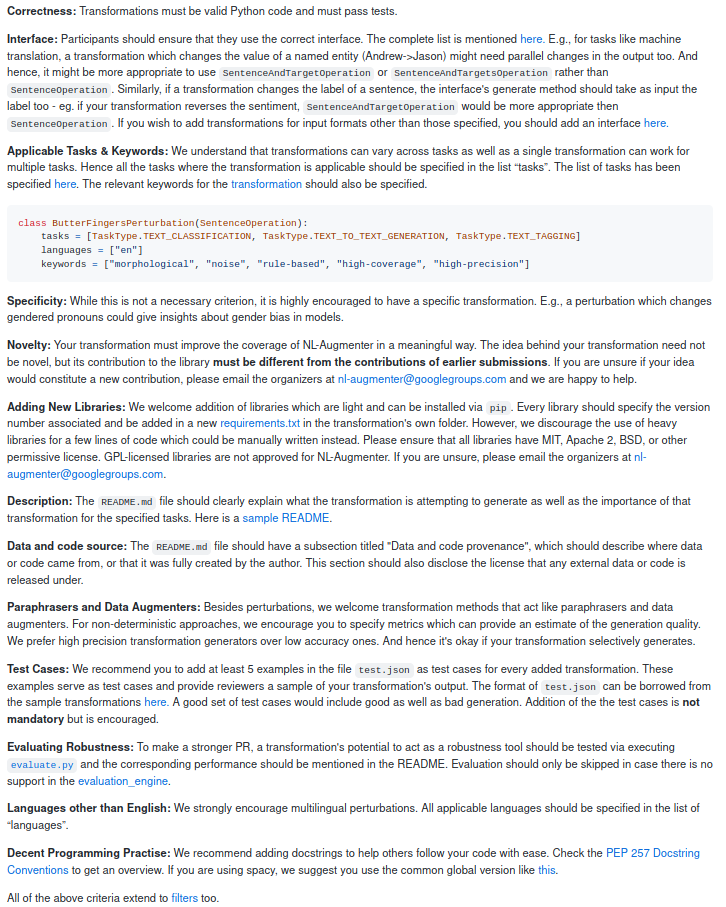}
    \caption{Participants and reviewers were provided with a set of review criteria.}
    \label{fig:rc}
\end{figure*}

Figure \ref{fig:rc} shows the detailed review criteria used for evaluating the transformation and filters submissions.

%% file: tf_list.tex
\section{Transformations}

The following is the list of all accepted transformations to NL-Augmenter project. Many of the transformations tokenize the sentences using SpaCy\footnote{\url{https://spacy.io/}} or NLTK~\cite{bird2006nltk} tokenizers. We discuss the implementations of each alongwith their limitations. The title of each transformation subsection is clickable and redirects to the actual python implementation. Many of the transformations use external libraries and we urge readers to look at each implementation and its corresponding `requirements.txt` files.

\transformation{abbreviation\_transformation}{Abbreviation Transformation}
\label{transfo:abbreviation-transformation}
This transformation replaces a word or phrase with its abbreviated counterpart ``homework'' -> ``hwk'' using a web-scraped slang dictionary.\footnote{Scraped from \url{https://www.noslang.com/dictionary}}

\texttt{\manicule \remove{You} \arrow\ \add{yu} driving at 80 \remove{miles per hour} \arrow\ \add{mph} is why insurance \remove{is} \arrow\ \add{tis} so \remove{freaking} \arrow\ \add{friggin} expensive.}

\transformation{add\_hashtags}{Add Hash-Tags}
\label{transfo:add-hashtags}
This transformation uses words in the text to generate hashtags. These hastags are then appeneded to the original text. Using the same words appearing in the sentence to generate the hashtags acts as redundant noise that models should learn to ignore. Hashtags are widespread in social media channels and are used to draw attention to the source text and also as a quick stylistic device.

\texttt{\manicule I love domino's pizza. \arrow\ \add{\#LovePizza \#Love \#I \#Pizza}}

\transformation{adjectives\_antonyms\_switch}{Adjectives Antonyms Switch}
\label{transfo:adjectives-antonyms-switch}
This transformation switches English adjectives in a sentence with their WordNet~\cite{miller1998wordnet} antonyms to generate new sentences with possibly different meanings and can be useful for tasks like Paraphrase Detection, Paraphrase Generation, Semantic Similarity, and Recognizing Textual Entailment.

\texttt{\manicule Amanda's mother was very \remove{beautiful} \arrow\ \add{ugly} .}

\transformation{americanize\_britishize\_english}{AmericanizeBritishizeEnglish}
\label{transfo:AmericanizeBritishizeEnglish}
This transformation takes a sentence and tries to convert it from British English to American English and vice-versa. A select set of words have been taken from~\citet{hyperreality}.

\texttt{\manicule I love the pastel \remove{colours} \arrow\ \add{colors}}

\transformation{antonyms\_substitute}{AntonymsSubstitute}
\label{transfo:AntonymsSubstitute}

This transformation introduces semantic diversity by replacing an even number of adjective/adverb antonyms in a given text. We assume that an even number of antonyms transforms will revert back sentence semantics; however, an odd number of transforms will revert the semantics. Thus, our transform only applies to the sentence that has an even number of revertible adjectives or adverbs. We called this mechanism double negation.

\texttt{\manicule Steve is \remove{able} \arrow\ \add{unable} to recommend movies that depicts the lives of \remove{beautiful} \arrow\ \add{ugly} minds.}

\transformation{auxiliary\_negation\_removal}{Auxiliary Negation Removal}
\label{transfo:auxiliary-negation-removal}
This is a low-coverage transformation which targets sentences that contain negations. It removes negations in English auxillaries and attempts to generate new sentences with the oppposite meaning.

\texttt{\manicule Ujjal Dev Dosanjh was \remove{not} \arrow\ Ujjal Dev Dosanjh was the 1st Premier of British Columbia from 1871 to 1872.}

\transformation{azerty\_qwerty\_chars\_swap}{AzertyQwertyCharsSwap}
\label{transfo:AzertyQwertyCharsSwap}

\texttt{\manicule Preferably use the above download link, as the release tarballs \remove{are generated deterministically} \arrow\ \add{qre generqted deterministicqlly} whereas GitHub's are not.}

\transformation{back\_translation}{BackTranslation}
\label{transfo:back-translation}
This transformation translates a given English sentence into German and back to English.This transformation acts like a light paraphraser. Multiple variations can be easily created via changing parameters like the language as well as the translation models which are available in plenty. Backtranslation has been quite popular now and has been a quick way to augment examples~\cite{Li_2019,sugiyama-yoshinaga-2019-data}.

\texttt{\manicule Andrew \remove{finally returned} \arrow\ \add{eventually gave} Chris the French book the French book I bought last week.}

\transformation{back\_translation\_ner}{BackTranslation for Named Entity Recognition}
\label{transfo:back-translation-ner}
This transformation splits the token sequences into segments of entity mention(s) and ``contexts'' around the entity mention(s). Backtranslation is used to paraphrase the contexts around the entity mention(s), thus resulting in a different surface form from the original token sequence. The resultant tokens are also assigned new tags. Exploiting this transformation has shown to empirically benefit named entity tagging~\cite{yaseen-and-langer-backtranslation-ner} and hence could arguably benefit other low-resource tagging tasks~\cite{bhatt2020benchmarking, Khachatrian2019BioRelEx1B, gupta2021candle}.

\transformation{butter\_fingers\_perturbation}{Butter Fingers Perturbation}
\label{transfo:butter-fingers-perturbation}
This perturbation adds noise to all types of text sources (sentence, paragraph, etc.) proportional to noise erupting from keyboard typos making common spelling errors. Few letters picked at random are replaced with letters which are at keyboard positions near the source letter. The implementation has been borrowed from here~\cite{ButterFingers} as used in~\cite{mille2021automatic}. There has also been some recent work in NoiseQA~\cite{ravichander2021noiseqa} to mimick keyboard typos.

\texttt{\manicule \remove{Sentences} \arrow\ \add{Senhences} with gapping, such as Paul likes \remove{coffee} \arrow\ \add{coffwe} and Mary tea, lack an overt predicate to \remove{indicate} \arrow\ \add{indicatx} the \remove{relation} \arrow\ \add{relauion} between two or more \remove{arguments} \arrow\ \add{argumentd} .}

\transformation{butter\_fingers\_perturbation\_for\_IL}{Butter Fingers Perturbation For Indian Languages}
\label{transfo:butter-fingers-perturbation-for-IL}
This implements the butter fingers perturbation as used above for 7 Indian languages: Bangla, Gujarati, Hindi, Kannada, Malayalam, Oriya, Punjabi, Tamil, and Telugu. The implementation considers the InScript keyboard~\footnote{\url{https://en.wikipedia.org/wiki/InScript\_keyboard}} which is decreed as a standard for Indian scripts. 

% \texttt{\manicule \remove{আপনি} \arrow\ \add{আপনন} কি বাংলা বলতে পারেন?

\transformation{change\_char\_case}{Change Character Case}
\label{transfo:change-char-case}
This transformation acts like a perturbation and randomly swaps the casing of some of the letters. The transformation's outputs will not work with uncased models or languages without casing. 

\texttt{\manicule Alice in Wonderland is a 2010 American live- \remove{action} \arrow\ \add{actIon} / \remove{animated} \arrow\ \add{anImated} dark \remove{fantasy} \arrow\ \add{faNtasy} adventure film.}

\transformation{change\_date\_format}{Change Date Format}
\label{transfo:change-date-format}
This transformation changes the format of dates.

\texttt{\manicule The first known case of COVID-19 was identified in Wuhan, China in \remove{December} \arrow\ \add{Dec} 2019.}

\transformation{change\_person\_named\_entities}{Change Person Named Entities}
\label{transfo:change-person-named-entities}
This perturbation changes the name of the person from one name to another by making use of the lexicon of person names in ~\citet{ribeiro2020beyond}. 

\texttt{\manicule \remove{Andrew} \arrow\ \add{Nathaniel} finally returned the French book to Chris that I bought last week}

\transformation{change\_two\_way\_ne}{Change Two Way Named Entities}
\label{transfo:change-two-way-ne}
This perturbation also changes the name of the person but also makes a parallel change in the label or reference text with the same name making it useful for text-to-text generation tasks. 

\texttt{\manicule He finally returned the French book to \remove{Chris} \arrow\ \add{Austin} that I bought last week}

\transformation{chinese\_antonym\_synonym\_substitution}{Chinese Antonym and Synonym Substitution}
\label{transfo:chinese-antonym-synonym-substitution}
%Chinese Antonym (反义词) and Synonym (同义词) Substitution
This transformation substitutes Chinese words with their synonyms or antonyms by using the Chinese dictionary\footnote{Chinese Dictionary: \url{https://github.com/guotong1988/chinese_dictionary}} and NLP Chinese Data Augmentation dictionary\footnote{NLP Chinese Data Augmentation: \url{https://github.com/425776024/nlpcda}}.

\transformation{chinese\_butter\_fingers\_perturbation}{Chinese Pinyin Butter Fingers Perturbation}
\label{transfo:chinese-butter-fingers-perturbation}
This transformation implements the Butter Fingers Perturbation for Chinese characters. Few Chinese words and characters that are picked at random will be substituted with others that have similar pinyin (based on the default Pinyin keyboards in Windows and Mac OS). It uses a database of 16142 Chinese characters~\footnote{\url{https://github.com/pwxcoo/chinese-xinhua}} and its associated pinyins to generate the perturbations for Chinese characters. A smaller database of 3500~\footnote{\url{https://github.com/elephantnose/characters}} more frequently seen Chinese characters are also used in the perturbations with a higher probability of being used compared to less frequently seen Chinese characters. It also uses a database of 575173 words~\footnote{\url{http://thuocl.thunlp.org/}} that are combined from several sources~\footnote{\url{https://github.com/fighting41love/Chinese\_from\_dongxiexidian}} in order to generate perturbations for Chinese words.

%\texttt{\manicule \remove{恰当的运用反义词，可以形成鲜明的对比，把事物的特点表达得更充分，给人留下深刻难忘的印象} \arrow\ \add{恰当的运用反义词，可以形成鲜明的对比，把石屋的特点表达得更充分，给人溜下深刻难忘的印象} 。

\transformation{chinese\_person\_named\_entities\_gender}{Chinese Person Named Entities and Gender Perturbation}
\label{transfo:chinese-person-named-entities-gender}
This perturbation adds noise to all types of text sources containing Chinese names (sentence, paragraph, etc.) by swapping a Chinese name with another Chinese name whilst also allowing the possibility of gender swap. CLUENER~\cite{xu2020cluener2020, zhao2019uer} is used for tagging named entities in Chinese. The list of names is taken from the Chinese Names Corpus!~\cite{Chinese-Names-Corpus}. It can provide assistance in detecting biases present in language models and the ability to infer implicit gender information when presented with gender-specific names. This can also be useful in mitigating representation biases in the input text. 

%\texttt{\manicule \remove{我的名字是曾祥莉，我的年龄是20岁。我在江西武宁，我的工作是前台} \arrow\ \add{我的名字是陈充全，我的年龄是20岁。我在江西武宁，我的工作是前台} 。

\transformation{chinese\_simplified\_traditional\_perturbation}{Chinese (Simplified \& Traditional) Perturbation}
\label{transfo:chinese-simplified-traditional-perturbation}
This perturbation adds noise to all types of text sources containing Chinese words and characters (sentence, paragraph, etc.) by changing the words and characters between Simplified and Traditional Chinese as well as other variants of Chinese Characters such as Japanese Kanji, character-level and phrase-level conversion, character variant conversion and regional idioms among Mainland China, Taiwan and Hong Kong, all available as configurations originally in the OpenChineseConvert project~\footnote{\url{https://github.com/BYVoid/OpenCC}}.

\transformation{city\_names\_transformation}{City Names Transformation}
\label{transfo:city-names-transformation}
This transformation replaces instances of populous and well-known cities in Spanish and English sentences with instances of less populous and less well-known cities to help reveal demographic biases~\cite{DBLP:journals/corr/abs-2008-03415} prevelant in named entity recognition models. The choice of cities have been taken from the World Cities Dataset~\footnote{\url{https://www.kaggle.com/juanmah/world-cities}}.

\texttt{\manicule The team was established in \remove{Dallas} \arrow\ \add{Viera West} in 1898 and was a charter member of the NFL in 1920.}

\transformation{close\_homophones\_swap}{Close Homophones Swap}
\label{transfo:close-homophones-swap}
Humans are generally guided by their senses and are unconsciously robust against phonetic attacks. Such types of attacks are highly popular in languages like English which has an irregular mapping between pronunciation and spelling~\cite{eger2020hero}.
This transformation mimics writing behaviors where users swap words with similar homophones either intentionally or by accident. This transformation acts like a perturbation to test robustness. Few words picked at random are replaced with words with similar homophones which sound similar or look similar. Some of the word choices might not be completely natural to normal human behavior, since humans "prefer" some words over others even they sound exactly the same. So it might not be fully reflecting the natural distribution of intentional or unintentional swapping of words.

\texttt{\manicule Sentences with gapping, such as Paul likes coffee and Mary \remove{tea} \arrow\ \add{Tee} , lack an overt predicate to indicate \remove{the} \arrow\ \add{Thee} relation between two or \remove{more} \arrow\ \add{Morr} arguments.}

\transformation{color\_transformation}{Color Transformation}
\label{transfo:color-transformation}
This transformation augments the input sentence by randomly replacing mentioned colors with different ones from the 147 extended color keywords specified by the World Wide Web Consortium (W3C)~\footnote{\url{https://www.w3.org/TR/2021/REC-css-color-3-20210805/}}. Some of the colors include ``dark sea green'', ``misty rose'', ``burly wood''.

\texttt{\manicule Tom bought 3 apples, 1 \remove{orange} \arrow\ \add{misty rose} , and 4 bananas and paid \$10.}

\transformation{concat}{Concatenate Two Random Sentences (Bilingual)}
\label{transfo:concat}
Given a dataset, this  transformation concatenates a sentence with a previously occuring sentence as explained in~\cite{nguyen2021data}. A monolingual version is mentioned in the subsequent subsection below. This concatenation would benefit all text tasks that use a transformer (and likely other sequence-to-sequence architectures). Previously published work~\cite{nguyen2021data} has shown a large gain in performance of low-resource machine translation using this method. In particular, the learned model is stronger due to being able to see training data that has context diversity, length diversity, and (to a lesser extent) position shifting.

\transformation{concat\_monolingual}{Concatenate Two Random Sentences (Monolingual)}
\label{transfo:concat-monolingual}
This is the monolingual counterpart of the above.

\texttt{\manicule I am just generating a very very very long sentence to make sure that the method is able to handle it. It does not even need to be a sentence. Right? This is not splitting on punctuation... I am just generating a very very very long sentence to make sure that the method is able to handle it. It does not even need to be a sentence. Right? This is not splitting on punctuation...}

\transformation{concept2sentence}{Concept2Sentence}
\label{transfo:concept2sentence}
This transformation intakes a sentence, its associated integer label, and (optionally) a dataset name that is supported by huggingface/datasets~\cite{lhoest2021datasets, quentin_lhoest_2021_5579268}. It works by extracting keyword concepts from the original sentence, passing them into a BART~\cite{lewis2020bart} transformer trained on CommonGen~\cite{lin2019commongen} to generate a new, related sentence which reflects the extracted concepts. Providing a dataset allows the function to use transformers-interpret~\cite{Pierse_Transformers_Interpret_2021} to identify the most critical concepts for use in the generative step. Underneath the hood, this transform makes use of the Sibyl tool~\cite{sibyl}, which is capable of also transforming the label as well. However, this particular implementation of C2S generates new text that is invariant (INV) with respect to the label. Since the model is trained on CommonGen, which is focussed on image captioning, the style of the output sentence would be geared towards scenic descritions and might not necessarily adhere to the syntax of the original sentence. Besides, it can be hard to argue that a handful subset of keywords could provide a complete description of the original sentence.

\transformation{contextual\_meaning\_perturbation}{Contextual Meaning Perturbation}
\label{transfo:contextual-meaning-perturbation}
This transformation was designed to model the "Chinese Whispers" or "Telephone" children's game: The transformed sentence appears fluent and somewhat logical, but the meaning of the original sentence might not be preserved. To achieve logical coherence, a pre-trained language model is used to replace words with alternatives that match the context of the sentence. Grammar mistakes are reduced by limiting the type of words considered for changes (based on POS tagging) and replacing adjectives with adjectives, nouns with nouns, etc. where possible.

This transformation benefits users who seek perturbations that preserve fluency but not the meaning of the sentence. For instance, it can be used in scenarios where the meaning is relevant to the task, but the model shows a tendency to over-rely on simpler features such as the grammatical correctness and general coherence of the sentence. A real-world example would be the training of quality estimation models for machine translation (does the translation maintain the meaning of the source?) or for text summarisation (does the summary capture the content of the source?).

Word substitution with pre-trained language models has been explored in different settings. For example, the augmentation library nlpaug~\cite{ma2019nlpaug} and the adversarial attack library TextAttack~\cite{morris2020textattack} include contextual perturbation methods. However, their implementations do not offer control over the type of words that should be perturbed and introduce a large number of grammar mistakes. If the aim is to change the sentence's meaning while preserving its fluency, this transformation can help to get the same effect with significantly fewer grammatical errors. ~\citet{li2020contextualized} propose an alternative approach to achieve a similar objective.

\transformation{contraction\_expansions}{Contractions and Expansions Perturbation}
\label{transfo:contraction-expansions}
This perturbation substitutes the text with popular expansions and contractions, e.g., ``I'm'' is changed to ``I am''and vice versa. The list of commonly used contractions \& expansions and the implementation of perturbation has been taken from Checklist~\cite{ribeiro2020beyond}.

\texttt{\manicule He often does \remove{n't} \arrow\ \add{not} come to school.}

\transformation{correct\_common\_misspellings}{Correct Common Misspellings}
\label{transfo:correct-common-misspellings}
This transformation acts like a lightweight spell-checker and corrects common misspellings appearing in text by looking for words in Wikipedia's Lists of Common Misspellings.

\texttt{\manicule Andrew \remove{andd} \arrow\ \add{and} Alice finally \remove{returnd} \arrow\ \add{returned} the French book that I bought \remove{lastr} \arrow\ \add{last} week}

\transformation{country\_state\_abbreviation\_transformation}{Country/State Abbreviation}
\label{transfo:country-state-abbreviation-transformation}
This transformation replaces country and state names with their common abbreviations\footnote{Countries States Cities Database: \url{https://github.com/dr5hn/countries-states-cities-database}}. Abbreviations can be common across different locations: \manicule ``MH'' can refer to Country Meath in Ireland as well as the state of Maharashtra in India and hence this transformation might result in a slight loss of information, especially if the surrounding context doesn't have enough signals.

\texttt{\manicule One health officer and one epidemiologist have boarded the ship in San Diego, \remove{CA} \arrow\ \add{California} on April 13, 2015 to conduct an environmental health assessment.}

\transformation{decontextualize\_sentence}{Decontextualisation of the main Event}
\label{transfo:decontextualize-sentence}
Semantic Role Labelling (SRL) is a powerful shallow semantic representation to determine who did what to whom, when, and where (and why and how etc). The core arguments generally talk about the participants involved in the event. Addtionally, contextual arguments on the other hand provide more specific information about the event. After tagging a sentence with an appropriate semantic role labels using an SRL labeller~\cite{jindal2020improved, shi2019simple}. This transformation crops out contextual arguments to create a new sentence with a minimal description of the event. Helping to generate textual pairs for entailment.

\transformation{diacritic\_removal}{Diacritic Removal}
\label{transfo:diacritic-removal}
``Diacritics are marks placed above or below (or sometimes next to) a letter in a word to indicate a particular pronunciation — in regard to accent, tone, or stress — as well as meaning, especially when a homograph exists without the marked letter or letters.'' \citet{mw:diacritic}. This transformation removes these diacritics or accented characters, and replaces them with their non-accented versions. It can be common for non-native or inexperienced speakers to miss out on any accents and specify non-accented versions. 

\texttt{\manicule She \remove{lookèd} \arrow\ \add{looked} east an she \remove{lookèd} \arrow\ \add{looked} west.}

\transformation{disability\_transformation}{Disability/Differently Abled Transformation}
\label{transfo:disability-transformation}
Disrespectful language can make people feel excluded and represent an obstacle towards their full participation in the society~\cite{RespectfulDisability}. This low-coverage transformation substitutes outdated references to references of disabilities with more appropriate and respectful ones which avoid negative connotations. A small list of inclusive words and phrases have been taken from a public article on  ~\href{https://www.gov.uk/government/publications/inclusive-communication/inclusive-language-words-to-use-and-avoid-when-writing-about-disability}{inclusive communication}, Wikipedia's list of ~\href{https://en.wikipedia.org/wiki/List_of_disability-related_terms_with_negative_connotations}{disability-related terms} with negative connotations, ~\href{https://ncdj.org/2015/09/terms-to-avoid-when-writing-about-disability/}{terms to avoid while writing about disability}.

\texttt{\manicule They are \remove{deaf} \arrow\ \add{person or people with a hearing} \add{disability}.}

\transformation{discourse\_marker\_substitution}{Discourse Marker Substitution}
\label{transfo:discourse-marker-substitution}
This perturbation replaces a discourse marker in a sentence by a semantically equivalent marker. Previous work has identified discourse markers that have low ambiguity~\cite{pitler-etal-2008-easily}. This transformation uses the corpus analysis on PDTB 2.0~\cite{prasad-etal-2008-penn} to identify discourse markers that are associated with a discourse relation with a chance of at least 0.5. Then, a marker is replaced with a different marker that is associated to the same semantic class.

\texttt{\manicule It has plunged 13\% \remove{since} \arrow\ \add{inasmuch as} July to around 26 cents a pound. A year ago ethylene sold for 33 cents}

\transformation{diverse\_paraphrase}{Diverse Paraphrase Generation Using SubModular Optimization and Diverse Beam Search}
\label{transfo:diverse-paraphrase}
This transformation generates multiple paraphrases of a sentence by employing 4 candidate selection methods on top of a base set of backtranslation models. 1) DiPS~\cite{dips2019} 2) Diverse Beam Search~\cite{AAAI1817329} 3) Beam Search~\cite{wiseman-rush-2016-sequence} 4) Random. Unlike beam search which generally focusses on the top-k candidates, DiPS introduces a novel formulation of using submodular optimisation to focus on generating more diverse paraphrases and has been proven to be an effective data augmenter for tasks like intent recognition and paraphrase detection~\cite{dips2019}. Diverse Beam Search attempts to generate diverse sequences by employing a diversity promoting alternative to the classical beam search~\cite{wiseman-rush-2016-sequence}.

\transformation{dyslexia\_words\_swap}{Dislexia Words Swap}
\label{transfo:dyslexia-words-swap}
This transformation acts like a perturbation by altering some words of the sentences with abberations~\cite{Dyslexia} that are likely to happen in the context of dyslexia.

\texttt{\manicule Biden hails \remove{your} \arrow\ \add{you're} relationship with Australia just days after new partnership drew ire from France.}

\transformation{emoji\_icon\_transformation}{Emoji Icon Transformation}
\label{transfo:emoji-icon-transformation}
\NewDocumentCommand\emojismile{}{
    \includegraphics[scale=0.08]{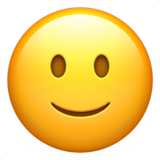}}

This transformation converts emojis into their equivalent keyboard format (e.g., \emojismile -> ":)" ) and vice versa (e.g., ":)" -> \emojismile ).

%\texttt{\manicule I am happy \remove{:)} \arrow\ \add{☺️} but you are sad \remove{:(} \arrow\ \add{☹️}}

\transformation{emojify}{Emojify}
\label{transfo:emojify}
\NewDocumentCommand\emojiapple{}{
    \includegraphics[scale=0.08]{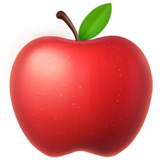}}
\NewDocumentCommand\emojieyes{}{
    \includegraphics[scale=0.08]{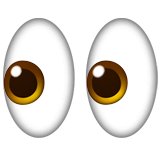}}
\NewDocumentCommand\emojishopping{}{
    \includegraphics[scale=0.08]{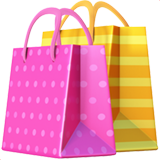}}
\NewDocumentCommand\emojiunitedkingdom{}{
    \includegraphics[scale=0.08]{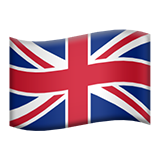}}
\NewDocumentCommand\emojikeycapone{}{
    \includegraphics[scale=0.08]{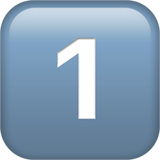}}
\NewDocumentCommand\emojikeycaptwo{}{
    \includegraphics[scale=0.08]{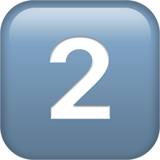}}
\NewDocumentCommand\emojikeycapthree{}{
    \includegraphics[scale=0.08]{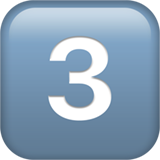}}

This transformation augments the input sentence by swapping words with emojis of similar meanings. Emojis, introduced in 1997 as a set of pictograms used in digital messaging, have become deeply integrated into our daily communication. More than 10\% of tweets\footnote{\url{https://blog.twitter.com/en\_us/a/2015/emoji-usage-in-tv-conversation}} and more than 35\% of Instagram posts\footnote{\url{https://instagram-engineering.com/}} include one or more emojis in 2015. Given the ubiquitousness of emojis, there is a growing body of work researching the linguistic and cultural aspects of emojis~\cite{guntuku2019studying} and how we can leverage the use of emojis to help solve NLP tasks~\cite{eisner-etal-2016-emoji2vec}. 

\texttt{\manicule Apple is looking at buying U.K. startup for \$132 billion. \arrow\ \emojiapple\ is \emojieyes\ at \emojishopping\ \emojiunitedkingdom\ startup for \$\emojikeycapone\emojikeycapthree\emojikeycaptwo.}

\transformation{english\_inflectional\_variation}{English Inflectional Variation}
\label{transfo:english-inflectional-variation}
This transformation adds inflectional variation to English words and can be used to test the robustness of models against inflectional variations. In English, each inflection generally maps to a Part-Of-Speech tag~\footnote{\href{https://www.ling.upenn.edu/courses/Fall_2003/ling001/penn_treebank_pos.html}{Penn TreeBank POS}} in the Penn Treebank~\cite{marcus-etal-1993-building}. For each content word in the sentence, it is first lemmatised before randomly sampling a valid POS category and reinflecting the word according to the new category. The sampling process for each word is constrained using its POS tag to maintain the original sense for polysemous words. This has been adapted from the Morpheus~\cite{tan-etal-2020-morphin} adversarial attack.

\texttt{\manicule Ujjal Dev Dosanjh \remove{served} \arrow\ \add{serve} as 33rd \remove{Premier} \arrow\ \add{Premiers} of British Columbia from 2000 to 2001}

\transformation{entity\_mention\_replacement\_ner}{English Mention Replacement for NER}
\label{transfo:entity-mention-replacement-ner}
This transformation randomly swaps an entity mention with another entity mention of the same entity type.
Exploiting this transformation as a data augmentation strategy has been empirically shown to improve the performance of underlying (NER) models~\cite{dai-adel-2020-analysis}.

\transformation{filler\_word\_augmentation}{Filler Word Augmentation}
\label{transfo:filler-word-augmentation}
This augmentation adds noise in the form of colloquial filler phrases. 23 different phrases are chosen across 3 different categories: general filler words and phrases ("uhm", "err", "actually", "like", "you know"...), phrases emphasizing speaker opinion/mental state ("I think/believe/mean", "I would say"...) \& phrases indicating uncertainty ("maybe", "perhaps", "probably", "possibly", "most likely").The latter two categories had shown promising results~\citet{kovatchev-etal-2021-vectors} when they were concatenated at the beginning of the sentence unlike this implementation which perform insertions at any random positions. Filler words are based on the work of~\citet{laserna2014like} but have not been explored in the context of data augmentation.

\transformation{formality\_change}{Style Transfer from Informal to Formal}
\label{transfo:formality-change}
This transformation transfers the style of text from formal to informal and vice versa. It uses the implementation of Styleformer~\cite{Styleformer}.

\texttt{\manicule What you \remove{upto} \arrow\ \add{currently doing} ?}

\transformation{french\_conjugation\_transformation}{French Conjugation Substitution}
\label{transfo:french-conjugation-transform}
This transformation change the conjugation of verbs for simple french sentences with a specified tense. It detects the pronouns used in the sentence in order to conjugate accordingly whenever a sentence contains differents verbs. This version only works for  indicative tenses. It also only works for simple direct sentences (subject, verb, COD/COI), which contains a pronoun as subject (il, elle, je etc.). It does not detect when the subject is a couple of nouns ("les enfants" or "la jeune femme").

\transformation{gender\_culture\_diverse\_name}{Gender And Culture Diversity Name Changer (1-way and 2-way)}
\label{transfo:gender-culture-diverse-name}

Corpora exhibits many representational biases and this transformation focuses on one particular mediator, the personal names. It diversifies names in the corpora along two critical dimensions, gender and cultural background. Technically, the transformation samples a (country, gender) pair and then randomly draws a name from that (country, gender) pair to replace the original name. We collected 42812 distinct names from 141 countries.They are primarily from the World Gender Name Dictionary \cite{DVN/MSEGSJ_2021}.

Common name augmentations do not consider their gender and cultural implication. Thus, they do not necessarily mitigate biases or promote the minority's representation because the augmented name may be from the same gender and cultural background. This is the case, for example in the CheckList's \cite{ribeiro2020beyond} implemented name augmentation. Taking the interaction of the names therein with ours, 34.0\%, 33.5\%, 31.9\%, 30.8\% of them are popular names in US, Canada, Australia, and UK, respectively. Only 0.4\%, 0.4\%, 0.5\%, 2.1\% of them are from India, Korea, China, and Kazakhstan.

\texttt{\manicule \remove{Rachel} \arrow\ \add{Charity} Green, a sheltered but friendly woman, flees her wedding day and wealthy yet unfulfilling life.}

%The ~\href{https://github.com/GEM-benchmark/NL-Augmenter/tree/main/transformations/gender\_culture\_diverse\_name\_two\_way}{2-way} implementation makes a parallel substitution for the reference or target sentence too so that it could benefit text-text generation tasks.

\transformation{gender\_neopronouns}{Neopronoun Substitution}
\label{transfo:gender-neopronouns}
This transformation performs grammatically correct substitution from English to English of the gendered pronouns, he/she, in a given sentence with their neopronoun counterparts, based on a list compiled by UNC Greensboro and LGBTA WIKI\footnote{\url{https://intercultural.uncg.edu/wp-content/uploads/Neopronouns-Explained-UNCG-Intercultural-Engagement.pdf}}. NLP models, such as those for neural machine translation, often fail to recognize the neopronouns and treat them as proper nouns. This transformation seeks to render the training data used in NLP pipelines more neopronoun aware to reduce the risk of trans-erasure. The reason why a simple look-up-table approach might not work is due to the fact that the case may differ depending on the context.

\texttt{\manicule \remove{She} \arrow\ \add{They} had \remove{her} \arrow\ \add{their} friends tell \remove{her} \arrow\ \add{them} about the event.}

\transformation{gender\_neutral\_rewrite}{Gender Neutral Rewrite}
\label{transfo:gender-neutral-rewrite}
This transformation involves rewriting an English sentence containing a single gendered entity  with its gender-neutral variant. One application is machine translation, when translating from a language with gender-neutral pronouns (e.g. Turkish) to a language with gendered pronouns (e.g. English). This transformation is based on the algorithm proposed by \citet{genderrewrite}.

\texttt{\manicule \remove{His} \arrow\ \add{Their} dream is to be a \remove{fireman} \arrow\ \add{firefighter} when \remove{he} \arrow\ \add{they} \remove{grows} \arrow\ \add{grow} up.}
%\texttt{\manicule This is \remove{his} \arrow\ \add{their} pen.}

\transformation{gender\_swap}{GenderSwapper}
\label{transfo:gender-swap}
This transformation introduces gender diversity to the given data. If used as data augmentation for training, the transformation might mitigate gender bias, as shown in~\citet{dinan-etal-2020-queens}. It also might be used to create a gender-balanced evaluation dataset to expose the gender bias of pre-trained models. This transformation performs lexical substitution of the opposite gender. The list of gender pairs (shepherd <--> shepherdess) is taken from~\citet{lu2019gender}. Genderwise names used from~\citet{ribeiro2020beyond} are also randomly swapped.

\transformation{geo\_names\_transformation}{GeoNames Transformation}
\label{transfo:GeoNames-Transformation}
This transformation augments the input sentence with information based on location entities (specifically cities and countries) available in the Geo\-Names database\footnote{\url{http://download.geonames.org/export/dump/}}. E.g., if a country name is found, the name of the country is appended with information about the country like its capital city, its neighbouring countries, its continent, etc. Some initial ideas of this nature were explored in~\citet{paisthesis}.

\transformation{german\_gender\_swap}{German Gender Swap}
\label{transfo:GermanGenderSwap}
This transformation replaces the masculine nouns and pronouns with their female counterparts for German sentences from a total of 2226 common German names.\footnote{\url{https://de.wiktionary.org/wiki/Verzeichnis:Deutsch/Namen}}

\texttt{\manicule \remove{Er} \arrow\ \add{Sie} ist \remove{ein Arzt} \arrow\ \add{eine Ärztin} und \remove{mein Vater} \arrow\ \add{meine Mutter} .}

\transformation{grapheme\_to\_phoneme\_transformation}{Grapheme to Phoneme Substitution}
\label{transfo:grapheme-to-phoneme-transformation}
This transformation adds noise to a sentence by randomly converting words to their phonemes. Grapheme-to-phoneme substitution is useful in NLP systems operating on speech. An example of grapheme to phoneme substitution is ``permit'' $\rightarrow$ \texttt{P ER0 M IH1 T'}.

\transformation{greetings\_and\_farewells}{Greetings and Farewells}
\label{transfo:Greetings-and-Farewells}
This transformation replaces greetings (e.g. "Hi", "Howdy") and farewells (e.g. "See you", "Good night") with their synonymous equivalents.

\texttt{\manicule \remove{Hey} \arrow\ \add{Hi} everyone. \remove{It's nice} \arrow\ \add{Pleased} to meet you. How \remove{have} \arrow\ \add{are} you \remove{been} ?}

\transformation{hashtagify}{Hashtagify}
\label{transfo:hashtagify}
This transformation modifies an input sentence by identifying named entities and other common words and turning them into hashtags, as often used in social media.

\transformation{insert\_abbreviation}{Insert English and French Abbreviations}
\label{transfo:insert-abbreviation}
This perturbation replaces in texts some well known English and French words or expressions with (one of) their abbreviations. Many of the abbreviations covered here are quite common on social medias platforms, even though some of them are quite generic. This implementation is partly inspired by recent work in Machine Translation~\cite{berard2019naver}.

\transformation{leet\_letters}{Leet Transformation}
\label{transfo:leet-letters}
Visual perturbations are often used to disguise offensive comments on social media (e.g., “!d10t”) or as a distinct writing style (“1337” in “leet speak”)~\cite{eger-etal-2019-text}, especially common in scenarios like video gaming. Humans are unconsciously robust to such visually similar texts. This perturbation replaces letters with their visually similar ``leet'' counterparts.\footnote{\url{https://simple.wikipedia.org/wiki/Leet}}

\texttt{\manicule \remove{Ujjal Dev Dosanjh served} \arrow\ \add{U7jal 0ev D0san74 serv3d} as 33rd \remove{Premier of British Columbia from} \arrow\ \add{Pr33i3r 0f 8ritis4 00lu36ia fr0m} 2000 \remove{to} \arrow\ \add{t0} 2001}

\transformation{lexical\_counterfactual\_generator}{Lexical Counterfactual Generator}
\label{transfo:lexical-counterfactual-generator}
This transformation generates counterfactuals by simply substituting negative words like ``not'', ``neither'' in one sentence of a semantically similar sentence pair. The substituted sentence is then backtranslated in an attempt to correct for grammaticality. This transformation would be useful for tasks like entailment and paraphrase detection.

\transformation{longer\_location\_ner}{Longer Location for NER}
\label{transfo:longer-location-ner}
This transformation augments data for Named Entity Recognition (NER) tasks by augmenting examples which have a Location Tag. Names of locations are expanded by appending them with cardinal directions like ``south'', ``N'', ``northwest'', etc. The transformation ensures that the tags of the new sentence are accordingly modified.

\transformation{longer\_location\_ner}{Longer Location Names for testing NER}
\label{transfo:longer-location-ner2}
This transformation augments data for Named Entity Recognition (NER) tasks by augmenting examples that have a Location (LOC) Tag. Names of location are expanded by inserting random prefix or postfix word(s). The transformation also ensures that the labels of the new tags are accordingly modified.

\transformation{longer\_names\_ner}{Longer Names for NER}
\label{transfo:longer-names-ner}
This transformation augments data for Named Entity Recognition (NER) tasks by augmenting examples which have a Person Tag. Names of people are expanded by inserting random characters as initials. The transformation also ensures that the labels of the new tags are accordingly modified. 

\transformation{lost\_in\_translation}{Lost in Translation}
\label{transfo:lost-in-translation}
This transformation is a generalization of the BackTranslation transformation to any sequence of languages supported by the Helsinki-NLP OpusMT models \citep{TiedemannThottingal:EAMT2020}. 

\texttt{\manicule Andrew \remove{finally returned} \arrow\ \add{brought Chris back} the French book the French book I bought last week I bought last week}

\transformation{mixed\_language\_perturbation}{Mixed Language Perturbation}
\label{transfo:mixed-language-perturbation}
Mixed language training has been effective for cross-lingual tasks~\cite{liu2020attention}, to help generate data for low-resource scenarios~\cite{liu2021continual} and for multilingual translation~\cite{fan2021beyond}. Two transformations translate randomly picked words in the text from English to other languages (e.g., German). It can be used to test the robustness of a model in a multilingual setting. 

\texttt{\manicule Andrew finally returned \remove{the} \arrow\ \add{die} Comic book to Chris that I bought last \remove{week} \arrow\ \add{woche}}

\transformation{mix\_transliteration}{Mix transliteration}
\label{transfo:mix-transliteration}
This transformation transliterates randomly picked words from the input sentence (of given source languae script) to a target language script. It can be used to train/test multilingual models to improve/evaluate their ability to understand complete or partially transliterated text.

\transformation{mr\_value\_replacement}{MR Value Replacement}
\label{transfo:mr-value-replacement}
This perturbation adds noise to a key-value meaning representation (MR) (and its corresponding sentence) by randomly substituting values/words with their synonyms (or related words).
This transformation uses a simple strategy to align values of a MR and tokens in the corresponding sentence inspired by how synonyms are substituted for tasks like machine translation~\cite{fadaee-etal-2017-data}. This way, there could be some problems in complex sentences. Besides, the transformation might introduce non-grammatical segments.

\transformation{multilingual\_back\_translation}{Multilingual Back Translation}
\label{transfo:multilingual-back-translation}
This transformation translates a given sentence from a 
given language into a pivot language and then back to the original language. This transformation is a simple paraphraser that works on 100 different languages. Back Translation has been quite popular now and has been a quick way to augment ~\cite{Li_2019,sugiyama-yoshinaga-2019-data, Fan2020BeyondEM}.

\texttt{\manicule \remove{Being honest} \arrow\ \add{Honesty} should be one of our most important \remove{character traits} \arrow\ \add{characteristics} }

\transformation{multilingual\_dictionary\_based\_code\_switch}{Multilingual Dictionary Based Code Switch}
\label{transfo:multilingual-dictionary-based-code-switch}
This transformation generates multi-lingual code-switching data to fine-tune encoders of large language models~\cite{ijcai2020-533, tan-joty-2021-code-mixing, wang2019cross} by making use of bilingual dictionaries of MUSE~\cite{lample2018word}. 

\transformation{multilingual\_lexicon\_perturbation}{Multilingual Lexicon Perturbation}
\label{transfo:multilingual-lexicon-perturbation}
This perturbation helps to creates code-mixed sentences for both high-resource and low-resource languages by randomly translating words with a specified probability from any supported languages (e.g., English) to other supported languages (e.g., Chinese) by using a multilingual lexicon. Thus, it can be used to generate code-mixed training data to improve models for multilingual and cross-lingual settings. As of now 100 languages are supported and 3000 common English words listed on ef.com~\footnote{\url{https://www.ef.com/wwen/english-resources/english-vocabulary/top-3000-words/}} are supported. The lexicon implementation is also 160x faster than its model based counterpart.

%\texttt{\manicule Ok setting \remove{your medicine} \arrow\ \add{你的 医学} appointment \remove{for} \arrow\ \add{为} 7pm}

\transformation{negate\_strengthen}{Causal Negation \& Strengthening}
\label{transfo:negate-strengthen}
This transformation is targeted at augmenting Causal Relations in text and adapts the code from the paper Causal Augmentation for Causal Sentence Classification~\cite{tan-etal-2021-causal}. There are two operations:
1. Causal Negation: Negative words like "not, no, did not" are introduced into sentences to unlink the causal relation. 
2. Causal Strengthening: Causal meaning is strengthened by converting weaker modal words into stronger ones like "may" to "will" to assert causal strength.

The implementation provides users with the option to amend causal meaning automatically from the root word of the sentence, or by explicitly highlighting the index of the word they wish to amend. Additionally, we include WordNet~\cite{miller1998wordnet} synonyms and tense matching to allow for more natural augmentations.

\texttt{\manicule The rs7044343 polymorphism \remove{could be} \arrow\ \add{was} involved in regulating the production of IL-33.}

\transformation{neural\_question\_paraphraser}{Question Rephrasing transformation}
\label{transfo:neural-question-paraphraser}
This implementation rephrases questions for sentence tasks by using the T5 model used in~\ref{transfo:protaugment-diverse-paraphrase} for Question Answering tasks.

\transformation{noun\_compound\_paraphraser}{English Noun Compound Paraphraser [N+N]}
\label{transfo:noun-compound-paraphraser}
This transformation replaces two-word noun compounds with a paraphrase, based on the compound paraphrase dataset from SemEval 2013 Task 4~\cite{hendrickx2013semeval}. It currently only works for English. Any two-word compound that appears in a dataset of noun compound paraphrases will be replaced by a paraphrase. If more than one two-word compound appears, then all combinations of compound paraphrases (including no paraphrase at all) will be returned. For example, the paraphrases of ``club house'' include ``house for club activities'', ``house for club members'', ``house in which a club meets'', etc. We start with replacing paraphrases with the highest score (the specified frequency in the annotated dataset), and paraphrases with the same score (ties) are sorted randomly.
This transformation currently only checks for noun compounds from~\citet{hendrickx2013semeval} and therefore has low coverage. To improve it, other datasets could be added, e.g., from~\citet{ponkiya2018treat} or~\citet{lauer1995designing}. To attain even wider-coverage (at the expense of lower precision), machine learning approaches such as~\citet{shwartz2018paraphrase} or~\citet{ponkiya2020looking} could be considered. In addition, some of the the paraphrases in~\citet{hendrickx2013semeval} sound a little odd (e.g., "blood cell" -> "cell of blood") and may not fit well in context.

\transformation{number-to-word}{Number to Word}
\label{transfo:number-to-word}
This transformation acts like a perturbation to improve robustness on processing numerical values. The perturbated sentence contains the same information as the initial sentence but with a different representation of numbers.

\transformation{numeric\_to\_word}{Numeric to Word}
\label{transfo:numeric-to-word}
This transformation translates numbers in numeric form to their textual representations. This includes general numbers, long numbers, basic math characters, currency, date, time, phone numbers, etc.

\transformation{ocr\_perturbation}{OCR Perturbation}
\label{transfo:ocr-perturbation}
This transformation directly induces Optical Character Recognition (OCR) errors into the input text. It renders the input sentence as an image and recognizes the rendered text using the OCR engine Tesseract 4~\citep{smith2007overview}. It works with text in English, French, Spanish, and German. The implementation follows previous work by \citet{namysl-etal-2021-empirical}.

\transformation{p1\_noun\_transformation}{Add Noun Definition}
\label{transfo:p1-noun-transformation}
This transformation appends noun definitions onto the original nouns in a sentence. Definitions of nouns are collected from Wikidata~\footnote{\url{https://www.wikidata.org/wiki/Wikidata:Main\_Page}}.

\transformation{pig\_latin}{Pig Latin Cipher}
\label{transfo:pig-latin}

This transformation translates the original text into pig latin. Pig Latin is a well-known deterministic transformation of English words, and can be viewed as a cipher which can be deciphered by a human with relative ease. The resulting sentences are completely unlike examples typically used in language model training. As such, this augmentation change the input into inputs which are difficult for a language model to interpret, while being relatively easy for a human to interpret.

\transformation{pinyin}{Pinyin Chinese Character Transcription}
\label{transfo:pinyin}
This transformation transcribes Chinese characters into their Mandarin pronunciation using the Pinyin romanization scheme. The Character-to-Pinyin converter at the core of this transformation is a neural model by \citet{park2020neural}.

\transformation{propbank\_srl\_roles}{SRL Argument Exchange}
\label{transfo:propbank-srl-roles}
This perturbation adds noise to all types of English text sources (sentence, paragraph, etc.) proportional to the number of arguments identified by SRL BERT~\citep{DBLP:journals/corr/abs-1904-05255}. Different rules are applied to deterministically modify the sentence in a meaning-preserving manner. Rules look as follows: \textit{if ARGM-LOC and ARGM-TMP both present, exchange them.}

\noindent Example: \texttt{[ARG0: Alex] [V: left] [ARG2: for Delhi] [ARGM-COM: with his wife] [ARGM-TMP: at 5 pm] . $\rightarrow$ Alex left for Delhi at 5 pm with his wife.}

\noindent The transformation relies on propbank annotations~\citep{bonial2012english,DBLP:conf/lrec/KingsburyP02,DBLP:journals/coling/PalmerKG05,DBLP:conf/acl/GildeaP02}.

\transformation{protaugment\_diverse\_paraphrase}{ProtAugment Diverse Paraphrasing}
\label{transfo:protaugment-diverse-paraphrase}

This transformation utilizes the \textsc{ProtAugment} method by \citet{DBLP:journals/corr/abs-2105-12995}. The paraphrase generation model is a BART model~\citep{lewis2020bart}, finetuned on the paraphrase generation task using 3 datasets: Google-PAWS~\citep{DBLP:conf/naacl/ZhangBH19}, MSR~\citep{dolan2005automatically}, Quora\footnote{\url{https://quoradata.quora.com/First-Quora-Dataset-Release-Question-Pairs}}. 

When parpahrasing a sentence, the transformation useu Diverse Beam Search~\citep{DBLP:journals/corr/VijayakumarCSSL16} to generate diverse outputs. The diversity penalty term is by default set to 0.5 but can be set to custom values. Additionally, the transformation can use the following generation constraints: (1) A fraction of the words in the input sentence are forbidden in the paraphrase (default 0.7). (2)  All bi-grams in the input sentence are forbidden in the paraphrase. This means the paraphrase cannot contain any bi-gram that are in the input sentence. This constraint enforces the paraphrase generation model to change the sentence structure. 

\transformation{punctuation}{Punctuation}
\label{transfo:punctuation}

This transformation removes/adds punctuation from an English sentence. This transformation was first introduced by \citet{mille2021automatic} and used as an example implemention for NL-Augmenter.

\transformation{quora\_trained\_t5\_for\_qa}{Question-Question Paraphraser for QA}
\label{transfo:qq-paraphraser}
This transformation creates new QA pairs by generating question paraphrases from a T5 model fine-tuned on Quora Question pairs~\footnote{\href{https://www.quora.com/q/quoradata/First-Quora-Dataset-Release-Question-Pairs}{Quora Question Pairs}}. Generated questions can have a very different surface form from the original question making it a strong paraphrase generator. A T5 model~\cite{raffel2019exploring, wolf-etal-2020-transformers} fine tuned~\footnote{\url{https://huggingface.co/ramsrigouthamg/t5\_paraphraser}} on the Quora Question Pairs dataset was being used to generate question paraphrases. This transformation would benefit Question Answering, Question Generation as well as other tasks which could indirectly benefit eg. for dialog tasks~\cite{shrivastava-etal-2021-saying, dhole2020resolving}. 

\transformation{redundant\_context\_for\_qa}{Question in CAPS}
\label{transfo:question-in-caps}

This transformation upper-cases the context of a question answering example. It also adds upper-cased versions of the original answers to the set of acceptable model responses. 

\transformation{random\_deletion}{Random Word Deletion}
\label{transfo:random-deletion}

This transformation randomly removes a word with a given probability $p$ (by default 0.25). The transformation relies on whitespace tokenization and thus only works for English and other languages that are segmented via whitespace. Due to the destructive nature of the transformation, it is likely that the meaning of a sequence may be changed as a result of the change. A similar transformation was suggested by \citet{DBLP:conf/emnlp/WeiZ19}. Word dropout~\cite{goldberg2017neural} has been common to help models understand unknown words encountered during evaluation by exposing them to this unknown-word condition during training itself. 

\transformation{random\_upper\_transformation}{Random Upper-Case Transformation}
\label{transfo:random-upper-transformation}

This perturbation adds noise to all types of text sources (sentence, paragraph, etc.) by randomly adding upper cased letters. With a default probably of 0.1, each character in a sequence is upper-cased. This transformation does not rely on a tokenizer and thus works with all languages that have upper and lower-case letters. One limiation of this transformation is that it will not affect a tokenizer that does lower case for all input. A similar transformation was suggested by \citet{DBLP:conf/emnlp/WeiZ19}. Further improvement of this transformation exists by potentially relying on extreme value theory \cite{Jalalzai2020repersentations}. 

\transformation{redundant\_context\_for\_qa}{Double Context QA}
\label{transfo:redundant-context-for-qa}

This transformation repeats the context of a question answering example. This should not change the result in any way.

\transformation{replace\_abbreviation\_and\_acronyms}{Replace Abbreviations and Acronyms}
\label{transfo:replace-abbreviation-and-acronyms}

This transformation changes abbreviations and acronyms appearing in an English text to their expanded form and respectively, changes expanded abbreviations and acronyms appearing in a text to their shorter form. For example, ``send this file asap to human resources'' might be changed to ``send this file as soon as possible to HR''.  The list of abbreviations and acronyms used in this transformation where manually gathered focusing on common abbreviations present in business communications. When abbreviation are context-dependent or highly specific, the induced change may change the meaning of a text, or an abbreviation may not be available in the lookup. The transformation was first introduced by \citet{DBLP:journals/corr/abs-2007-02033}.

\transformation{replace\_financial\_amounts}{Replace Financial Amounts}
\label{transfo:replace-financial-amounts}

This transformation replaces financial amounts throughout a text with the same value in a different currency. The replacement changes the amount, the writing format as well as the currency of the financial amount. For example, the sentence ``\textit{I owe Fred € 20 and I need € 10 for the bus.}'' might be changed to ``\textit{I owe Fred 2 906.37 Yen and I need 1 453.19 Yen for the bus.}''
The transformation was first introduced by \citet{DBLP:journals/corr/abs-2007-02033}.

\transformation{replace\_numerical\_values}{Replace Numerical Values}
\label{transfo:replace-numerical-values}

This transformation looks for numerical values in an English text and replaces it with another random value of the same cardinality. For example, ``\textit{6.9}'' may be replaced by ``\textit{4.2}'', or ``\textit{333}'' by ``\textit{789}''.
The transformation was first introduced by \citet{mille2021automatic}.

\transformation{replace\_spelling}{Replace Spelling}
\label{transfo:replace-spelling}

This transformation adds noise to all types of English text sources (sentence, paragraph, etc.) using corpora of common spelling errors introduced by \citet{deorowicz2005correcting}. Each word with a common misspelling is replaced by the version with mistake with a probability $p$ which by default is set to $0.2$.

\transformation{replace\_with\_hyponyms\_hypernyms}{Replace nouns with hyponyms or hypernyms}
\label{transfo:replace-with-hyponyms-hypernyms}

This transformation replaces common nouns with other related words that are either hyponyms or hypernyms. Hyponyms of a word are more specific in meaning (such as a sub-class of the word), eg: 'spoon' is a hyponym of 'cutlery'. Hypernyms are related words with a broader meaning (such as a generic category /super-class of the word), eg: 'colour' is a hypernym of 'red'. Not every word will have a hypernym or hyponym.

\transformation{sentence\_additions}{Sampled Sentence Additions}
\label{transfo:sentence-additions}

This transformation adds generated sentence to all types of English text sources (sentence, paragraph, etc.) by passing the input text to a GPT-2 model~\citep{radford2019language}.
By default, GPT-XL is used, together with the prompt ``\textit{paraphrase:}'' appended to the original text, after which up to 75 tokens are sampled. Since the additional text is sampled from a model, the model may introduce harmful language or generate text that contradicts the earlier text or changes its meaning. 
The idea to sample one or more additional sentences was first introduced by \citet{DBLP:conf/emnlp/JiaL17}.

\transformation{sentence\_reordering}{Sentence Reordering}
\label{transfo:sentence-reordering}
This perturbation adds noise to all types of text sources (paragraph, document, etc.) by randomly shuffling the order of sentences in the input text~\cite{lewis2020bart}. Sentences are first partially decontextualized by resolving coreference~\cite{lee2018higher}. 

This transformation is limited to input text that has more than one sentence. There are still cases where coreference can not be enough for decontextualization. For example, there could be occurences of ellipsis as demonstrated by~\citet{gangal2021nareor} or events could be mentioned in a narrative style which makes it difficult to perform re-ordering or shuffling~\cite{kovcisky2018narrativeqa} while keeping the context of the discourse intact.

\transformation{sentiment\_emoji\_augmenter}{Emoji Addition for Sentiment Data}
\label{transfo:sentiment-emoji-augmenter}

This transformation adds positive emojis and smileys to positive sentiment data and negative emojis to negative sentiment data. For non-labelled data, it adds neutral smileys.

\transformation{shuffle\_within\_segments}{Shuffle Within Segments}
\label{transfo:shuffle-within-segments}

In this transformation, a token sequence, for example BIO-tagged, is split into coherent segments. Thus, each segment corresponds to either a mention or a sequence of out-of-mention tokens. For example, a sentence ``\textit{She did not complain of headache or any other neurological symptoms .}'' with tags O O O O O B-problem O B-problem I-problem I-problem I-problem O is split into five segments: [\textit{She did not complain of}], [\textit{headache}], [\textit{or}], [\textit{any other neurological symptoms}], [\textit{.}]. Then for each segment, a binomial distribution (p=0.5) is used to decide whether it should be shuffled. If yes, the order of the tokens within the segment is shuffled while the label order is kept unchanged.
This transformation is inspired by \citet{dai-adel-2020-analysis}.

\transformation{simple\_ciphers}{Simple Ciphers}
\label{transfo:simple-ciphers}
This transformation modifies the text in ways that a human could rapidly decipher, but which make the input sequences almost completely unlike typical input sequences which are used during language model training. This transformation includes the following text modifications: double the characters, double the words, add spaces between the characters,  reverse all characters in the text, reverse the characters within each word, reverse the order of the words in the text, substitute homoglyphs, rot13 cipher.

\transformation{slangificator}{Slangificator}
\label{transfo:slangificator}

This transformation replaces some of the words (in particular, nouns, adjectives, and adverbs) of an English text with their corresponding slang. The replacement is done with the subset of the "Dictionary of English Slang \& Colloquialisms".\footnote{\url{http://www.peevish.co.uk/slang/index.htm}} The amount of replacement is proportional to the corresponding probabilities of replacement (by default, 0.5 for nouns, adjectives, and adverbs each).

\transformation{spanish\_gender\_swap}{Spanish Gender Swap}
\label{transfo:spanish-gender-swap}

This transformation changes the gender of all animate entities (mostly referring to people, and some animals) in a given Spanish sentence from masculine to feminine. This includes masculine nouns with feminine equivalents (e.g., \textit{doctor} → \textit{doctora}), nouns with a common gender (``sustantivos comunes en cuanto al género'', e.g., \textit{el violinista} → \textit{la violinista}), personal pronouns, and (optionally) given names often used with a given gender (e.g., \textit{Pedro} → \textit{Alicia}). Epicene nouns are excluded. In addition, the gender of adjectives, determiners, pronouns and participles are modified in order to maintain the grammatical agreement.

\transformation{speech\_disfluency\_perturbation}{Speech Disfluency Perturbation}
\label{transfo:speech-disfluency-perturbation}

This perturbation randomly inserts speech disfluencies in the form of filler words into English texts. With a given probability (0.2 by default), a speech disfluency is inserted between words. The default disfluencies are "um", "uh", "erm", "ah", and "er". At least one filler word is always inserted by this transformation.

\transformation{style\_paraphraser}{Paraphrasing through Style Transfer}
\label{transfo:style-paraphraser}

This transformation provides a range of possible styles of writing English language. The following styles can be chosen:

\begin{itemize}
    \item Shakespeare - Trained on written works by Shakespeare.
    \item Switchboard - Trained on a collection of conversational speech transcripts.
    \item Tweets - Trained on 5.2M English tweets.
    \item Bible - Trained on texts from the Bible.
    \item Romantic poetry - Trained on romantic poetry.
    \item Basic - A light, basic paraphraser with no specific style.
\end{itemize}

The transformation follows the models and formulations by \citet{krishna-etal-2020-reformulating}.

\transformation{subject\_object\_switch}{Subject Object Switch}
\label{transfo:subject-object-switch}

This transformation switches the subject and object of English sentences to generate new sentences with a very high surface similarity but very different meaning. This can be used, for example, for augmenting data for models that assess Semantic Similarity.

\transformation{summarization\_transformation}{Sentence Summarizaiton}
\label{transfo:summarization-transformation}

This transformation compresses English sentences by extracting subjects, verbs, and objects of the sentence. It also retains any negations.
For example, ``\textit{Stillwater is not a 2010 American live-action/animated dark fantasy adventure film}'' turns into ``\textit{Stillwater !is film}''.
\citet{zhang2021smat} used a similar idea to this transformation.

\transformation{suspecting\_paraphraser}{Suspecting Paraphraser for QA}
\label{transfo:suspecting-paraphraser}
This paraphraser transforms a yes/no question into a declarative sentence with a question tag~\footnote{\url{https://www.englishclub.com/grammar/tag-questions.htm}}, which helps to add more question specific informality to the dataset. Example: ''Did the American National Shipment company really break its own fleet?'' -> ''The American National Shipment company really broke its own fleet, didn't it''.

\transformation{swap\_characters}{Swap Characters Perturbation}
\label{transfo:swap-characters}
This perturbation randomly swaps two adjacent characters in a sentence or a paragraph with a default probability \cite{zhang2019adversarial}.

\transformation{synonym\_insertion}{Synonym Insertion}
\label{transfo:synonym-insertion}
This perturbation adds noise to all types of text sources (sentence, paragraph, etc.) by randomly inserting synonyms of randomly selected words excluding punctuations and stopwords \cite{marivate2020improving}.

\transformation{synonym\_substitution}{Synonym Substitution}
\label{transfo:synonym-substitution}
This perturbation randomly substitutes some words in an English text with their WordNet~\cite{miller1998wordnet} synonyms. 

\transformation{syntactically\_diverse\_paraphrase}{Syntactically Diverse Paraphrasing using Sow Reap models}
\label{transfo:syntactically-diverse-paraphrase}
This transformation is capable of generating multiple syntactically diverse paraphrases for a given sentence based on the work of ~\citet{goyal-durrett-2020-neural}. The model paraphrases inputs using a two step framework: 1) SOW (Source Order reWriting): This step enumerates multiple feasible syntactic transformations of the input sentence. 2) REAP (REarrangement Aware Paraphrasing): This step conditions on the multiple reorderings/ rearragements produced by SOW and outputs diverse paraphrases corresponding to these reoderings. The transformation is designed to work only on single-sentence inputs. Multi-sentence inputs results in an empty string/no transformation. The model are trained on the ParaNMT-50M dataset~\cite{wieting-17-millions, wieting-17-backtrans}, which can be argued to be a bit noisy. 

\transformation{tag\_subsequence\_substitution}{Subsequence Substitution for Sequence Tagging}
\label{transfo:tag-subsequence-substitution}

This transformation performs same-label subsequence substitution for the task of sequence tagging, which replaces a subsequence of the input tokens with another one that has the same sequence of tags~\cite{shi-etal-2021-substructure}.
This is done as follows: (1) Draw a subsequence A from the input (tokens, tags) tuple. (2) Draw a subsequence B within the whole dataset, with the same tag subsequence. (3) Substitute A with B in the input example.

\transformation{tense}{Change English Tense}
\label{transfo:tense}

This transformation converts English sentences from one tense to the other, for example simple present to simple past. This transformation was introduced by \citet{DBLP:conf/nips/LogeswaranLB18}.

\transformation{token\_replacement}{Token Replacement Based on Lookup Tables}
\label{transfo:token-replacement}

This transformation replaces input tokens with their perturbed versions sampled from a given lookup table of replacement candidates. Lookup tables containing OCR errors and misspellings from prior work are given as examples. Thus, by default, the transformation induces plausible OCR errors and human typos to the input sentence.

The transformation is an adapted and improved version of the lookup table-based noise induction method from \citet{namysl-etal-2020-nat}. The OCR lookup table is from \citet{namysl-etal-2021-empirical} and the misspellings from \citet{piktus-etal-2019-misspelling}.

\transformation{transformer\_fill}{Transformer Fill}
\label{transfo:transformer-fill}

This perturbation replaces words based on recommendations from a masked language model. The transformation can limit replacements to certain POS tags (all enabled by default). Many previous papers have used this technique for data augmentation~\citep[][inter alia]{ribeiro2020beyond,DBLP:journals/corr/abs-2009-07502}.

\transformation{underscore\_trick}{Underscore Trick}
\label{transfo:underscore-trick}
This perturbation adds noise to the text sources like sentence, paragraph, etc.
This transformation acts like a perturbation to test robustness. It replaces some random spaces with underscores (or even other selected symbols). This perturbation would benefit all tasks which have a sentence/paragraph/document as input like text classification and text generation, especially on tasks related to understanding/generating scripts.

\transformation{unit\_converter}{Unit converter}
\label{transfo:unit-converter}
This transformation converts length and weight measures to different units (e.g., kilometers to miles) picking at random the new unit but converting accurately the quantity. The transformation conserves the format of the original quantity: "100 pounds" is converted to "1600 ounces" but "one-hundred pounds" is converted to "one thousand, six hundred ounces". Generated transformations display high similarity to the source sentences.

\transformation{urban\_dict\_swap}{Urban Thesaurus Swap}
\label{transfo:urban-dict-swap}
This perturbation randomly picks nouns from the input source to convert to related terms drawn from the Urban Dictionary~\footnote{\url{https://www.urbandictionary.com/}} resource. It can be applied to an input text to produce semantically-similar output texts in order to generate more robust test sets. We first select nouns at random, then query the Urban Thesaurus website~\footnote{\url{https://urbanthesaurus.org/}} to obtain a list of related terms to swap in \cite{wilson-etal-2020-urban}.

\transformation{use\_acronyms}{Use Acronyms}
\label{transfo:use-acronyms}
This transformation changes groups of words for their equivalent acronyms. It's a simple substitution of groups of words for their acronyms. It helps to increase the size of the dataset as well as improving the understanding of acronyms of models trained on data augmented with this transformation. This transformations works to increase the data for any task that has input texts. It is specially interesting for tasks on semantic similarity, where models should be aware of the equivalence between a set of words and their acronym. The quality of the transformation depends on the list of acronyms. As of now, this list was scraped from wikipedia's List of Acronyms~\footnote{\url{https://en.wikipedia.org/wiki/Lists\_of\_acronyms}} and naively filtered, which leaves space for improvement .

\transformation{visual\_attack\_letters}{Visual Attack Letter}
\label{transfo:visual-attack-letters}
This perturbation replaces letters with visually similar, but different, letters. Every letter was embedded into 576-dimensions. The nearest neighbors are obtained through cosine distance. To obtain the embeddings the letter was resized into a 24x24 image, then flattened and scaled. This follows the Image Based Character Embedding (ICES)~\cite{eger-etal-2019-text}.

The top neighbors from each letter are chosen. Some were removed by judgment (e.g. the nearest neighbors for 'v' are many variations of the letter 'y') which did not qualify from the image embedding \cite{DBLP:journals/corr/abs-1903-11508}.

\transformation{weekday\_month\_abbreviation}{Weekday Month Abbreviation}
\label{transfo:weekday-month-abbreviation}
This transformation abbreviates or expands the names of months and weekdays, e.g. Mon. -> Monday. Generated transformations display high similarity to the source sentences and does not alter the meaning and the semantic of the original texts. It does not abbreviate plural names, e.g. Sundays, and does not influence texts without names of weekdays or months.

\transformation{whitespace\_perturbation}{Whitespace Perturbation}
\label{transfo:whitespace-perturbation}
This perturbation adds noise to text by randomly removing or adding whitespaces.

\transformation{word\_noise}{Context Noise for QA}
\label{transfo:word-noise}
This transformation chooses a set of words at random from the context and the question and forms a sentence out of them. The sentence is then prepended or appended to the context to create a new QA pair. The transformation is inspired by the the \textbf{AddAny} method described in Adversarial SQUAD~\cite{jia-liang-2017-adversarial}. However, instead of probing the model to generate adversaries, random words from the context and the question are simply selected and joined together into a sentence, ignoring grammaticality. The transformation attempts to create novel QA pairs assuming that the introduction of random words to the context is less likely to change the answer choice to an asked question.

\transformation{writing\_system\_replacement}{Writing System Replacement}
\label{transfo:writing-system-replacement}
This transformation replaces the writing system of the input with another writing system. We use CJK Unified Ideographs\footnote{\url{https://en.wikipedia.org/wiki/CJK\_Unified\_Ideographs}} as the source of characters for the generated writing systems. The transformation would benefit text classification tasks, especially in the cases where the input writing system is undeciphered.

%\begin{CJK*}{UTF8}
%\texttt{\manicule I love potatoes \arrow\ 驿掩㑇㕶誨, I love potatoes \arrow\ 之 笓䒉㘔䆇 躓䒉蝲討蝲䒉䆇䁣}
%\end{CJK*}

\transformation{yes\_no\_question}{Yes-No Question Perturbation}
\label{transfo:yes-no-question}
This transformation turns English non-compound statements into yes-no questions. The generated questions can be answered by the statements that were used to generate them. The text is left largely unchanged other than the fronted/modified/added auxiliaries and be-verbs.

The transformation works by getting dependency parse and POS tags from a machine learning model and applying human-engineered, rule-based transformations to those parses/tags. This transformation would particularly benefit question-answering and question-generation tasks, as well as providing surplus legal text for language modeling and masked language modeling.

\transformation{yoda\_transform}{Yoda Transformation}
\label{transfo:yoda-transform}
This perturbation modifies sentences to flip the clauses such that it reads like "Yoda Speak". For example, "Much to learn, you still have". This form of construction is sometimes called "XSV", where "the “X” being a stand-in for whatever chunk of the sentence goes with the verb", and appears very rarely in English normally. The rarity of this construction in ordinary language makes it particularly well suited for NL augmentation and serves as a relatively easy but potentially powerful test of robustness.

%% file: filter_list.tex
\section{Filters}

The following is the list of all submitted filters to NL-Augmenter. Filters are used to filter data and create subpopulations of given inputs, according to features such as input complexity, input size, etc. Therefore, the output of a filter is a boolean value, indicating that whether the input meet the filter criterion. We discuss the implementations of each filter alongwith their limitations. The title of each filter subsection is clickable and redirects to the actual python implementation.

\filter{code\_mixing}{Code-Mixing Filter}
\label{filter:code-mixing}
This filter identifies whether the input text is code-mixed. It checks that there is at least one sentence in the text where there are tokens representing at least `k' unique languages (with at least a `threshold` level of confidence that the token is of that language). It is useful for collecting code-mixed data to test the model's performance on multilingual tasks. The filter relies on \texttt{ftlid}\footnote{\url{https://pypi.org/project/ftlid/}} for language detection, therefore, this filter might be limited by the performance of the language detection tool.

\manicule (containing code-mixing) Yo estaba con Esteban yesterday, he was telling me about lo que su esposa vio en los Estados Unidos. \arrow \add{\chk True}

\filter{diacritic\_filter}{Diacritics Filter}
\label{filter:diacritic}
This filter checks whether any character in the sentence has a diacritic. It can be used to create splits of the dataet where the sentences have diacritics. Accented characters are typically among the rarer characters and checking the model performance on such a split might help investigate model robustness.

\manicule (containing diacritics) She lookèd east an she lookèd west. \arrow \add{\chk True}

\filter{encoding}{Encoding Filter}
\label{filter:encoding}
This filter filters examples which contain characters outside a given encoding. It can be used to find examples containing e.g. non-ASCII Unicode characters. Filtering out and testing examples that contain these characters can provide feedback on how to improve the models accordingly, since most models are trained with plain English text, which contains mostly ASCII characters. Sometimes non-ASCII character are even explicitly stripped away.

\manicule (containing non-ASCII characters) That souvenir sure was expensive at 60£.. or was it 60€? \arrow \add{\chk True}

\filter{englishness}{Englishness Filter}
\label{filter:englishness}
This filter identifies texts that contain uniquely British spellings, vocabulary, or slang. The filter uses a vocabulary of common British words/phrases and checks the number of occurence of British words in the given texts. The text is selected if the number exceeds a pre-defined threshold. 

\manicule (containing British spellings) Colour is an attribute of light that is perceived by the human eye. \arrow \add{\chk True}

\filter{gender\_bias}{Gender Bias Filter}
\label{filter:gender-bias}
This filter filters a text corpus to measure gender fairness with respect to a female gender representation. It supports four languages (i.e. English, French, Polish and Russian) and can be used to define whether the female gender is sufficiently represented in a tested subset of sentences. The filter uses a list of lexicals, which includes filter categories such as personal pronouns, words defining the relation, titles and names, corresponding to the female and male genders accordingly.

\manicule (texts with unbalanced representation) "He went home", "He drives a car", "She has returned" \arrow \add{\chk True}

\filter{group\_inequity}{Group Inequity Filter}
\label{filter:group-inequity}
This is a bilingual filter (for English and French languages), which helps to discover potential group inequity issues in the text corpus. It is a topic agnostic filter which accepts user-defined parameters, consisting of keywords inherent to minor group (which potentially might suffer from the discrimination), major group, minor factor and major factor. The filter first flags the sentences as belonging to the minor, and the major groups, and then, the sentences from each of the groups are used to define the intersection with both factors. The filter then compares whether the percentage of major factors exceeds that of the minor factors to determine if the sentences have group inequity issues.

\manicule (containing group inequity issues) "He is a doctor", "She is a nurse", "She works at the hospital" \arrow \add{\chk True}

\filter{keywords}{Keyword Filter}
\label{filter:keywords}
This is a simple filter, which filters examples based on a pre-defined set of keywords. It can be useful in creating splits for a specific domain.

\manicule (containing keyword "at") Andrew played cricket in India \arrow \add{\chk True}

\filter{lang}{Language Filter}
\label{filter:lang}
This filter selects texts that match any of a given set of ISO 639-1 language codes (the default language being English). Language matching is performed using a pre-trained \texttt{langid.py} model instance. The model provides normalized confidence scores. A minimum threshold score needs to be set, and all sentences with confidence scores above this threshold are accepted by the filter.

\manicule (is English texts) Mein Luftkissenfahrzeug ist voller Aale \arrow \remove{\xxx False}

\filter{length}{Length Filter}
\label{filter:length}
This filter filters data with the input text length matching a specified threshold. It can be useful in creating data with different length distributions.

\manicule (containing more than 3 words) Andrew played cricket in India \arrow \add{\chk True} 

\filter{named-entity-count}{Named-entity-count Filter}
\label{filter:named-entity-count}
This filter filters data where the number of Named Entities in the input match a specified threshold (based on the supported conditions). 

\manicule (containing more than 1 named entity) Novak Djokovic is the greatest tennis player of all time. \arrow \add{\chk True}

\filter{numeric}{Numeric Filter}
\label{filter:numeric}
This filter filters example which contain a numeric value. In the tasks like textual entailment, question answering etc., a quantity (number) could directly affect the final label/response. This filter can be used to create splits to measure the performance separately on texts containing numeric values.

\manicule (containing numbers in texts) John bought a car worth dollar twenty five thousand . \arrow \add{\chk True} 

\filter{oscillatory\_hallucination}{Oscillatory Hallucinations Filter}
\label{filter:oscillatory-hallucination}
This filter is designed to operate in text generation systems' outputs, with the purpose of extracting oscillatory hallucinations. Oscillatory hallucinations are one class of hallucinations characterized by repeating bigram structure in the output\cite{raunak-etal-2021-curious}. Typically, these behaviors are observed in models trained on noisy corpora. The filter counts the frequency of bigrams in both source and target texts, and compare the frequency difference with a pre-set threshold to determine whether the texts includes oscillatory hallucinations.

\manicule (containing hallucinations in target texts) Source: "Community, European Parliament common regional policy, Mediterranean region", Target: "Arbeitsbedingungen, berufliche Bildung, berufliche Bildung, berufliche Bildung" \arrow \add{\chk True} 

\filter{polarity}{Polarity Filter}
\label{filter:polarity}
This filter filters a transformed text if it does not retain the same polarity as an original text. This filter helps not to distort training data during augmentation for sentiment analysis-related tasks. While generating new data for a sentiment analysis task, it is important to make sure that generated data is labelled correctly. 

\manicule (texts retaining polarity) "Hotel is terrible", "Hotel is great" \arrow \remove{\xxx False} 

\filter{quantitative\_ques}{Quantitative Question Filter}
\label{filter:quantitative-ques}
This is a simple rule-based filter that can be used to identify quantitative questions. It can help to analyse models' performance on questions which require numerical understanding. It is also useful to study possible biases in question generation. 

\manicule (being quantitative question) How long does the journey take? \arrow \add{\chk True}

\filter{question\_filter}{Question type filter}
\label{filter:question-filter}
This filter helps identify the question category of a question answering example based on the question word or the named entity type of the answer. Knowledge of the question type can help in the development of question answering systems~\cite{DBLP:journals/corr/abs-1904-02651} as well as for assessing performance on individual splits.

\manicule (being where question) Where is Delhi located ? \arrow \add{\chk True}

\filter{repetitions}{Repetitions Filter}
\label{filter:repetitions}
This filter finds texts with repetitions with simple heuristic rules. It might be helpful in finding repetitions that frequently occur in the spoken language data.

\manicule (containing repetitions in texts) I I want to sleep \arrow \add{\chk True}

\filter{soundex}{Phonetic Match Filter}
\label{filter:soundex}
This filter selects texts that contain matching entries to a list of supplied keywords. It first transform the input sentence and the keywords into phenetic units and then compare whether the two phenetic unit sets have overlap.

\manicule (containing homophones of keyword "trombone") I left my trombno on the train \arrow \add{\chk True}

\filter{special\_casing}{Special Casing Filter}
\label{filter:special-casing}
This filter checks if the input sentence has a special casing, i.e. the string is either all lowercased, all uppercased or has title casing. It might be useful for creating splits that contain texts with unusual casing, e.g. misspellings.

\manicule (text being uppercased/lowercased/titlecased) let's go to chipotle \arrow \add{\chk True}

\filter{speech\_tag}{Speech-Tag Filter}
\label{filter:speech-tag}
This filter filters an example text based on a set of speech tags and identifies whether the count of selected POS tags meet the pre-defined conditions (e.g. above the threshold).

\manicule (containing 1 verb and 2 numbers in texts) It all happened between November 2007 and November 2008. \arrow \add{\chk True}

\filter{token\_amount}{Token-Amount filter}
\label{filter:token-amount}
This filter filters an example text based on whether certain keywords are present in a specified amount.

\manicule (containing 2 occurances of "in") Andrew played cricket in a soccer stadium in India at 9pm \arrow \add{\chk True}

\filter{toxicity}{Toxicity Filter}
\label{filter:toxicity}
This filter filters an example text which has a toxicity value matching a particular threshold. It uses a pre-trained toxicity detector, which can provide 7 toxicity scores. All the 7 types of toxicity scores can be used as criteria for the filtering.

\manicule (text being toxic) I disagree. It is not supposed to work that way. \arrow \remove{\xxx False}

\filter{universal\_bias}{Universal Bias Filter}
\label{filter:universal-bias}
This filter works the same way as the Gender Bias Filter, but measures balance of representation for more categories (religion, race, ethnicity, gender, sexual orientation, age, appearance, disability, experience, education, economic status). The lexical seeds representing these categories are currently available in English only, however the pool of languages can be extended by a simple addition of the lexical seeds in a desired language to the lexicals.json file.

\manicule (texts being biased) "He is going to make a cake.", "She is going to program", "Nobody likes washing dishes", "She agreed to help him" \arrow \remove{\xxx False}

\filter{yesno\_question}{Yes/no question filter}
\label{filter:yesno-question}
This filter allows to select questions that can be correctly answered with either 'yes' or 'no'. Since it is rule-based, the limitation of this filter is that questions that are ambiguous might not be recognized.

\manicule (text being yes/no question) Wasn't she angry when you told her about the accident? \arrow \add{\chk True}

%% file: nertstyle.bbl
\begin{thebibliography}{139}
\expandafter\ifx\csname natexlab\endcsname\relax\def\natexlab#1{#1}\fi

\bibitem[{Res(2006)}]{RespectfulDisability}
 2006.
\newblock {Respectful Disability Language: Here’s What’s Up!}
\newblock \url{https://www.aucd.org/docs/add/sa_summits/Language%20Doc.pdf}.

\bibitem[{Bamman(2017)}]{bamman2017natural}
David Bamman. 2017.
\newblock Natural language processing for the long tail.
\newblock In \emph{DH}.

\bibitem[{Berard et~al.(2019)Berard, Calapodescu, and Roux}]{berard2019naver}
Alexandre Berard, Ioan Calapodescu, and Claude Roux. 2019.
\newblock Naver labs europe's systems for the wmt19 machine translation
  robustness task.
\newblock \emph{arXiv preprint arXiv:1907.06488}.

\bibitem[{Bhagat and Hovy(2013)}]{bhagat2013paraphrase}
Rahul Bhagat and Eduard Hovy. 2013.
\newblock What is a paraphrase?
\newblock \emph{Computational Linguistics}, 39(3):463--472.

\bibitem[{Bhatt and Dhole(2020)}]{bhatt2020benchmarking}
Abhinav Bhatt and Kaustubh~D. Dhole. 2020.
\newblock \href {http://arxiv.org/abs/2006.00533} {Benchmarking biorelex for
  entity tagging and relation extraction}.

\bibitem[{Bird(2006)}]{bird2006nltk}
Steven Bird. 2006.
\newblock Nltk: the natural language toolkit.
\newblock In \emph{Proceedings of the COLING/ACL 2006 Interactive Presentation
  Sessions}, pages 69--72.

\bibitem[{Board(2021)}]{Dyslexia}
Smorga's Board. 2021.
\newblock \href
  {https://www.teacherspayteachers.com/Product/Frequently-Misspelled-Word-List-for-Dyslexia-5295631}
  {Frequently misspelled word list for dyslexia}.

\bibitem[{Bonial et~al.(2012)Bonial, Hwang, Bonn, Conger, Babko-Malaya, and
  Palmer}]{bonial2012english}
Claire Bonial, Jena Hwang, Julia Bonn, Kathryn Conger, Olga Babko-Malaya, and
  Martha Palmer. 2012.
\newblock English propbank annotation guidelines.
\newblock \emph{Center for Computational Language and Education Research
  Institute of Cognitive Science University of Colorado at Boulder}, 48.

\bibitem[{Chen et~al.(2021)Chen, Tam, Raffel, Bansal, and
  Yang}]{chen2021empirical}
Jiaao Chen, Derek Tam, Colin Raffel, Mohit Bansal, and Diyi Yang. 2021.
\newblock \href {http://arxiv.org/abs/2106.07499} {An empirical survey of data
  augmentation for limited data learning in nlp}.

\bibitem[{Dai and Adel(2020)}]{dai-adel-2020-analysis}
Xiang Dai and Heike Adel. 2020.
\newblock \href {https://doi.org/10.18653/v1/2020.coling-main.343} {An analysis
  of simple data augmentation for named entity recognition}.
\newblock In \emph{Proceedings of the 28th International Conference on
  Computational Linguistics}, pages 3861--3867, Barcelona, Spain (Online).
  International Committee on Computational Linguistics.

\bibitem[{Damodaran()}]{Styleformer}
Prithiviraj Damodaran.
\newblock \href {https://github.com/PrithivirajDamodaran/Styleformer}
  {Styleformer}.

\bibitem[{Deorowicz and Ciura(2005)}]{deorowicz2005correcting}
Sebastian Deorowicz and Marcin~G Ciura. 2005.
\newblock Correcting spelling errors by modelling their causes.
\newblock \emph{International journal of applied mathematics and computer
  science}, 15:275--285.

\bibitem[{Dhole(2020)}]{dhole2020resolving}
Kaustubh~D. Dhole. 2020.
\newblock \href {http://arxiv.org/abs/2008.07559} {Resolving intent ambiguities
  by retrieving discriminative clarifying questions}.

\bibitem[{Dinan et~al.(2020)Dinan, Fan, Williams, Urbanek, Kiela, and
  Weston}]{dinan-etal-2020-queens}
Emily Dinan, Angela Fan, Adina Williams, Jack Urbanek, Douwe Kiela, and Jason
  Weston. 2020.
\newblock \href {https://doi.org/10.18653/v1/2020.emnlp-main.656} {Queens are
  powerful too: Mitigating gender bias in dialogue generation}.
\newblock In \emph{Proceedings of the 2020 Conference on Empirical Methods in
  Natural Language Processing (EMNLP)}, pages 8173--8188, Online. Association
  for Computational Linguistics.

\bibitem[{Dolan and Brockett(2005)}]{dolan2005automatically}
William~B Dolan and Chris Brockett. 2005.
\newblock Automatically constructing a corpus of sentential paraphrases.
\newblock In \emph{Proceedings of the Third International Workshop on
  Paraphrasing (IWP2005)}.

\bibitem[{Dopierre et~al.(2021)Dopierre, Gravier, and
  Logerais}]{DBLP:journals/corr/abs-2105-12995}
Thomas Dopierre, Christophe Gravier, and Wilfried Logerais. 2021.
\newblock \href {http://arxiv.org/abs/2105.12995} {Protaugment: Unsupervised
  diverse short-texts paraphrasing for intent detection meta-learning}.
\newblock \emph{CoRR}, abs/2105.12995.

\bibitem[{Eger and Benz(2020)}]{eger2020hero}
Steffen Eger and Yannik Benz. 2020.
\newblock \href {http://arxiv.org/abs/2010.05648} {From hero to zéroe: A
  benchmark of low-level adversarial attacks}.

\bibitem[{Eger et~al.(2019{\natexlab{a}})Eger, {\c{S}}ahin, R{\"u}ckl{\'e},
  Lee, Schulz, Mesgar, Swarnkar, Simpson, and Gurevych}]{eger-etal-2019-text}
Steffen Eger, G{\"o}zde~G{\"u}l {\c{S}}ahin, Andreas R{\"u}ckl{\'e}, Ji-Ung
  Lee, Claudia Schulz, Mohsen Mesgar, Krishnkant Swarnkar, Edwin Simpson, and
  Iryna Gurevych. 2019{\natexlab{a}}.
\newblock \href {https://doi.org/10.18653/v1/N19-1165} {Text processing like
  humans do: Visually attacking and shielding {NLP} systems}.
\newblock In \emph{Proceedings of the 2019 Conference of the North {A}merican
  Chapter of the Association for Computational Linguistics: Human Language
  Technologies, Volume 1 (Long and Short Papers)}, pages 1634--1647,
  Minneapolis, Minnesota. Association for Computational Linguistics.

\bibitem[{Eger et~al.(2019{\natexlab{b}})Eger, Sahin, R{\"{u}}ckl{\'{e}}, Lee,
  Schulz, Mesgar, Swarnkar, Simpson, and
  Gurevych}]{DBLP:journals/corr/abs-1903-11508}
Steffen Eger, G{\"{o}}zde~G{\"{u}}l Sahin, Andreas R{\"{u}}ckl{\'{e}}, Ji{-}Ung
  Lee, Claudia Schulz, Mohsen Mesgar, Krishnkant Swarnkar, Edwin Simpson, and
  Iryna Gurevych. 2019{\natexlab{b}}.
\newblock \href {http://arxiv.org/abs/1903.11508} {Text processing like humans
  do: Visually attacking and shielding {NLP} systems}.
\newblock \emph{CoRR}, abs/1903.11508.

\bibitem[{Eisner et~al.(2016)Eisner, Rockt{\"a}schel, Augenstein,
  Bo{\v{s}}njak, and Riedel}]{eisner-etal-2016-emoji2vec}
Ben Eisner, Tim Rockt{\"a}schel, Isabelle Augenstein, Matko Bo{\v{s}}njak, and
  Sebastian Riedel. 2016.
\newblock \href {https://doi.org/10.18653/v1/W16-6208} {emoji2vec: Learning
  emoji representations from their description}.
\newblock In \emph{Proceedings of The Fourth International Workshop on Natural
  Language Processing for Social Media}, pages 48--54, Austin, TX, USA.
  Association for Computational Linguistics.

\bibitem[{Fadaee et~al.(2017)Fadaee, Bisazza, and Monz}]{fadaee-etal-2017-data}
Marzieh Fadaee, Arianna Bisazza, and Christof Monz. 2017.
\newblock \href {https://doi.org/10.18653/v1/P17-2090} {Data augmentation for
  low-resource neural machine translation}.
\newblock In \emph{Proceedings of the 55th Annual Meeting of the Association
  for Computational Linguistics (Volume 2: Short Papers)}, pages 567--573,
  Vancouver, Canada. Association for Computational Linguistics.

\bibitem[{Fan et~al.(2021)Fan, Bhosale, Schwenk, Ma, El-Kishky, Goyal, Baines,
  Celebi, Wenzek, Chaudhary et~al.}]{fan2021beyond}
Angela Fan, Shruti Bhosale, Holger Schwenk, Zhiyi Ma, Ahmed El-Kishky,
  Siddharth Goyal, Mandeep Baines, Onur Celebi, Guillaume Wenzek, Vishrav
  Chaudhary, et~al. 2021.
\newblock Beyond english-centric multilingual machine translation.
\newblock \emph{Journal of Machine Learning Research}, 22(107):1--48.

\bibitem[{Fan et~al.(2020)Fan, Bhosale, Schwenk, Ma, El-Kishky, Goyal, Baines,
  Çelebi, Wenzek, Chaudhary, Goyal, Birch, Liptchinsky, Edunov, Grave, Auli,
  and Joulin}]{Fan2020BeyondEM}
Angela Fan, Shruti Bhosale, Holger Schwenk, Zhiyi Ma, Ahmed El-Kishky,
  Siddharth Goyal, Mandeep Baines, Onur Çelebi, Guillaume Wenzek, Vishrav
  Chaudhary, Naman Goyal, Tom Birch, Vitaliy Liptchinsky, Sergey Edunov,
  Edouard Grave, Michael Auli, and Armand Joulin. 2020.
\newblock Beyond english-centric multilingual machine translation.
\newblock \emph{ArXiv}, abs/2010.11125.

\bibitem[{Feng et~al.(2021)Feng, Gangal, Wei, Chandar, Vosoughi, Mitamura, and
  Hovy}]{feng2021survey}
Steven~Y Feng, Varun Gangal, Jason Wei, Sarath Chandar, Soroush Vosoughi,
  Teruko Mitamura, and Eduard Hovy. 2021.
\newblock A survey of data augmentation approaches for nlp.
\newblock \emph{arXiv preprint arXiv:2105.03075}.

\bibitem[{Galton(1907)}]{galton1907vox}
Francis Galton. 1907.
\newblock Vox populi (the wisdom of crowds).
\newblock \emph{Nature}, 75(7):450--451.

\bibitem[{Gangal et~al.(2021)Gangal, Feng, Hovy, and
  Mitamura}]{gangal2021nareor}
Varun Gangal, Steven~Y Feng, Eduard Hovy, and Teruko Mitamura. 2021.
\newblock Nareor: The narrative reordering problem.
\newblock \emph{arXiv preprint arXiv:2104.06669}.

\bibitem[{Gardner et~al.(2020)Gardner, Artzi, Basmov, Berant, Bogin, Chen,
  Dasigi, Dua, Elazar, Gottumukkala et~al.}]{gardner2020evaluating}
Matt Gardner, Yoav Artzi, Victoria Basmov, Jonathan Berant, Ben Bogin, Sihao
  Chen, Pradeep Dasigi, Dheeru Dua, Yanai Elazar, Ananth Gottumukkala, et~al.
  2020.
\newblock Evaluating models’ local decision boundaries via contrast sets.
\newblock In \emph{Proceedings of the 2020 Conference on Empirical Methods in
  Natural Language Processing: Findings}, pages 1307--1323.

\bibitem[{Gauthier et~al.(2020)Gauthier, Hu, Wilcox, Qian, and
  Levy}]{gauthier2020syntaxgym}
Jon Gauthier, Jennifer Hu, Ethan Wilcox, Peng Qian, and Roger Levy. 2020.
\newblock Syntaxgym: An online platform for targeted evaluation of language
  models.
\newblock In \emph{Proceedings of the 58th Annual Meeting of the Association
  for Computational Linguistics: System Demonstrations}, pages 70--76.

\bibitem[{Gehrmann et~al.(2021)Gehrmann, Adewumi, Aggarwal, Ammanamanchi,
  Aremu, Bosselut, Chandu, Clinciu, Das, Dhole, Du, Durmus, Du{\v{s}}ek,
  Emezue, Gangal, Garbacea, Hashimoto, Hou, Jernite, Jhamtani, Ji, Jolly, Kale,
  Kumar, Ladhak, Madaan, Maddela, Mahajan, Mahamood, Majumder, Martins,
  McMillan-Major, Mille, van Miltenburg, Nadeem, Narayan, Nikolaev,
  Niyongabo~Rubungo, Osei, Parikh, Perez-Beltrachini, Rao, Raunak, Rodriguez,
  Santhanam, Sedoc, Sellam, Shaikh, Shimorina, Sobrevilla~Cabezudo, Strobelt,
  Subramani, Xu, Yang, Yerukola, and Zhou}]{gehrmann2021gem}
Sebastian Gehrmann, Tosin Adewumi, Karmanya Aggarwal, Pawan~Sasanka
  Ammanamanchi, Anuoluwapo Aremu, Antoine Bosselut, Khyathi~Raghavi Chandu,
  Miruna-Adriana Clinciu, Dipanjan Das, Kaustubh Dhole, Wanyu Du, Esin Durmus,
  Ond{\v{r}}ej Du{\v{s}}ek, Chris~Chinenye Emezue, Varun Gangal, Cristina
  Garbacea, Tatsunori Hashimoto, Yufang Hou, Yacine Jernite, Harsh Jhamtani,
  Yangfeng Ji, Shailza Jolly, Mihir Kale, Dhruv Kumar, Faisal Ladhak, Aman
  Madaan, Mounica Maddela, Khyati Mahajan, Saad Mahamood, Bodhisattwa~Prasad
  Majumder, Pedro~Henrique Martins, Angelina McMillan-Major, Simon Mille, Emiel
  van Miltenburg, Moin Nadeem, Shashi Narayan, Vitaly Nikolaev, Andre
  Niyongabo~Rubungo, Salomey Osei, Ankur Parikh, Laura Perez-Beltrachini,
  Niranjan~Ramesh Rao, Vikas Raunak, Juan~Diego Rodriguez, Sashank Santhanam,
  Jo{\~a}o Sedoc, Thibault Sellam, Samira Shaikh, Anastasia Shimorina,
  Marco~Antonio Sobrevilla~Cabezudo, Hendrik Strobelt, Nishant Subramani, Wei
  Xu, Diyi Yang, Akhila Yerukola, and Jiawei Zhou. 2021.
\newblock \href {https://doi.org/10.18653/v1/2021.gem-1.10} {The {GEM}
  benchmark: Natural language generation, its evaluation and metrics}.
\newblock In \emph{Proceedings of the 1st Workshop on Natural Language
  Generation, Evaluation, and Metrics (GEM 2021)}, pages 96--120, Online.
  Association for Computational Linguistics.

\bibitem[{Gehrmann et~al.(2022)Gehrmann, Bhattacharjee, Mahendiran, Wang,
  Papangelis, Madaan, McMillan-Major, Shvets, Upadhyay, Yao, Wilie,
  Bhagavatula, You, Thomson, Garbacea, Wang, Deutsch, Xiong, Jin, Gkatzia,
  Radev, Clark, Durmus, Ladhak, Ginter, Winata, Strobelt, Hayashi, Novikova,
  Kanerva, Chim, Zhou, Clive, Maynez, Sedoc, Juraska, Dhole, Chandu,
  Perez-Beltrachini, Ribeiro, Tunstall, Zhang, Pushkarna, Creutz, White, Kale,
  Eddine, Daheim, Subramani, Dusek, Liang, Ammanamanchi, Zhu, Puduppully, Kriz,
  Shahriyar, Cardenas, Mahamood, Osei, Cahyawijaya, Štajner, Montella,
  {Shailza}, Jolly, Mille, Hasan, Shen, Adewumi, Raunak, Raheja, Nikolaev,
  Tsai, Jernite, Xu, Sang, Liu, and Hou}]{gem2-0}
Sebastian Gehrmann, Abhik Bhattacharjee, Abinaya Mahendiran, Alex Wang,
  Alexandros Papangelis, Aman Madaan, Angelina McMillan-Major, Anna Shvets,
  Ashish Upadhyay, Bingsheng Yao, Bryan Wilie, Chandra Bhagavatula, Chaobin
  You, Craig Thomson, Cristina Garbacea, Dakuo Wang, Daniel Deutsch, Deyi
  Xiong, Di~Jin, Dimitra Gkatzia, Dragomir Radev, Elizabeth Clark, Esin Durmus,
  Faisal Ladhak, Filip Ginter, Genta~Indra Winata, Hendrik Strobelt, Hiroaki
  Hayashi, Jekaterina Novikova, Jenna Kanerva, Jenny Chim, Jiawei Zhou, Jordan
  Clive, Joshua Maynez, João Sedoc, Juraj Juraska, Kaustubh Dhole,
  Khyathi~Raghavi Chandu, Laura Perez-Beltrachini, Leonardo F.~R. Ribeiro,
  Lewis Tunstall, Li~Zhang, Mahima Pushkarna, Mathias Creutz, Michael White,
  Mihir~Sanjay Kale, Moussa~Kamal Eddine, Nico Daheim, Nishant Subramani,
  Ondrej Dusek, Paul~Pu Liang, Pawan~Sasanka Ammanamanchi, Qi~Zhu, Ratish
  Puduppully, Reno Kriz, Rifat Shahriyar, Ronald Cardenas, Saad Mahamood,
  Salomey Osei, Samuel Cahyawijaya, Sanja Štajner, Sebastien Montella,
  {Shailza}, Shailza Jolly, Simon Mille, Tahmid Hasan, Tianhao Shen, Tosin
  Adewumi, Vikas Raunak, Vipul Raheja, Vitaly Nikolaev, Vivian Tsai, Yacine
  Jernite, Ying Xu, Yisi Sang, Yixin Liu, and Yufang Hou. 2022.
\newblock \href {https://doi.org/10.48550/ARXIV.2206.11249} {Gemv2:
  Multilingual nlg benchmarking in a single line of code}.

\bibitem[{Gildea and Palmer(2002)}]{DBLP:conf/acl/GildeaP02}
Daniel Gildea and Martha~Stone Palmer. 2002.
\newblock \href {https://doi.org/10.3115/1073083.1073124} {The necessity of
  parsing for predicate argument recognition}.
\newblock In \emph{Proceedings of the 40th Annual Meeting of the Association
  for Computational Linguistics, July 6-12, 2002, Philadelphia, PA, {USA}},
  pages 239--246. {ACL}.

\bibitem[{Goel et~al.(2021)Goel, Rajani, Vig, Tan, Wu, Zheng, annd
  Mohit~Bansal, and R\'e}]{goel2021robustness}
Karan Goel, Nazneen Rajani, Jesse Vig, Samson Tan, Jason Wu, Stephan Zheng,
  Caiming~Xiong annd Mohit~Bansal, and Christopher R\'e. 2021.
\newblock \href {https://arxiv.org/abs/2101.04840} {Robustness {G}ym:
  {U}nifying the {NLP} evaluation landscape}.
\newblock \emph{arXiv preprint arXiv:2101.04840}.

\bibitem[{Goldberg(2017)}]{goldberg2017neural}
Yoav Goldberg. 2017.
\newblock Neural network methods for natural language processing.
\newblock \emph{Synthesis lectures on human language technologies},
  10(1):1--309.

\bibitem[{Goyal and Durrett(2020)}]{goyal-durrett-2020-neural}
Tanya Goyal and Greg Durrett. 2020.
\newblock \href {https://doi.org/10.18653/v1/2020.acl-main.22} {Neural
  syntactic preordering for controlled paraphrase generation}.
\newblock In \emph{Proceedings of the 58th Annual Meeting of the Association
  for Computational Linguistics}, pages 238--252, Online. Association for
  Computational Linguistics.

\bibitem[{Guntuku et~al.(2019)Guntuku, Li, Tay, and
  Ungar}]{guntuku2019studying}
Sharath~Chandra Guntuku, Mingyang Li, Louis Tay, and Lyle~H Ungar. 2019.
\newblock Studying cultural differences in emoji usage across the east and the
  west.
\newblock In \emph{Proceedings of the International AAAI Conference on Web and
  Social Media}, volume~13, pages 226--235.

\bibitem[{Gupta et~al.(2021)Gupta, Dhole, Tarway, Prabhakar, and
  Shrivastava}]{gupta2021candle}
Aadesh Gupta, Kaustubh~D. Dhole, Rahul Tarway, Swetha Prabhakar, and Ashish
  Shrivastava. 2021.
\newblock \href {http://arxiv.org/abs/2107.03884} {Candle: Decomposing
  conditional and conjunctive queries for task-oriented dialogue systems}.

\bibitem[{Harel-Canada(2021)}]{sibyl}
Fabrice Harel-Canada. 2021.
\newblock \href {https://github.com/fabriceyhc/Sibyl} {{Sibyl}}.

\bibitem[{Hendrickx et~al.(2013)Hendrickx, Kozareva, Nakov, S{\'e}aghdha,
  Szpakowicz, and Veale}]{hendrickx2013semeval}
Iris Hendrickx, Zornitsa Kozareva, Preslav Nakov, Diarmuid~{\'O} S{\'e}aghdha,
  Stan Szpakowicz, and Tony Veale. 2013.
\newblock Semeval-2013 task 4: Free paraphrases of noun compounds.
\newblock In \emph{Second Joint Conference on Lexical and Computational
  Semantics (* SEM), Volume 2: Proceedings of the Seventh International
  Workshop on Semantic Evaluation (SemEval 2013)}, pages 138--143.

\bibitem[{hyperreality@GitHub()}]{hyperreality}
hyperreality@GitHub.
\newblock American british english translator.
\newblock
  \url{https://github.com/hyperreality/American-British-English-Translator }.

\bibitem[{Jalalzai et~al.(2020)Jalalzai, Colombo, Clavel, Gaussier, Varni,
  Vignon, and Sabourin}]{Jalalzai2020repersentations}
Hamid Jalalzai, Pierre Colombo, Chlo\'{e} Clavel, Eric Gaussier, Giovanna
  Varni, Emmanuel Vignon, and Anne Sabourin. 2020.
\newblock \href
  {https://proceedings.neurips.cc/paper/2020/file/2cfa3753d6a524711acb5fce38eeca1a-Paper.pdf}
  {Heavy-tailed representations, text polarity classification \&amp; data
  augmentation}.
\newblock In \emph{Advances in Neural Information Processing Systems},
  volume~33, pages 4295--4307. Curran Associates, Inc.

\bibitem[{Jia and Liang(2017{\natexlab{a}})}]{DBLP:conf/emnlp/JiaL17}
Robin Jia and Percy Liang. 2017{\natexlab{a}}.
\newblock \href {https://doi.org/10.18653/v1/d17-1215} {Adversarial examples
  for evaluating reading comprehension systems}.
\newblock In \emph{Proceedings of the 2017 Conference on Empirical Methods in
  Natural Language Processing, {EMNLP} 2017, Copenhagen, Denmark, September
  9-11, 2017}, pages 2021--2031. Association for Computational Linguistics.

\bibitem[{Jia and Liang(2017{\natexlab{b}})}]{jia-liang-2017-adversarial}
Robin Jia and Percy Liang. 2017{\natexlab{b}}.
\newblock \href {https://doi.org/10.18653/v1/D17-1215} {Adversarial examples
  for evaluating reading comprehension systems}.
\newblock In \emph{Proceedings of the 2017 Conference on Empirical Methods in
  Natural Language Processing}, pages 2021--2031, Copenhagen, Denmark.
  Association for Computational Linguistics.

\bibitem[{Jindal et~al.(2020)Jindal, Aharonov, Brahma, Zhu, and
  Li}]{jindal2020improved}
Ishan Jindal, Ranit Aharonov, Siddhartha Brahma, Huaiyu Zhu, and Yunyao Li.
  2020.
\newblock Improved semantic role labeling using parameterized neighborhood
  memory adaptation.
\newblock \emph{arXiv preprint arXiv:2011.14459}.

\bibitem[{Kaushik et~al.(2019)Kaushik, Hovy, and Lipton}]{kaushik2019learning}
Divyansh Kaushik, Eduard Hovy, and Zachary~C Lipton. 2019.
\newblock Learning the difference that makes a difference with
  counterfactually-augmented data.
\newblock \emph{arXiv preprint arXiv:1909.12434}.

\bibitem[{Khachatrian et~al.(2019)Khachatrian, Nersisyan, Hambardzumyan,
  Galstyan, Hakobyan, Arakelyan, Rzhetsky, and
  Galstyan}]{Khachatrian2019BioRelEx1B}
Hrant Khachatrian, Lilit Nersisyan, Karen Hambardzumyan, Tigran Galstyan, Anna
  Hakobyan, Arsen Arakelyan, A.~Rzhetsky, and A.~G. Galstyan. 2019.
\newblock Biorelex 1.0: Biological relation extraction benchmark.
\newblock In \emph{BioNLP@ACL}.

\bibitem[{Kiela et~al.(2021)Kiela, Bartolo, Nie, Kaushik, Geiger, Wu, Vidgen,
  Prasad, Singh, Ringshia, Ma, Thrush, Riedel, Waseem, Stenetorp, Jia, Bansal,
  Potts, and Williams}]{kiela-etal-2021-dynabench}
Douwe Kiela, Max Bartolo, Yixin Nie, Divyansh Kaushik, Atticus Geiger,
  Zhengxuan Wu, Bertie Vidgen, Grusha Prasad, Amanpreet Singh, Pratik Ringshia,
  Zhiyi Ma, Tristan Thrush, Sebastian Riedel, Zeerak Waseem, Pontus Stenetorp,
  Robin Jia, Mohit Bansal, Christopher Potts, and Adina Williams. 2021.
\newblock \href {https://doi.org/10.18653/v1/2021.naacl-main.324} {Dynabench:
  Rethinking benchmarking in {NLP}}.
\newblock In \emph{Proceedings of the 2021 Conference of the North American
  Chapter of the Association for Computational Linguistics: Human Language
  Technologies}, pages 4110--4124, Online. Association for Computational
  Linguistics.

\bibitem[{Kingsbury and Palmer(2002)}]{DBLP:conf/lrec/KingsburyP02}
Paul~R. Kingsbury and Martha Palmer. 2002.
\newblock \href
  {http://www.lrec-conf.org/proceedings/lrec2002/sumarios/283.htm} {From
  treebank to propbank}.
\newblock In \emph{Proceedings of the Third International Conference on
  Language Resources and Evaluation, {LREC} 2002, May 29-31, 2002, Las Palmas,
  Canary Islands, Spain}. European Language Resources Association.

\bibitem[{Ko{\v{c}}isk{\`y} et~al.(2018)Ko{\v{c}}isk{\`y}, Schwarz, Blunsom,
  Dyer, Hermann, Melis, and Grefenstette}]{kovcisky2018narrativeqa}
Tom{\'a}{\v{s}} Ko{\v{c}}isk{\`y}, Jonathan Schwarz, Phil Blunsom, Chris Dyer,
  Karl~Moritz Hermann, G{\'a}bor Melis, and Edward Grefenstette. 2018.
\newblock The narrativeqa reading comprehension challenge.
\newblock \emph{Transactions of the Association for Computational Linguistics},
  6:317--328.

\bibitem[{Kovatchev et~al.(2021)Kovatchev, Smith, Lee, and
  Devine}]{kovatchev-etal-2021-vectors}
Venelin Kovatchev, Phillip Smith, Mark Lee, and Rory Devine. 2021.
\newblock \href {https://doi.org/10.18653/v1/2021.acl-long.96} {Can vectors
  read minds better than experts? comparing data augmentation strategies for
  the automated scoring of children{'}s mindreading ability}.
\newblock In \emph{Proceedings of the 59th Annual Meeting of the Association
  for Computational Linguistics and the 11th International Joint Conference on
  Natural Language Processing (Volume 1: Long Papers)}, pages 1196--1206,
  Online. Association for Computational Linguistics.

\bibitem[{Krishna et~al.(2020)Krishna, Wieting, and
  Iyyer}]{krishna-etal-2020-reformulating}
Kalpesh Krishna, John Wieting, and Mohit Iyyer. 2020.
\newblock \href {https://doi.org/10.18653/v1/2020.emnlp-main.55} {Reformulating
  unsupervised style transfer as paraphrase generation}.
\newblock In \emph{Proceedings of the 2020 Conference on Empirical Methods in
  Natural Language Processing (EMNLP)}, pages 737--762, Online. Association for
  Computational Linguistics.

\bibitem[{Kumar et~al.(2019)Kumar, Bhattamishra, Bhandari, and
  Talukdar}]{dips2019}
Ashutosh Kumar, Satwik Bhattamishra, Manik Bhandari, and Partha Talukdar. 2019.
\newblock \href {https://www.aclweb.org/anthology/N19-1363} {Submodular
  optimization-based diverse paraphrasing and its effectiveness in data
  augmentation}.
\newblock In \emph{Proceedings of the 2019 Conference of the North {A}merican
  Chapter of the Association for Computational Linguistics: Human Language
  Technologies, Volume 1 (Long and Short Papers)}, pages 3609--3619,
  Minneapolis, Minnesota. Association for Computational Linguistics.

\bibitem[{Lample et~al.(2018)Lample, Conneau, Ranzato, Denoyer, and
  Jégou}]{lample2018word}
Guillaume Lample, Alexis Conneau, Marc'Aurelio Ranzato, Ludovic Denoyer, and
  Hervé Jégou. 2018.
\newblock \href {https://openreview.net/forum?id=H196sainb} {Word translation
  without parallel data}.
\newblock In \emph{International Conference on Learning Representations}.

\bibitem[{Laserna et~al.(2014)Laserna, Seih, and Pennebaker}]{laserna2014like}
Charlyn~M Laserna, Yi-Tai Seih, and James~W Pennebaker. 2014.
\newblock Um... who like says you know: Filler word use as a function of age,
  gender, and personality.
\newblock \emph{Journal of Language and Social Psychology}, 33(3):328--338.

\bibitem[{Lauer(1995)}]{lauer1995designing}
Mark Lauer. 1995.
\newblock \emph{Designing Statistical Language Learners: Experiments on Noun
  Compounds}.
\newblock Ph.D. thesis.

\bibitem[{Lee et~al.(2018)Lee, He, and Zettlemoyer}]{lee2018higher}
Kenton Lee, Luheng He, and Luke Zettlemoyer. 2018.
\newblock Higher-order coreference resolution with coarse-to-fine inference.
\newblock In \emph{Proceedings of the 2018 Conference of the North American
  Chapter of the Association for Computational Linguistics: Human Language
  Technologies, Volume 2 (Short Papers)}, pages 687--692.

\bibitem[{Lewis et~al.(2020)Lewis, Liu, Goyal, Ghazvininejad, Mohamed, Levy,
  Stoyanov, and Zettlemoyer}]{lewis2020bart}
Mike Lewis, Yinhan Liu, Naman Goyal, Marjan Ghazvininejad, Abdelrahman Mohamed,
  Omer Levy, Veselin Stoyanov, and Luke Zettlemoyer. 2020.
\newblock Bart: Denoising sequence-to-sequence pre-training for natural
  language generation, translation, and comprehension.
\newblock In \emph{Proceedings of the 58th Annual Meeting of the Association
  for Computational Linguistics}, pages 7871--7880.

\bibitem[{Lhoest et~al.(2021{\natexlab{a}})Lhoest, del Moral, Jernite, Thakur,
  von Platen, Patil, Chaumond, Drame, Plu, Tunstall, Davison, Šaško,
  Chhablani, Malik, Brandeis, Scao, Sanh, Xu, Patry, McMillan-Major, Schmid,
  Gugger, Delangue, Matussière, Debut, Bekman, Cistac, Goehringer, Mustar,
  Lagunas, Rush, and Wolf}]{lhoest2021datasets}
Quentin Lhoest, Albert~Villanova del Moral, Yacine Jernite, Abhishek Thakur,
  Patrick von Platen, Suraj Patil, Julien Chaumond, Mariama Drame, Julien Plu,
  Lewis Tunstall, Joe Davison, Mario Šaško, Gunjan Chhablani, Bhavitvya
  Malik, Simon Brandeis, Teven~Le Scao, Victor Sanh, Canwen Xu, Nicolas Patry,
  Angelina McMillan-Major, Philipp Schmid, Sylvain Gugger, Clément Delangue,
  Théo Matussière, Lysandre Debut, Stas Bekman, Pierric Cistac, Thibault
  Goehringer, Victor Mustar, François Lagunas, Alexander~M. Rush, and Thomas
  Wolf. 2021{\natexlab{a}}.
\newblock \href {http://arxiv.org/abs/2109.02846} {Datasets: A community
  library for natural language processing}.

\bibitem[{Lhoest et~al.(2021{\natexlab{b}})Lhoest, del Moral, von Platen, Wolf,
  Šaško, Jernite, Thakur, Tunstall, Patil, Drame, Chaumond, Plu, Davison,
  Brandeis, Sanh, Scao, Xu, Patry, Liu, McMillan-Major, Schmid, Gugger, Raw,
  Lesage, Lozhkov, Carrigan, Matussière, von Werra, Debut, Bekman, and
  Delangue}]{quentin_lhoest_2021_5579268}
Quentin Lhoest, Albert~Villanova del Moral, Patrick von Platen, Thomas Wolf,
  Mario Šaško, Yacine Jernite, Abhishek Thakur, Lewis Tunstall, Suraj Patil,
  Mariama Drame, Julien Chaumond, Julien Plu, Joe Davison, Simon Brandeis,
  Victor Sanh, Teven~Le Scao, Kevin~Canwen Xu, Nicolas Patry, Steven Liu,
  Angelina McMillan-Major, Philipp Schmid, Sylvain Gugger, Nathan Raw, Sylvain
  Lesage, Anton Lozhkov, Matthew Carrigan, Théo Matussière, Leandro von
  Werra, Lysandre Debut, Stas Bekman, and Clément Delangue.
  2021{\natexlab{b}}.
\newblock \href {https://doi.org/10.5281/zenodo.5579268} {huggingface/datasets:
  1.14.0}.

\bibitem[{Li et~al.(2020{\natexlab{a}})Li, Zhang, Peng, Chen, Brockett, Sun,
  and Dolan}]{li2020contextualized}
Dianqi Li, Yizhe Zhang, Hao Peng, Liqun Chen, Chris Brockett, Ming-Ting Sun,
  and Bill Dolan. 2020{\natexlab{a}}.
\newblock Contextualized perturbation for textual adversarial attack.
\newblock \emph{arXiv preprint arXiv:2009.07502}.

\bibitem[{Li et~al.(2020{\natexlab{b}})Li, Zhang, Peng, Chen, Brockett, Sun,
  and Dolan}]{DBLP:journals/corr/abs-2009-07502}
Dianqi Li, Yizhe Zhang, Hao Peng, Liqun Chen, Chris Brockett, Ming{-}Ting Sun,
  and Bill Dolan. 2020{\natexlab{b}}.
\newblock \href {http://arxiv.org/abs/2009.07502} {Contextualized perturbation
  for textual adversarial attack}.
\newblock \emph{CoRR}, abs/2009.07502.

\bibitem[{Li and Specia(2019)}]{Li_2019}
Zhenhao Li and Lucia Specia. 2019.
\newblock \href {https://doi.org/10.18653/v1/d19-5543} {Improving neural
  machine translation robustness via data augmentation: Beyond
  back-translation}.
\newblock \emph{Proceedings of the 5th Workshop on Noisy User-generated Text
  (W-NUT 2019)}.

\bibitem[{Lin et~al.(2019)Lin, Zhou, Shen, Zhou, Bhagavatula, Choi, and
  Ren}]{lin2019commongen}
Bill~Yuchen Lin, Wangchunshu Zhou, Ming Shen, Pei Zhou, Chandra Bhagavatula,
  Yejin Choi, and Xiang Ren. 2019.
\newblock Commongen: A constrained text generation challenge for generative
  commonsense reasoning.
\newblock \emph{arXiv preprint arXiv:1911.03705}.

\bibitem[{Liu et~al.(2021)Liu, Winata, and Fung}]{liu2021continual}
Zihan Liu, Genta~Indra Winata, and Pascale Fung. 2021.
\newblock Continual mixed-language pre-training for extremely low-resource
  neural machine translation.
\newblock \emph{arXiv preprint arXiv:2105.03953}.

\bibitem[{Liu et~al.(2020)Liu, Winata, Lin, Xu, and Fung}]{liu2020attention}
Zihan Liu, Genta~Indra Winata, Zhaojiang Lin, Peng Xu, and Pascale Fung. 2020.
\newblock Attention-informed mixed-language training for zero-shot
  cross-lingual task-oriented dialogue systems.
\newblock In \emph{Proceedings of the AAAI Conference on Artificial
  Intelligence}, volume~34, pages 8433--8440.

\bibitem[{Logeswaran et~al.(2018)Logeswaran, Lee, and
  Bengio}]{DBLP:conf/nips/LogeswaranLB18}
Lajanugen Logeswaran, Honglak Lee, and Samy Bengio. 2018.
\newblock \href
  {https://proceedings.neurips.cc/paper/2018/hash/7cf64379eb6f29a4d25c4b6a2df713e4-Abstract.html}
  {Content preserving text generation with attribute controls}.
\newblock In \emph{Advances in Neural Information Processing Systems 31: Annual
  Conference on Neural Information Processing Systems 2018, NeurIPS 2018,
  December 3-8, 2018, Montr{\'{e}}al, Canada}, pages 5108--5118.

\bibitem[{Lu et~al.(2019)Lu, Mardziel, Wu, Amancharla, and
  Datta}]{lu2019gender}
Kaiji Lu, Piotr Mardziel, Fangjing Wu, Preetam Amancharla, and Anupam Datta.
  2019.
\newblock \href {http://arxiv.org/abs/1807.11714} {Gender bias in neural
  natural language processing}.

\bibitem[{Ma(2019)}]{ma2019nlpaug}
Edward Ma. 2019.
\newblock Nlp augmentation.
\newblock https://github.com/makcedward/nlpaug.

\bibitem[{Maas et~al.(2011)Maas, Daly, Pham, Huang, Ng, and
  Potts}]{maas2011learning}
Andrew Maas, Raymond~E Daly, Peter~T Pham, Dan Huang, Andrew~Y Ng, and
  Christopher Potts. 2011.
\newblock Learning word vectors for sentiment analysis.
\newblock In \emph{Proceedings of the 49th annual meeting of the association
  for computational linguistics: Human language technologies}, pages 142--150.

\bibitem[{Marcus et~al.(1993)Marcus, Santorini, and
  Marcinkiewicz}]{marcus-etal-1993-building}
Mitchell~P. Marcus, Beatrice Santorini, and Mary~Ann Marcinkiewicz. 1993.
\newblock \href {https://aclanthology.org/J93-2004} {Building a large annotated
  corpus of {E}nglish: The {P}enn {T}reebank}.
\newblock \emph{Computational Linguistics}, 19(2):313--330.

\bibitem[{Marivate and Sefara(2020)}]{marivate2020improving}
Vukosi Marivate and Tshephisho Sefara. 2020.
\newblock Improving short text classification through global augmentation
  methods.
\newblock In \emph{International Cross-Domain Conference for Machine Learning
  and Knowledge Extraction}, pages 385--399. Springer.

\bibitem[{McCoy et~al.(2019)McCoy, Pavlick, and Linzen}]{mccoy-etal-2019-right}
Tom McCoy, Ellie Pavlick, and Tal Linzen. 2019.
\newblock \href {https://doi.org/10.18653/v1/P19-1334} {Right for the wrong
  reasons: Diagnosing syntactic heuristics in natural language inference}.
\newblock In \emph{Proceedings of the 57th Annual Meeting of the Association
  for Computational Linguistics}, pages 3428--3448, Florence, Italy.
  Association for Computational Linguistics.

\bibitem[{Merriam-Webster()}]{mw:diacritic}
Merriam-Webster.
\newblock \href
  {https://www.merriam-webster.com/words-at-play/how-to-use-and-understand-diacritics-diacritical-marks}
  {What is a diacritic, anyway?}

\bibitem[{Mille et~al.(2021)Mille, Dhole, Mahamood, Perez-Beltrachini, Gangal,
  Kale, van Miltenburg, and Gehrmann}]{mille2021automatic}
Simon Mille, Kaustubh~D. Dhole, Saad Mahamood, Laura Perez-Beltrachini, Varun
  Gangal, Mihir Kale, Emiel van Miltenburg, and Sebastian Gehrmann. 2021.
\newblock \href {http://arxiv.org/abs/2106.09069} {Automatic construction of
  evaluation suites for natural language generation datasets}.

\bibitem[{Miller(1998)}]{miller1998wordnet}
George~A Miller. 1998.
\newblock \emph{WordNet: An electronic lexical database}.
\newblock MIT press.

\bibitem[{Mishra et~al.(2020)Mishra, He, and
  Belli}]{DBLP:journals/corr/abs-2008-03415}
Shubhanshu Mishra, Sijun He, and Luca Belli. 2020.
\newblock \href {http://arxiv.org/abs/2008.03415} {Assessing demographic bias
  in named entity recognition}.
\newblock \emph{CoRR}, abs/2008.03415.

\bibitem[{Morris et~al.(2020)Morris, Lifland, Yoo, Grigsby, Jin, and
  Qi}]{morris2020textattack}
John Morris, Eli Lifland, Jin~Yong Yoo, Jake Grigsby, Di~Jin, and Yanjun Qi.
  2020.
\newblock Textattack: A framework for adversarial attacks, data augmentation,
  and adversarial training in nlp.
\newblock In \emph{Proceedings of the 2020 Conference on Empirical Methods in
  Natural Language Processing: System Demonstrations}, pages 119--126.

\bibitem[{Namysl et~al.(2020)Namysl, Behnke, and
  K{\"o}hler}]{namysl-etal-2020-nat}
Marcin Namysl, Sven Behnke, and Joachim K{\"o}hler. 2020.
\newblock \href {https://doi.org/10.18653/v1/2020.acl-main.138} {{NAT}:
  Noise-aware training for robust neural sequence labeling}.
\newblock In \emph{Proceedings of the 58th Annual Meeting of the Association
  for Computational Linguistics}, pages 1501--1517, Online. Association for
  Computational Linguistics.

\bibitem[{Namysl et~al.(2021)Namysl, Behnke, and
  K{\"o}hler}]{namysl-etal-2021-empirical}
Marcin Namysl, Sven Behnke, and Joachim K{\"o}hler. 2021.
\newblock \href {https://doi.org/10.18653/v1/2021.findings-acl.27} {Empirical
  error modeling improves robustness of noisy neural sequence labeling}.
\newblock In \emph{Findings of the Association for Computational Linguistics:
  ACL-IJCNLP 2021}, pages 314--329, Online. Association for Computational
  Linguistics.

\bibitem[{Nekoto et~al.(2020)Nekoto, Marivate, Matsila, Fasubaa, Fagbohungbe,
  Akinola, Muhammad, Kabongo~Kabenamualu, Osei, Sackey, Niyongabo, Macharm,
  Ogayo, Ahia, Berhe, Adeyemi, Mokgesi-Selinga, Okegbemi, Martinus, Tajudeen,
  Degila, Ogueji, Siminyu, Kreutzer, Webster, Ali, Abbott, Orife, Ezeani,
  Dangana, Kamper, Elsahar, Duru, Kioko, Espoir, van Biljon, Whitenack,
  Onyefuluchi, Emezue, Dossou, Sibanda, Bassey, Olabiyi, Ramkilowan, {\"O}ktem,
  Akinfaderin, and Bashir}]{nekoto-etal-2020-participatory}
Wilhelmina Nekoto, Vukosi Marivate, Tshinondiwa Matsila, Timi Fasubaa, Taiwo
  Fagbohungbe, Solomon~Oluwole Akinola, Shamsuddeen Muhammad, Salomon
  Kabongo~Kabenamualu, Salomey Osei, Freshia Sackey, Rubungo~Andre Niyongabo,
  Ricky Macharm, Perez Ogayo, Orevaoghene Ahia, Musie~Meressa Berhe, Mofetoluwa
  Adeyemi, Masabata Mokgesi-Selinga, Lawrence Okegbemi, Laura Martinus,
  Kolawole Tajudeen, Kevin Degila, Kelechi Ogueji, Kathleen Siminyu, Julia
  Kreutzer, Jason Webster, Jamiil~Toure Ali, Jade Abbott, Iroro Orife, Ignatius
  Ezeani, Idris~Abdulkadir Dangana, Herman Kamper, Hady Elsahar, Goodness Duru,
  Ghollah Kioko, Murhabazi Espoir, Elan van Biljon, Daniel Whitenack,
  Christopher Onyefuluchi, Chris~Chinenye Emezue, Bonaventure F.~P. Dossou,
  Blessing Sibanda, Blessing Bassey, Ayodele Olabiyi, Arshath Ramkilowan, Alp
  {\"O}ktem, Adewale Akinfaderin, and Abdallah Bashir. 2020.
\newblock \href {https://doi.org/10.18653/v1/2020.findings-emnlp.195}
  {Participatory research for low-resourced machine translation: A case study
  in {A}frican languages}.
\newblock In \emph{Findings of the Association for Computational Linguistics:
  EMNLP 2020}, pages 2144--2160, Online. Association for Computational
  Linguistics.

\bibitem[{Nguyen et~al.(2021)Nguyen, Murray, and Chiang}]{nguyen2021data}
Toan~Q Nguyen, Kenton Murray, and David Chiang. 2021.
\newblock \href {https://arxiv.org/abs/2105.01691} {Data augmentation by
  concatenation for low-resource translation: A mystery and a solution}.
\newblock In \emph{Proceedings of the International Workshop on Spoken Language
  Translation}, Online. Association for Computational Linguistics.

\bibitem[{Pais(2019)}]{paisthesis}
Vasile~Florian Pais. 2019.
\newblock \emph{Contributions to semantic processing of texts; Identification
  of entities and relations between textual units; Case study on Romanian
  language}.
\newblock Ph.D. thesis.

\bibitem[{Palmer et~al.(2005)Palmer, Kingsbury, and
  Gildea}]{DBLP:journals/coling/PalmerKG05}
Martha Palmer, Paul~R. Kingsbury, and Daniel Gildea. 2005.
\newblock \href {https://doi.org/10.1162/0891201053630264} {The proposition
  bank: An annotated corpus of semantic roles}.
\newblock \emph{Comput. Linguistics}, 31(1):71--106.

\bibitem[{Parikh et~al.(2019)Parikh, Sai, Nema, and
  Khapra}]{DBLP:journals/corr/abs-1904-02651}
Soham Parikh, Ananya~B. Sai, Preksha Nema, and Mitesh~M. Khapra. 2019.
\newblock \href {http://arxiv.org/abs/1904.02651} {Eliminet: {A} model for
  eliminating options for reading comprehension with multiple choice
  questions}.
\newblock \emph{CoRR}, abs/1904.02651.

\bibitem[{Park and Lee(2020)}]{park2020neural}
Kyubyong Park and Seanie Lee. 2020.
\newblock \href {http://arxiv.org/abs/2004.03136} {g2pm: {A} neural
  grapheme-to-phoneme conversion package for mandarin chinese based on a new
  open benchmark dataset}.
\newblock \emph{CoRR}, abs/2004.03136.

\bibitem[{Pierse(2021)}]{Pierse_Transformers_Interpret_2021}
Charles Pierse. 2021.
\newblock \href {https://github.com/cdpierse/transformers-interpret}
  {{Transformers Interpret}}.

\bibitem[{Piktus et~al.(2019)Piktus, Edizel, Bojanowski, Grave, Ferreira, and
  Silvestri}]{piktus-etal-2019-misspelling}
Aleksandra Piktus, Necati~Bora Edizel, Piotr Bojanowski, Edouard Grave, Rui
  Ferreira, and Fabrizio Silvestri. 2019.
\newblock \href {https://doi.org/10.18653/v1/N19-1326} {Misspelling oblivious
  word embeddings}.
\newblock In \emph{Proceedings of the 2019 Conference of the North {A}merican
  Chapter of the Association for Computational Linguistics: Human Language
  Technologies, Volume 1 (Long and Short Papers)}, pages 3226--3234,
  Minneapolis, Minnesota. Association for Computational Linguistics.

\bibitem[{Pitler et~al.(2008)Pitler, Raghupathy, Mehta, Nenkova, Lee, and
  Joshi}]{pitler-etal-2008-easily}
Emily Pitler, Mridhula Raghupathy, Hena Mehta, Ani Nenkova, Alan Lee, and
  Aravind Joshi. 2008.
\newblock \href {https://www.aclweb.org/anthology/C08-2022} {Easily
  identifiable discourse relations}.
\newblock In \emph{Coling 2008: Companion volume: Posters}, pages 87--90,
  Manchester, UK. Coling 2008 Organizing Committee.

\bibitem[{Ponkiya et~al.(2020)Ponkiya, Murthy, Bhattacharyya, and
  Palshikar}]{ponkiya2020looking}
Girishkumar Ponkiya, Rudra Murthy, Pushpak Bhattacharyya, and Girish Palshikar.
  2020.
\newblock Looking inside noun compounds: Unsupervised prepositional and free
  paraphrasing using language models.
\newblock In \emph{Proceedings of the 2020 Conference on Empirical Methods in
  Natural Language Processing: Findings}, pages 4313--4323.

\bibitem[{Ponkiya et~al.(2018)Ponkiya, Patel, Bhattacharyya, and
  Palshikar}]{ponkiya2018treat}
Girishkumar Ponkiya, Kevin Patel, Pushpak Bhattacharyya, and Girish Palshikar.
  2018.
\newblock Treat us like the sequences we are: Prepositional paraphrasing of
  noun compounds using lstm.
\newblock In \emph{Proceedings of the 27th International Conference on
  Computational Linguistics}, pages 1827--1836.

\bibitem[{Prasad et~al.(2008)Prasad, Dinesh, Lee, Miltsakaki, Robaldo, Joshi,
  and Webber}]{prasad-etal-2008-penn}
Rashmi Prasad, Nikhil Dinesh, Alan Lee, Eleni Miltsakaki, Livio Robaldo,
  Aravind Joshi, and Bonnie Webber. 2008.
\newblock \href
  {http://www.lrec-conf.org/proceedings/lrec2008/pdf/754_paper.pdf} {The {P}enn
  {D}iscourse {T}ree{B}ank 2.0.}
\newblock In \emph{Proceedings of the Sixth International Conference on
  Language Resources and Evaluation ({LREC}'08)}, Marrakech, Morocco. European
  Language Resources Association (ELRA).

\bibitem[{Pruksachatkun et~al.(2021)Pruksachatkun, Krishna, Dhamala, Gupta, and
  Chang}]{pruksachatkun-etal-2021-robustness}
Yada Pruksachatkun, Satyapriya Krishna, Jwala Dhamala, Rahul Gupta, and Kai-Wei
  Chang. 2021.
\newblock \href {https://doi.org/10.18653/v1/2021.findings-acl.294} {Does
  robustness improve fairness? approaching fairness with word substitution
  robustness methods for text classification}.
\newblock In \emph{Findings of the Association for Computational Linguistics:
  ACL-IJCNLP 2021}, pages 3320--3331, Online. Association for Computational
  Linguistics.

\bibitem[{Qin et~al.(2020)Qin, Ni, Zhang, and Che}]{ijcai2020-533}
Libo Qin, Minheng Ni, Yue Zhang, and Wanxiang Che. 2020.
\newblock \href {https://doi.org/10.24963/ijcai.2020/533} {Cosda-ml:
  Multi-lingual code-switching data augmentation for zero-shot cross-lingual
  nlp}.
\newblock In \emph{Proceedings of the Twenty-Ninth International Joint
  Conference on Artificial Intelligence, {IJCAI-20}}, pages 3853--3860.
  International Joint Conferences on Artificial Intelligence Organization.
\newblock Main track.

\bibitem[{Radford et~al.(2019)Radford, Wu, Child, Luan, Amodei, and
  Sutskever}]{radford2019language}
Alec Radford, Jeff Wu, Rewon Child, David Luan, Dario Amodei, and Ilya
  Sutskever. 2019.
\newblock Language models are unsupervised multitask learners.

\bibitem[{Raffel et~al.(2019)Raffel, Shazeer, Roberts, Lee, Narang, Matena,
  Zhou, Li, and Liu}]{raffel2019exploring}
Colin Raffel, Noam Shazeer, Adam Roberts, Katherine Lee, Sharan Narang, Michael
  Matena, Yanqi Zhou, Wei Li, and Peter~J. Liu. 2019.
\newblock \href {http://arxiv.org/abs/1910.10683} {Exploring the limits of
  transfer learning with a unified text-to-text transformer}.

\bibitem[{Raffo(2021)}]{DVN/MSEGSJ_2021}
Julio Raffo. 2021.
\newblock \href {https://doi.org/10.7910/DVN/MSEGSJ} {{WGND 2.0}}.

\bibitem[{Raunak et~al.(2021)Raunak, Menezes, and
  Junczys-Dowmunt}]{raunak-etal-2021-curious}
Vikas Raunak, Arul Menezes, and Marcin Junczys-Dowmunt. 2021.
\newblock \href {https://doi.org/10.18653/v1/2021.naacl-main.92} {The curious
  case of hallucinations in neural machine translation}.
\newblock In \emph{Proceedings of the 2021 Conference of the North American
  Chapter of the Association for Computational Linguistics: Human Language
  Technologies}, pages 1172--1183, Online. Association for Computational
  Linguistics.

\bibitem[{Ravichander et~al.(2021)Ravichander, Dalmia, Ryskina, Metze, Hovy,
  and Black}]{ravichander2021noiseqa}
Abhilasha Ravichander, Siddharth Dalmia, Maria Ryskina, Florian Metze, Eduard
  Hovy, and Alan~W Black. 2021.
\newblock \href {https://arxiv.org/abs/2102.08345} {{NoiseQA: Challenge Set
  Evaluation for User-Centric Question Answering}}.
\newblock In \emph{Conference of the European Chapter of the Association for
  Computational Linguistics (EACL)}, Online.

\bibitem[{Regina et~al.(2020)Regina, Meyer, and
  Goutal}]{DBLP:journals/corr/abs-2007-02033}
Mehdi Regina, Maxime Meyer, and S{\'{e}}bastien Goutal. 2020.
\newblock \href {http://arxiv.org/abs/2007.02033} {Text data augmentation:
  Towards better detection of spear-phishing emails}.
\newblock \emph{CoRR}, abs/2007.02033.

\bibitem[{Ribeiro et~al.(2020)Ribeiro, Wu, Guestrin, and
  Singh}]{ribeiro2020beyond}
Marco~Tulio Ribeiro, Tongshuang Wu, Carlos Guestrin, and Sameer Singh. 2020.
\newblock \href {https://doi.org/10.18653/v1/2020.acl-main.442} {Beyond
  accuracy: Behavioral testing of {NLP} models with {C}heck{L}ist}.
\newblock In \emph{Proceedings of the 58th Annual Meeting of the Association
  for Computational Linguistics}, pages 4902--4912, Online. Association for
  Computational Linguistics.

\bibitem[{Shi et~al.(2021)Shi, Livescu, and
  Gimpel}]{shi-etal-2021-substructure}
Haoyue Shi, Karen Livescu, and Kevin Gimpel. 2021.
\newblock \href {https://doi.org/10.18653/v1/2021.findings-acl.307}
  {Substructure substitution: Structured data augmentation for {NLP}}.
\newblock In \emph{Findings of the Association for Computational Linguistics:
  ACL-IJCNLP 2021}, pages 3494--3508, Online. Association for Computational
  Linguistics.

\bibitem[{Shi and Lin(2019{\natexlab{a}})}]{shi2019simple}
Peng Shi and Jimmy Lin. 2019{\natexlab{a}}.
\newblock Simple bert models for relation extraction and semantic role
  labeling.
\newblock \emph{arXiv preprint arXiv:1904.05255}.

\bibitem[{Shi and Lin(2019{\natexlab{b}})}]{DBLP:journals/corr/abs-1904-05255}
Peng Shi and Jimmy Lin. 2019{\natexlab{b}}.
\newblock \href {http://arxiv.org/abs/1904.05255} {Simple {BERT} models for
  relation extraction and semantic role labeling}.
\newblock \emph{CoRR}, abs/1904.05255.

\bibitem[{Shrivastava et~al.(2021)Shrivastava, Dhole, Bhatt, and
  Raghunath}]{shrivastava-etal-2021-saying}
Ashish Shrivastava, Kaustubh Dhole, Abhinav Bhatt, and Sharvani Raghunath.
  2021.
\newblock \href {https://doi.org/10.18653/v1/2021.acl-short.13} {{S}aying {N}o
  is {A}n {A}rt: {C}ontextualized {F}allback {R}esponses for {U}nanswerable
  {D}ialogue {Q}ueries}.
\newblock In \emph{Proceedings of the 59th Annual Meeting of the Association
  for Computational Linguistics and the 11th International Joint Conference on
  Natural Language Processing (Volume 2: Short Papers)}, pages 87--92, Online.
  Association for Computational Linguistics.

\bibitem[{Shwartz and Dagan(2018)}]{shwartz2018paraphrase}
Vered Shwartz and Ido Dagan. 2018.
\newblock Paraphrase to explicate: Revealing implicit noun-compound relations.
\newblock In \emph{Proceedings of the 56th Annual Meeting of the Association
  for Computational Linguistics (Volume 1: Long Papers)}, pages 1200--1211.

\bibitem[{Si et~al.(2021)Si, Zhang, Qi, Liu, Wang, Liu, and
  Sun}]{si-etal-2021-better}
Chenglei Si, Zhengyan Zhang, Fanchao Qi, Zhiyuan Liu, Yasheng Wang, Qun Liu,
  and Maosong Sun. 2021.
\newblock \href {https://doi.org/10.18653/v1/2021.findings-acl.137} {Better
  robustness by more coverage: Adversarial and mixup data augmentation for
  robust finetuning}.
\newblock In \emph{Findings of the Association for Computational Linguistics:
  ACL-IJCNLP 2021}, pages 1569--1576, Online. Association for Computational
  Linguistics.

\bibitem[{Smith(2007)}]{smith2007overview}
R.~Smith. 2007.
\newblock \href {https://doi.org/10.1109/ICDAR.2007.4376991} {An overview of
  the tesseract {OCR} engine}.
\newblock In \emph{9th International Conference on Document Analysis and
  Recognition {(ICDAR} 2007), 23-26 September, Curitiba, Paran{\'{a}}, Brazil},
  pages 629--633. {IEEE} Computer Society.

\bibitem[{Socher et~al.(2013)Socher, Perelygin, Wu, Chuang, Manning, Ng, and
  Potts}]{socher2013recursive}
Richard Socher, Alex Perelygin, Jean Wu, Jason Chuang, Christopher~D. Manning,
  Andrew Ng, and Christopher Potts. 2013.
\newblock \href {https://www.aclweb.org/anthology/D13-1170} {Recursive deep
  models for semantic compositionality over a sentiment treebank}.
\newblock In \emph{Proceedings of the 2013 Conference on Empirical Methods in
  Natural Language Processing}, pages 1631--1642, Seattle, Washington, USA.
  Association for Computational Linguistics.

\bibitem[{Srivastava et~al.(2022)Srivastava, Rastogi, Rao, Shoeb, Abid, Fisch,
  Brown, Santoro, Gupta, Garriga-Alonso et~al.}]{srivastava2022beyond}
Aarohi Srivastava, Abhinav Rastogi, Abhishek Rao, Abu Awal~Md Shoeb, Abubakar
  Abid, Adam Fisch, Adam~R Brown, Adam Santoro, Aditya Gupta, Adri{\`a}
  Garriga-Alonso, et~al. 2022.
\newblock Beyond the imitation game: Quantifying and extrapolating the
  capabilities of language models.
\newblock \emph{arXiv preprint arXiv:2206.04615}.

\bibitem[{Sugiyama and Yoshinaga(2019)}]{sugiyama-yoshinaga-2019-data}
Amane Sugiyama and Naoki Yoshinaga. 2019.
\newblock \href {https://doi.org/10.18653/v1/D19-6504} {Data augmentation using
  back-translation for context-aware neural machine translation}.
\newblock In \emph{Proceedings of the Fourth Workshop on Discourse in Machine
  Translation (DiscoMT 2019)}, pages 35--44, Hong Kong, China. Association for
  Computational Linguistics.

\bibitem[{Sun et~al.(2021)Sun, Webster, Shah, Wang, and
  Johnson}]{genderrewrite}
Tony Sun, Kellie Webster, Apurva Shah, William~Yang Wang, and Melvin Johnson.
  2021.
\newblock \href {http://arxiv.org/abs/2102.06788} {They, them, theirs:
  Rewriting with gender-neutral english}.
\newblock \emph{CoRR}, abs/2102.06788.

\bibitem[{Tan et~al.(2021{\natexlab{a}})Tan, Hazarika, Ng, Poria, and
  Zimmermann}]{tan-etal-2021-causal}
Fiona~Anting Tan, Devamanyu Hazarika, See-Kiong Ng, Soujanya Poria, and Roger
  Zimmermann. 2021{\natexlab{a}}.
\newblock \href {https://aclanthology.org/2021.cinlp-1.1} {Causal augmentation
  for causal sentence classification}.
\newblock In \emph{Proceedings of the First Workshop on Causal Inference and
  NLP}, pages 1--20, Punta Cana, Dominican Republic. Association for
  Computational Linguistics.

\bibitem[{Tan and Joty(2021)}]{tan-joty-2021-code-mixing}
Samson Tan and Shafiq Joty. 2021.
\newblock \href {https://doi.org/10.18653/v1/2021.naacl-main.282} {Code-mixing
  on sesame street: Dawn of the adversarial polyglots}.
\newblock In \emph{Proceedings of the 2021 Conference of the North American
  Chapter of the Association for Computational Linguistics: Human Language
  Technologies}, pages 3596--3616, Online. Association for Computational
  Linguistics.

\bibitem[{Tan et~al.(2021{\natexlab{b}})Tan, Joty, Baxter, Taeihagh, Bennett,
  and Kan}]{tan-etal-2021-reliability}
Samson Tan, Shafiq Joty, Kathy Baxter, Araz Taeihagh, Gregory~A. Bennett, and
  Min-Yen Kan. 2021{\natexlab{b}}.
\newblock \href {https://doi.org/10.18653/v1/2021.acl-long.321} {Reliability
  testing for natural language processing systems}.
\newblock In \emph{Proceedings of the 59th Annual Meeting of the Association
  for Computational Linguistics and the 11th International Joint Conference on
  Natural Language Processing (Volume 1: Long Papers)}, pages 4153--4169,
  Online. Association for Computational Linguistics.

\bibitem[{Tan et~al.(2020)Tan, Joty, Kan, and Socher}]{tan-etal-2020-morphin}
Samson Tan, Shafiq Joty, Min-Yen Kan, and Richard Socher. 2020.
\newblock \href {https://doi.org/10.18653/v1/2020.acl-main.263} {It{'}s
  morphin{'} time! {C}ombating linguistic discrimination with inflectional
  perturbations}.
\newblock In \emph{Proceedings of the 58th Annual Meeting of the Association
  for Computational Linguistics}, pages 2920--2935, Online. Association for
  Computational Linguistics.

\bibitem[{Tiedemann and Thottingal(2020)}]{TiedemannThottingal:EAMT2020}
J{\"o}rg Tiedemann and Santhosh Thottingal. 2020.
\newblock {OPUS-MT} — {B}uilding open translation services for the {W}orld.
\newblock In \emph{Proceedings of the 22nd Annual Conferenec of the European
  Association for Machine Translation (EAMT)}, Lisbon, Portugal.

\bibitem[{Vijayakumar et~al.(2018)Vijayakumar, Cogswell, Selvaraju, Sun, Lee,
  Crandall, and Batra}]{AAAI1817329}
Ashwin Vijayakumar, Michael Cogswell, Ramprasaath Selvaraju, Qing Sun, Stefan
  Lee, David Crandall, and Dhruv Batra. 2018.
\newblock \href
  {https://www.aaai.org/ocs/index.php/AAAI/AAAI18/paper/view/17329} {Diverse
  beam search for improved description of complex scenes}.

\bibitem[{Vijayakumar et~al.(2016)Vijayakumar, Cogswell, Selvaraju, Sun, Lee,
  Crandall, and Batra}]{DBLP:journals/corr/VijayakumarCSSL16}
Ashwin~K. Vijayakumar, Michael Cogswell, Ramprasaath~R. Selvaraju, Qing Sun,
  Stefan Lee, David~J. Crandall, and Dhruv Batra. 2016.
\newblock \href {http://arxiv.org/abs/1610.02424} {Diverse beam search:
  Decoding diverse solutions from neural sequence models}.
\newblock \emph{CoRR}, abs/1610.02424.

\bibitem[{Wang et~al.(2019{\natexlab{a}})Wang, Singh, Michael, Hill, Levy, and
  Bowman}]{wang2018glue}
Alex Wang, Amanpreet Singh, Julian Michael, Felix Hill, Omer Levy, and
  Samuel~R. Bowman. 2019{\natexlab{a}}.
\newblock \href {https://openreview.net/forum?id=rJ4km2R5t7} {{GLUE:} {A}
  multi-task benchmark and analysis platform for natural language
  understanding}.
\newblock In \emph{7th International Conference on Learning Representations,
  {ICLR} 2019, New Orleans, LA, USA, May 6-9, 2019}. OpenReview.net.

\bibitem[{Wang et~al.(2021{\natexlab{a}})Wang, Liu, Gui, Zhang, Zou, Zhou, Ye,
  Zhang, Zheng, Pang, Wu, Li, Zhang, Ma, Fei, Cai, Zhao, Hu, Yan, Tan, Hu,
  Bian, Liu, Qin, Zhu, Xing, Fu, Zhang, Peng, Zheng, Zhou, Wei, Qiu, and
  Huang}]{wang-etal-2021-textflint}
Xiao Wang, Qin Liu, Tao Gui, Qi~Zhang, Yicheng Zou, Xin Zhou, Jiacheng Ye,
  Yongxin Zhang, Rui Zheng, Zexiong Pang, Qinzhuo Wu, Zhengyan Li, Chong Zhang,
  Ruotian Ma, Zichu Fei, Ruijian Cai, Jun Zhao, Xingwu Hu, Zhiheng Yan, Yiding
  Tan, Yuan Hu, Qiyuan Bian, Zhihua Liu, Shan Qin, Bolin Zhu, Xiaoyu Xing,
  Jinlan Fu, Yue Zhang, Minlong Peng, Xiaoqing Zheng, Yaqian Zhou, Zhongyu Wei,
  Xipeng Qiu, and Xuanjing Huang. 2021{\natexlab{a}}.
\newblock \href {https://doi.org/10.18653/v1/2021.acl-demo.41} {{T}ext{F}lint:
  Unified multilingual robustness evaluation toolkit for natural language
  processing}.
\newblock In \emph{Proceedings of the 59th Annual Meeting of the Association
  for Computational Linguistics and the 11th International Joint Conference on
  Natural Language Processing: System Demonstrations}, pages 347--355, Online.
  Association for Computational Linguistics.

\bibitem[{Wang et~al.(2019{\natexlab{b}})Wang, Che, Guo, Liu, and
  Liu}]{wang2019cross}
Yuxuan Wang, Wanxiang Che, Jiang Guo, Yijia Liu, and Ting Liu.
  2019{\natexlab{b}}.
\newblock Cross-lingual bert transformation for zero-shot dependency parsing.
\newblock In \emph{Proceedings of the 2019 Conference on Empirical Methods in
  Natural Language Processing and the 9th International Joint Conference on
  Natural Language Processing (EMNLP-IJCNLP)}, pages 5721--5727.

\bibitem[{Wang et~al.(2021{\natexlab{b}})Wang, Che, Titov, Cohen, Lei, and
  Liu}]{wang-etal-2021-closer}
Yuxuan Wang, Wanxiang Che, Ivan Titov, Shay~B. Cohen, Zhilin Lei, and Ting Liu.
  2021{\natexlab{b}}.
\newblock \href {https://doi.org/10.18653/v1/2021.findings-acl.207} {A closer
  look into the robustness of neural dependency parsers using better
  adversarial examples}.
\newblock In \emph{Findings of the Association for Computational Linguistics:
  ACL-IJCNLP 2021}, pages 2344--2354, Online. Association for Computational
  Linguistics.

\bibitem[{Wei and Zou(2019)}]{DBLP:conf/emnlp/WeiZ19}
Jason~W. Wei and Kai Zou. 2019.
\newblock \href {https://doi.org/10.18653/v1/D19-1670} {{EDA:} easy data
  augmentation techniques for boosting performance on text classification
  tasks}.
\newblock In \emph{Proceedings of the 2019 Conference on Empirical Methods in
  Natural Language Processing and the 9th International Joint Conference on
  Natural Language Processing, {EMNLP-IJCNLP} 2019, Hong Kong, China, November
  3-7, 2019}, pages 6381--6387. Association for Computational Linguistics.

\bibitem[{Wieting and Gimpel(2017)}]{wieting-17-millions}
John Wieting and Kevin Gimpel. 2017.
\newblock Pushing the limits of paraphrastic sentence embeddings with millions
  of machine translations.
\newblock In \emph{arXiv preprint arXiv:1711.05732}.

\bibitem[{Wieting et~al.(2017)Wieting, Mallinson, and
  Gimpel}]{wieting-17-backtrans}
John Wieting, Jonathan Mallinson, and Kevin Gimpel. 2017.
\newblock Learning paraphrastic sentence embeddings from back-translated
  bitext.
\newblock In \emph{Proceedings of Empirical Methods in Natural Language
  Processing}.

\bibitem[{Williams et~al.(2017)Williams, Nangia, and
  Bowman}]{williams2017broad}
Adina Williams, Nikita Nangia, and Samuel~R Bowman. 2017.
\newblock A broad-coverage challenge corpus for sentence understanding through
  inference.
\newblock \emph{arXiv preprint arXiv:1704.05426}.

\bibitem[{Wilson et~al.(2020)Wilson, Magdy, McGillivray, Garimella, and
  Tyson}]{wilson-etal-2020-urban}
Steven Wilson, Walid Magdy, Barbara McGillivray, Kiran Garimella, and Gareth
  Tyson. 2020.
\newblock \href {https://aclanthology.org/2020.lrec-1.586} {Urban dictionary
  embeddings for slang {NLP} applications}.
\newblock In \emph{Proceedings of the 12th Language Resources and Evaluation
  Conference}, pages 4764--4773, Marseille, France. European Language Resources
  Association.

\bibitem[{Wiseman and Rush(2016)}]{wiseman-rush-2016-sequence}
Sam Wiseman and Alexander~M. Rush. 2016.
\newblock \href {https://doi.org/10.18653/v1/D16-1137} {Sequence-to-sequence
  learning as beam-search optimization}.
\newblock In \emph{Proceedings of the 2016 Conference on Empirical Methods in
  Natural Language Processing}, pages 1296--1306, Austin, Texas. Association
  for Computational Linguistics.

\bibitem[{Wolf et~al.(2020)Wolf, Debut, Sanh, Chaumond, Delangue, Moi, Cistac,
  Rault, Louf, Funtowicz, Davison, Shleifer, von Platen, Ma, Jernite, Plu, Xu,
  Le~Scao, Gugger, Drame, Lhoest, and Rush}]{wolf-etal-2020-transformers}
Thomas Wolf, Lysandre Debut, Victor Sanh, Julien Chaumond, Clement Delangue,
  Anthony Moi, Pierric Cistac, Tim Rault, Remi Louf, Morgan Funtowicz, Joe
  Davison, Sam Shleifer, Patrick von Platen, Clara Ma, Yacine Jernite, Julien
  Plu, Canwen Xu, Teven Le~Scao, Sylvain Gugger, Mariama Drame, Quentin Lhoest,
  and Alexander Rush. 2020.
\newblock \href {https://doi.org/10.18653/v1/2020.emnlp-demos.6} {Transformers:
  State-of-the-art natural language processing}.
\newblock In \emph{Proceedings of the 2020 Conference on Empirical Methods in
  Natural Language Processing: System Demonstrations}, pages 38--45, Online.
  Association for Computational Linguistics.

\bibitem[{Wu et~al.(2021)Wu, Ribeiro, Heer, and Weld}]{wu2021polyjuice}
Tongshuang Wu, Marco~Tulio Ribeiro, Jeffrey Heer, and Daniel~S Weld. 2021.
\newblock Polyjuice: Generating counterfactuals for explaining, evaluating, and
  improving models.
\newblock In \emph{Proceedings of the 59th Annual Meeting of the Association
  for Computational Linguistics}.

\bibitem[{Xie et~al.(2020)Xie, Dai, Hovy, Luong, and Le}]{xie2020unsupervised}
Qizhe Xie, Zihang Dai, Eduard Hovy, Thang Luong, and Quoc Le. 2020.
\newblock Unsupervised data augmentation for consistency training.
\newblock \emph{Advances in Neural Information Processing Systems}, 33.

\bibitem[{Xu et~al.(2020)Xu, Dong, Yu, Tian, Liu, Li, and
  Zhang}]{xu2020cluener2020}
Liang Xu, Qianqian Dong, Cong Yu, Yin Tian, Weitang Liu, Lu~Li, and Xuanwei
  Zhang. 2020.
\newblock Cluener2020: Fine-grained name entity recognition for chinese.
\newblock \emph{arXiv preprint arXiv:2001.04351}.

\bibitem[{Yaseen and Langer(2021)}]{yaseen-and-langer-backtranslation-ner}
Usama Yaseen and Stefan Langer. 2021.
\newblock \href {http://arxiv.org/abs/2108.11703} {Data augmentation for
  low-resource named entity recognition using backtranslation}.
\newblock \emph{CoRR}, abs/2108.11703.

\bibitem[{Yi et~al.(2010)Yi, Steyvers, Lee, and Dry}]{yi2010wisdom}
Sheng~Kung Yi, Mark Steyvers, Michael Lee, and Matthew Dry. 2010.
\newblock Wisdom of the crowds in minimum spanning tree problems.
\newblock In \emph{Proceedings of the Annual Meeting of the Cognitive Science
  Society}, volume~32.

\bibitem[{Yorke()}]{ButterFingers}
Alex Yorke.
\newblock {butter-fingers}.
\newblock \url{https://github.com/alexyorke/butter-fingers}.

\bibitem[{Yunfei()}]{Chinese-Names-Corpus}
Yunfei.
\newblock { Chinese-Names-Corpus }.
\newblock \url{https://github.com/wainshine/Chinese-Names-Corpus}.

\bibitem[{Zhang et~al.(2021)Zhang, Shin, Choi, and Ho}]{zhang2021smat}
Jing Zhang, Bonggun Shin, Jinho~D Choi, and Joyce~C Ho. 2021.
\newblock Smat: An attention-based deep learning solution to the automation of
  schema matching.
\newblock In \emph{European Conference on Advances in Databases and Information
  Systems}, pages 260--274. Springer.

\bibitem[{Zhang et~al.(2019{\natexlab{a}})Zhang, Sheng, and
  Alhazmi}]{zhang2019adversarial}
Wei~Emma Zhang, Quan~Z. Sheng, and Ahoud Abdulrahmn~F. Alhazmi.
  2019{\natexlab{a}}.
\newblock \href {http://arxiv.org/abs/1901.06796} {Generating textual
  adversarial examples for deep learning models: {A} survey}.
\newblock \emph{CoRR}, abs/1901.06796.

\bibitem[{Zhang et~al.(2019{\natexlab{b}})Zhang, Baldridge, and
  He}]{DBLP:conf/naacl/ZhangBH19}
Yuan Zhang, Jason Baldridge, and Luheng He. 2019{\natexlab{b}}.
\newblock \href {https://doi.org/10.18653/v1/n19-1131} {{PAWS:} paraphrase
  adversaries from word scrambling}.
\newblock In \emph{Proceedings of the 2019 Conference of the North American
  Chapter of the Association for Computational Linguistics: Human Language
  Technologies, {NAACL-HLT} 2019, Minneapolis, MN, USA, June 2-7, 2019, Volume
  1 (Long and Short Papers)}, pages 1298--1308. Association for Computational
  Linguistics.

\bibitem[{Zhao et~al.(2019)Zhao, Chen, Zhang, Zhao, Liu, Lu, Chen, Deng, Ju,
  and Du}]{zhao2019uer}
Zhe Zhao, Hui Chen, Jinbin Zhang, Xin Zhao, Tao Liu, Wei Lu, Xi~Chen, Haotang
  Deng, Qi~Ju, and Xiaoyong Du. 2019.
\newblock Uer: An open-source toolkit for pre-training models.
\newblock \emph{EMNLP-IJCNLP 2019}, page 241.

\end{thebibliography}
